\documentclass{article} %
\usepackage{iclr2027_conference}
\usepackage{times}

\usepackage{amsmath,amsfonts,bm}

\def\eqref#1{equation~\ref{#1}}

\def\1{\bm{1}}

\DeclareMathAlphabet{\mathsfit}{\encodingdefault}{\sfdefault}{m}{sl}
\SetMathAlphabet{\mathsfit}{bold}{\encodingdefault}{\sfdefault}{bx}{n}

\usepackage{hyperref}
\usepackage{url}
\usepackage{graphicx}
\usepackage{amsmath}
\usepackage{amssymb}
\usepackage{float}
\usepackage{xspace}
\usepackage{booktabs}
\usepackage{graphicx}
\usepackage{subcaption}
\usepackage{xcolor}
\usepackage{wrapfig}
\usepackage{bm}
\usepackage{xparse}
\usepackage{booktabs}
\usepackage{arydshln}
\usepackage{tcolorbox}
\usepackage{fvextra}
\usepackage{newunicodechar}
\tcbuselibrary{breakable,skins}
\usepackage{colortbl}
\usepackage{makecell}
\usepackage{calc}

\usepackage{enumitem}

\newunicodechar{π}{\ensuremath{\pi}}
\newunicodechar{∀}{\ensuremath{\forall}}
\newunicodechar{ℕ}{\ensuremath{\mathbb{N}}}
\newunicodechar{λ}{\ensuremath{\lambda}}
\newunicodechar{∣}{\ensuremath{\mid}}
\newunicodechar{∨}{\ensuremath{\lor}}

\usepackage{titlesec}
\titlespacing*{\paragraph}{0pt}{0.05\baselineskip}{1em}

\NewDocumentCommand{\TODO}{g}{%
  \IfNoValueTF{#1}
    {\textcolor{red}{\textbf{[??]}}}
    {\textcolor{red}{\textbf{[TODO:}\ #1\textbf{]}}}%
}
\newcommand{\method}{\textsc{Magenta}\xspace}

\newtcolorbox{pipelinebox}[2][]{%
  enhanced,
  breakable,
  colback=white,
  colframe=black!45,
  colbacktitle=black!7,
  coltitle=black,
  fonttitle=\bfseries,
  title={#2},
  boxrule=0.65pt,
  arc=1.5mm,
  left=1.5mm,
  right=1.5mm,
  top=1.2mm,
  bottom=1.2mm,
  before skip=5pt,
  after skip=5pt,
  #1
}
\newcommand{\pipelinearrow}{%
  \begin{center}
    \vspace{-0.7em}\textcolor{black!45}{\Large$\Downarrow$}\vspace{-0.7em}
  \end{center}%
}

\floatstyle{plaintop}
\newfloat{algorithm}{t}{loa}
\floatname{algorithm}{Algorithm}
\restylefloat{algorithm}
\newlength{\algind}
\newcounter{plinenum}

\newcommand{\algstart}{\setcounter{plinenum}{0}\par\vspace{2pt}\hrule\vspace{4pt}%
  \small\setlength{\parindent}{0pt}\setlength{\parskip}{1pt}}
\newcommand{\algend}{\par\vspace{4pt}\hrule\vspace{2pt}}
\newcommand{\pline}[2]{\par\noindent\refstepcounter{plinenum}%
  \makebox[1.8em][r]{\theplinenum:}\hspace{0.5em}\hspace*{#1\algind}\ignorespaces#2}
\newcommand{\preq}[1]{\par\noindent\textbf{Require:}~#1}
\newcommand{\pret}[1]{\par\noindent\textbf{Ensure:}~#1}
\newcommand{\pcom}[1]{\hfill\footnotesize$\triangleright$~\emph{#1}}
\newcommand{\kw}[1]{\textbf{#1}}

\title{\textsc{Magenta}: Closing the Loop Between Mathematical Reasoning and Lean Verification}

\author{
Joshua Ong Jun Leang$^{1,2}$,  \textbf{Haonan Li}$^{1}$, Zheng Zhao$^{3}$, Xinyi Shang$^{4}$ \\
\textbf{Wenda Li}$^{3}$, \textbf{Zhengzhong Liu}$^{1}$,
\textbf{Eric Xing}$^{1}$, \textbf{Shay B. Cohen}$^{3}$, 
\textbf{Eleonora Giunchiglia}$^{2}$\medskip\\
$^{1}$Institute of Foundation Models, 
$^{2}$Imperial College London \\
$^{3}$University of Edinburgh, 
$^{4}$University College London \\
\texttt{\{Joshua.Ong,Haonan.Li\}@mbzuai.ac.ae}
}

\newcommand{\pnum}[1]{\makebox[\widthof{(- 00.00)}][r]{#1}}

\iclrfinalcopy

\begin{document}

\maketitle

\lhead{\textsc{Magenta}: Closing the Loop Between Mathematical Reasoning and Formal Verification}

\begin{abstract}
Most of mathematical knowledge has been 
communicated through so-called informal use of
mathematics and natural language. With large language models (LLMs) being highly adept in using natural language, they achieve strong performance, yet not perfect, in informal mathematical reasoning. Restraining LLMs to informal reasoning misses out on the opportunity to use the discrete verification abilities that machines offer through machine-checkable proofs. In this paper, we bridge the gap between informal and formal reasoning by integrating Lean signals into the informal reasoning process.
We introduce \method, a training-free agentic pipeline that, given only a natural-language problem, produces an answer, expresses it as a Lean~4 statement, and constructs a machine-checked proof.
A statement judge verifies whether the formalisation preserves the original problem, while an error-attribution judge routes failed attempts either to mathematical re-derivation or local Lean repair.
\method achieves $100\%$ accuracy across all evaluated olympiad benchmarks, including AIME~2025, AIME~2026, and HMMT February~2026. When paired with the open-weight \textsc{K2-Horizon-7B} reasoner, it solves all six IMO~2026 problems. Our analysis shows that statement adjudication is essential for preventing false certificates and that feedback-guided correction outperforms independent resampling on difficult problems.
\end{abstract}

\section{Introduction}

LLMs have become increasingly capable mathematical problem solvers~\citep{DBLP:journals/corr/abs-2601-03267,anthropic2026claude46systemcard,qwen3.6-27b}, solving International Mathematical Olympiad (IMO) problems~\citep{huang2025gemini} and beginning to contribute to mathematical research~\citep{DBLP:journals/corr/abs-2606-10806, DBLP:journals/corr/abs-2607-07779}.
Much of this progress relies on chain-of-thought (CoT) reasoning~\citep{wei2022chain} and test-time scaling~\citep{muennighoff2025s1}, where models generate long reasoning traces. %
Yet longer reasoning does not make correctness easier to guarantee: such traces routinely contain 
errors and dead ends~\citep{leang-etal-2025-comat,lyu-etal-2023-faithful}.
LLM judges~\citep{li2026rethinking} and process reward models~\citep{pmlr-v267-guan25f,zhang2025lessons} provide useful supervision over such traces, but ultimately remain learned proxies for correctness: they reproduce the failure modes of the models they evaluate and score plausible-but-incorrect reasoning~\citep{yosef2026rethinking,pmlr-v267-guan25f}.

Interactive theorem provers and proof assistants such as Lean~4~\citep{10.1007/978-3-030-79876-5_37}, Isabelle~\citep{DBLP:books/sp/Paulson94} and Rocq~\citep{CoquandHuet1988CoC} offer a different form of supervision: once a theorem is stated, a proof accepted by them provides a certificate of correctness. %
Systems built on this foundation have reached gold-medal level at the IMO~\citep{chen2025seed} and solve $99.2\%$ of miniF2F \citep{DBLP:journals/corr/abs-2109-00110} and $70.0\%$ of PutnamBench~\citep{tsoukalas2024putnambenchevaluatingneuraltheoremprovers,varambally2026hilbert}.
However, the models used are closed-source, or require parameter counts and search budgets beyond ordinary reach~\citep{chen2025seed, wang2026longcat}.
In addition, the results stated above largely address \emph{proof search after formalisation}.
Such benchmarks provide the system with a human-written formal statement and evaluate whether it can prove that statement~\citep{lin2025goedelproverv2scalingformaltheorem,DBLP:journals/corr/abs-2405-14333}.
They therefore assume away a central part of mathematical problem solving: determining precisely what the natural-language problem asks to prove.
Extending formal verification to natural-language mathematics consequently requires more than stronger proof search.

\begin{figure}[t]
    \centering
    \includegraphics[width=0.9\linewidth,page=1]{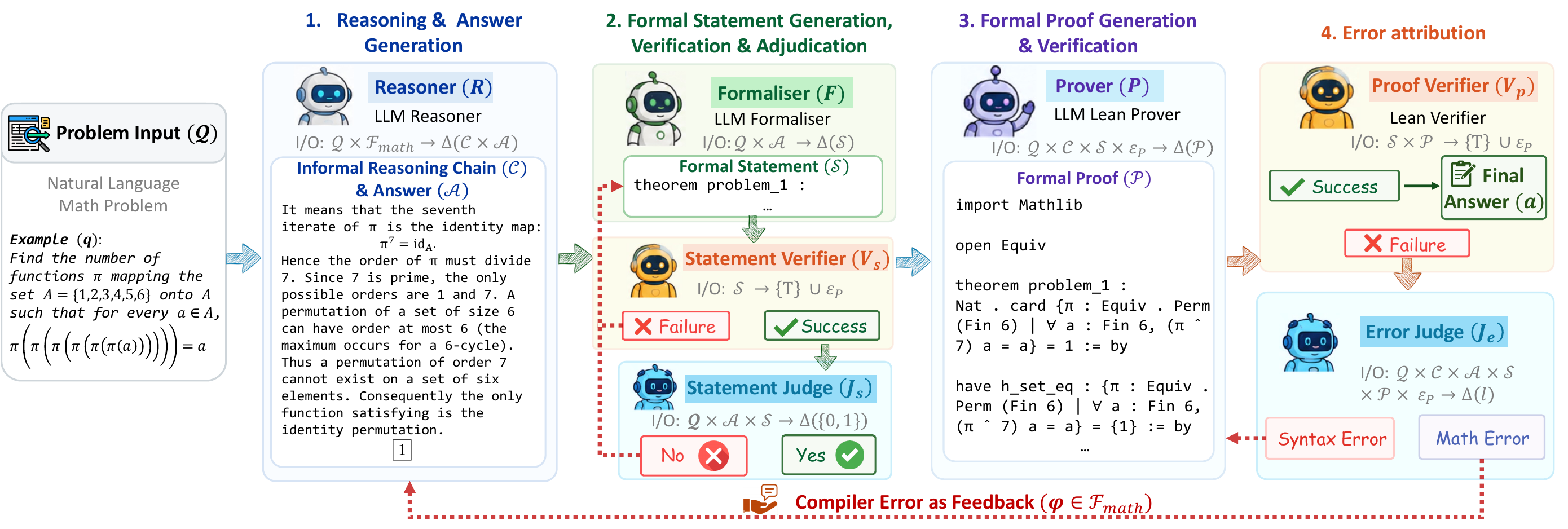}
    \caption{\textbf{Overview of \method.}
The pipeline maps a natural-language problem to an informal solution and candidate answer, a faithfully adjudicated Lean~4 statement, and a machine-checked proof. The illustrated example is taken from AIME~2026, where the initial reasoning incorrectly derives $\pi^{7}$ instead of the correct expression $\pi^{6}$. Rejected formal statements are resampled, while failed proof attempts are classified as either \textsc{Lean} or \textsc{Math} errors. \textsc{Lean} errors trigger local proof regeneration, whereas \textsc{Math} errors, such as the $\pi^{7}$ versus $\pi^{6}$ mistake shown here, are returned to the reasoner for mathematical re-derivation using verification feedback.}
    \label{fig:pipeline}
    \vspace{-4ex}
\end{figure}

To address these issues, we introduce \method, an end-to-end agentic pipeline, as shown in Figure~\ref{fig:pipeline}.
Given a natural-language problem, \method produces a reasoning chain and candidate answer, converts them into a Lean~4 statement, and constructs a machine-checked proof.
\method separates these tasks across four roles: A \emph{reasoner} derives an informal solution and commits to a candidate answer; a \emph{formaliser} converts the problem and that answer into a Lean~4 statement; a \emph{statement judge} explicitly checks whether that statement preserves the semantics of the original question and answer, rejecting omitted hypotheses, altered constants, and trivialising assumptions; and a \emph{prover} uses the informal reasoning as a proof plan.
The Lean kernel can certify only that a proof establishes the generated formal statement; it cannot determine whether that statement faithfully represents the original natural-language problem. This semantic gap is therefore the central reliability risk in our setting, and the statement judge mitigates it by checking their alignment before proof generation.
In addition, \method, in contrast to previous work, treats each certification failure as an \emph{attribution} problem, introducing an \emph{error judge} that labels it \textsc{Syntax} or \textsc{math} and reroutes the pipeline based on that.

\method verifies every problem on AIME~2025, AIME~2026, and HMMT February~2026 with four different reasoners, which indicates that these benchmarks are saturated under our evaluation setting.
Paired with \textsc{K2-Horizon-7B}, \method solves all six IMO~2026 problems, surpassing frontier closed models~\citep{DBLP:journals/corr/abs-2601-03267} and reasoners over two orders of magnitude larger~\citep{DBLP:journals/corr/abs-2607-24653}.
To our knowledge, this is the \textbf{\textit{smallest}} reported reasoner to solve all six problems.
Since perfect scores may raise concerns of contamination, we further test whether the results can be explained by one-shot recall: many problems reach a verified solution only after verification-guided self-correction, and we additionally re-evaluate the pipeline on answer-preserving paraphrases. Standalone reasoners degrade under such perturbations~\citep{DBLP:conf/iclr/MirzadehASTBF25,leang-etal-2025-comat}, whereas \method remains robust.
In summary, our contributions are as follows:
\begin{itemize}[topsep=0pt,leftmargin=*]
    \item 
    We introduce \method, a training-free agentic pipeline that bridges informal mathematical reasoning and formal verification, jointly producing an answer, a Lean~4 formalisation of the problem and answer, and a machine-checked proof through verification-guided self-correction.
    \item We further evaluate robustness to paraphrasing, characterise how verified performance scales with formalisation and proof-repair budgets, and ablate the two mechanisms that distinguish \method from a simple retry loop.

\end{itemize}

\section{Methodology}
\label{sec:method}

\method combines natural-language reasoning with verification in Lean. The reasoner proposes solutions but cannot guarantee their correctness, while Lean verifies proofs but cannot determine whether an error is due to an incorrect mathematical solution or a wrong formal implementation. \method closes this gap by attributing verification failures to the responsible component and routing the resulting feedback accordingly. %
Figure~\ref{fig:pipeline} gives an overview and Appendix~\ref{app:algorithms} states the notation and algorithm.

\subsection{Components}
\label{sec:components}

\method contains five components, each implemented by a prompted language model except for the deterministic verifier. To describe them, we first define the spaces over which they operate. 

Let $\mathcal{Q}$, $\mathcal{C}$, $\mathcal{A}$, $\mathcal{S}$ and $\mathcal{P}$ denote, respectively, the spaces of natural-language mathematics problems, informal reasoning chains, candidate answers, Lean~4 theorem statements and Lean~4 proof scripts. Let $\mathcal{E}_p$ and $\mathcal{E}_s$ denote the space of Lean diagnostics produced when a proof fails verification and when a statement fails to elaborate, respectively. Let $\varnothing\in\mathcal{E}_{s}\cap\mathcal{E}_{p}$ denote the absence of diagnostics, as occurs on the first iteration before any call to the Lean verifier. Let $\mathcal{F}_{\mathrm{math}} = (\mathcal{C} \times \mathcal{A} \times \mathcal{S} \times \mathcal{P} \times \mathcal{E}_p) \cup \{\varnothing\}$ denote the space of all correction feedback signals that get generated in case of a mathematical mistake, where $\varnothing$ denotes the absence of feedback. Given a set $\mathcal{B}$, we denote by $\Delta(\mathcal{B})$ the set of distributions over $\mathcal{B}$, and by $x \sim P$ a sample $x \in \mathcal{B}$ from a distribution $P \in \Delta(\mathcal{B})$.

\noindent{\bf Reasoner $R$.} The reasoner $R:\mathcal{Q}\times \mathcal{F}_{\mathrm{math}}\rightarrow\Delta(\mathcal{C}\times\mathcal{A})$ maps a problem $q\in\mathcal{Q}$ and a mathematical-correction signal $\varphi\in\mathcal{F}_{\mathrm{math}}$ to a distribution over pairs $(c,a)$, where $c\in\mathcal{C}$ is an informal reasoning chain and $a\in\mathcal{A}$ is a candidate answer. We write $(c,a)\sim R(q,\varphi)$ for a sample from the reasoner model obtained by inserting $q$ and $\varphi$ into its prompt and extracting $c$ and $a$ from the generated output. On the initial attempt, $\varphi=\varnothing$; a non-empty $\varphi$ arises only after the error judge returns \textsc{math}, indicating that the previous reasoning chain contains a mathematical error.

\noindent{\bf Formaliser $F$.} The formaliser $F:\mathcal{Q}\times\mathcal{A}\rightarrow\Delta(\mathcal{S})$ maps a problem $q \in\mathcal{Q}$ and candidate answer $a \in \mathcal{A}$ to a distribution over the space $\mathcal{S}$ of Lean~4 theorem statements. We write $s\sim F(q,a)$ for a sample from the formaliser model obtained by inserting $q$ and $a$ into its prompt and extracting the generated Lean statement. Conditioning on $a$ makes the statement assert the proposed answer rather than pose an open-ended existential goal. The resulting statement remains a candidate until it passes both Lean elaboration and semantic adjudication.

\noindent{\bf Prover $P$.} The prover $P:\mathcal{Q}\times\mathcal{C}\times\mathcal{S}\times \mathcal{E}_{\mathrm{p}} \rightarrow\Delta(\mathcal{P})$ maps a problem $q \in \mathcal{Q}$, an informal chain $c \in \mathcal{C}$, an accepted formal  statement $s \in \mathcal{S}$ and an optional diagnostic to a proof script $\epsilon \in \mathcal{E}_p$ to a distribution over the space $\mathcal{P}$ of Lean 4 proof scripts. We write $\pi\sim P(q,c,s,\epsilon)$, 
for a sample from the prover model obtained by inserting $q,c,s$ and $\epsilon$ into its prompt.
On the first attempt $\epsilon=\varnothing$, after a \textsc{Syntax} error the next attempt receives the latest diagnostic while $c$, $a$ and $s$ remain fixed.

\noindent{\bf Verifier $V$.} The verifier is the deterministic Lean~4 checking environment augmented with SafeVerify, a certification wrapper that rejects a submission whenever the declared theorem differs from the accepted statement or the script uses a forbidden construct such as \texttt{sorry}, \texttt{admit}, a new axiom or \texttt{native\_decide} (Appendix~\ref{app:hyperparameters}). The verifier consists of two deterministic checks: the {\sl statement verifier} $V_{\mathrm{s}}:\mathcal{S}\rightarrow\{\top\}\cup\mathcal{E}_{\mathrm{s}}$ and the {\sl proof verifier} $V_{\mathrm{p}}:\mathcal{S}\times\mathcal{P}\rightarrow\{\top\}\cup\mathcal{E}_{\mathrm{p}}$, where $\top$ denotes a successful check.  The verifier $V_s$ checks that a generated statement is closed and elaborates in Lean, which is a well-formedness check and does not establish that the statement is true or faithful to the problem. The verifier $V_p$ checks that $\pi$ is a complete, policy-compliant Lean proof of the unchanged statement $s$; otherwise, the verifier returns a diagnostic $\epsilon\in\mathcal{E}_{\mathrm{p}}$.

\noindent{\bf Judge $J$.} A single language model, prompted in two distinct ways, implements two distinct judging functions. The {\sl statement judge} $J_{\mathrm{s}}:\mathcal{Q}\times\mathcal{A}\times\mathcal{S}\rightarrow \Delta(\{0,1\})$ maps a problem $q\in\mathcal{Q}$, candidate answer $a\in\mathcal{A}$, and Lean 4 statement $s\in\mathcal{S}$ to a distribution over acceptance decisions. Acceptance
denotes that the Lean 4 statement faithfully represents the original problem and candidate answer. The {\sl error judge} $J_{\mathrm{e}}:\mathcal{Q}\times\mathcal{C}\times\mathcal{A}\times\mathcal{S}\times\mathcal{P}\times\mathcal{E}_{{p}}\rightarrow\Delta(\{\textsc{math},\textsc{Syntax}\})$, 
maps the complete context of a failed proof attempt to a distribution over error labels. We write $\ell\sim J_{\mathrm{e}}(q,c,a,s,\pi,\epsilon)$, where $\ell=\textsc{math}$ indicates that the mathematical reasoning should be revised, while $\ell=\textsc{Syntax}$ indicates that the error lies in the formal implementation. Thus, $J_{\mathrm{s}}$ determines whether the generated statement faithfully represents the intended claim, whereas $J_{\mathrm{e}}$ determines which component should be revised after proof verification fails.

\subsection{The \method procedure}

\method is an iterative verification-guided refinement procedure in which Lean~4 feedback is used to identify and correct errors in the reasoner’s proposed solution. Verification failures are attributed either to the mathematical reasoning or to its formal implementation, allowing the pipeline to revise only the component responsible for the failure. Below we describe each step of the procedure, while the pseudocode can be found in Appendix~\ref{app:algo}.

\paragraph{Step 1: Reasoning \& Answer Generation (Reasoner $R$).} Let $T$ be a user-defined upper bound on the number of calls to the reasoner, and let $t=0,\ldots,T-1$ index these calls. At iteration $t$, the reasoner samples
$(c_t,a_t)\sim R(q,\varphi_t)$,
where $\varphi_0=\varnothing$. For $t>0$, the reasoner is invoked only after the error judge $J_{\mathrm e}$ classifies the previous failed proof attempt as \textsc{math}; in this case, $\varphi_t\in\mathcal F_{\mathrm{math}}$ is constructed from the failed attempt and the corresponding verifier diagnostic. Thus, each \textsc{math} error triggers a new reasoning chain and candidate answer, while statement resampling and proof repair leave the current pair $(c_t,a_t)$ unchanged. The procedure terminates if a proof is verified or if the budget of $T$ reasoner calls is exhausted.

\paragraph{Step 2.a: Formal Statement Generation (Formaliser $F$).} Let $M$ be a user-defined upper bound on the number of statement samples drawn for a fixed candidate answer, and let $m=0,\ldots,M-1$ index these samples. At sample $m$, the formaliser draws $s_t^m\sim F(q,a_t)$, whose hypotheses encode the conditions of $q$ and the final answer provided by $a_t$.

\paragraph{Step 2.b: Formal Statement Verification (Statement Verifier $V_{\mathrm{s}}$).}
Each candidate statement $s_t^m$ is first submitted to the Lean~4 verifier. If $V_{\mathrm{s}}(s_t^m)\neq\top$, the statement is discarded and the formaliser $F$ is resampled. Otherwise, $s_t^m$ proceeds to statement adjudication. 
This verification step is restarted for every new reasoning pair $(c_t,a_t)$ and repeated within the same reasoning iteration until a statement passes verification, while $(c_t,a_t)$ remains fixed.

\paragraph{Step 2.c: Formal Statement Adjudication (Statement Judge $J_{\mathrm{s}}$).}
Given a candidate statement $s_t^m$ such that $V_{\mathrm{s}}(s_t^m)=\top$, the statement judge samples
$b\sim J_{\mathrm{s}}(q,a_t,s_t^m).$
The judge is tasked to accept $s_t^m$ only if it faithfully represents the original problem $q$ and assert the candidate answer $a_t$.
If $b=0$, the candidate is discarded and control returns to step~2, where the formaliser $F$ is resampled under the same statement budget $M$. If $b=1$, the candidate is accepted, denoted $s_t$, and passed to the prover.
A failure at either step~2.1 or step~2.2 does not trigger a new call to the reasoner; hence, statement resampling leaves the current reasoning pair $(c_t,a_t)$ unchanged. The statement judge therefore determines \emph{what} the pipeline attempts to prove, and is the component that connects Lean's guarantee about the accepted formal statement $s_t$ to the corresponding claim about the original problem $q$; \S\ref{sec:scope} discusses the resulting guarantee in detail.

\paragraph{Step 3: Formal Proof Generation (Prover $P$).} Let $K$ be a user-defined upper bound on the number of proof attempts for an accepted statement, and let $k=0,\ldots,K-1$ index these attempts. At attempt $k=0$, the prover samples $\pi_t^ k\sim P(q,c_t,s_t,\epsilon_k)$, where $\epsilon_0=\varnothing$ and the informal solution $c_t$ acts as a proof plan.

\paragraph{Step 4.a: Proof Verification (Proof Verifier $V_\mathrm{p}$).} Each candidate proof $\pi_t^k$ is submitted to the proof verifier $V_{\mathrm{p}}$. If $V_p(s_t, \pi_t^k) = \top$, then the Lean 4 verifier has accepted the proof, and the pipeline terminates and returns $(s_t, a_t, c_t, \pi_t^k)$. The associated informal solution $c_t$ and candidate answer $a_t$ are retained as the final reasoning chain and answer. Otherwise, $V_{\mathrm{p}}(s_t, \pi_t^k) = \epsilon_t^k$, i.e., the verifier has returned an error, which gets passed to the error judge.

\paragraph{Step 4.b: Error attribution (Error Judge $J_\mathrm{e}$).} Each candidate proof $\pi_t^k$ that failed the verifier check is submitted together with its corresponding error trace $\epsilon_t^k$ to the error judge $J_{\mathrm{e}}$, which returns $\ell\sim J_{\mathrm{e}}(q,c_t,a_t,s_t,\pi_t^k,\epsilon_t^k)$. If $\ell =$ \textsc{Syntax} then the algorithm loops back to step 3 a new candidate proof $\pi_t^{k+1}$ is sampled from the prover.  Thus, each \textsc{Syntax} error triggers a repair of the proof script alone, while $c_t,a_t$ and $s_t$ remain fixed.
If $\ell=\textsc{math}$ the informal solution is deemed false or insufficient, and the pipeline builds $\varphi_t\in\mathcal{F}_{\mathrm{math}}$ from the failed attempt and returns to step 1. Only a \textsc{math} label may revise the reasoning and answer.

\subsection{Role of the Judge and Scope of the Guarantee}
\label{sec:scope}

\paragraph{Why is statement adjudication necessary?}
Lean verifies a proof only relative to the generated statement:
$ V_{\mathrm{p}}(s,\pi)=\top$ entails that $ \pi \text{ is a valid proof of } s$,
but this does not imply that $s$ faithfully represents the original problem
$q$ and answer $a$. We refer to this as the \emph{autoformalisation gap}, which is due to the nature of LLMs and use of natural language.
The statement judge prevents these false certificates, working under the assumption that it is an easier task to confirm alignment between formal and informal statements compared to the generation of the formal statement. 
As this semantic check is
performed by a learned judge rather than the Lean kernel, the final output is a
\emph{soft certificate}: formally sound with respect to $s$, but only as
relevant to $q$ as the generated statement is faithful. Further ablation study of Statement Judge can be referred to \S\ref{sec:ablations}.

\paragraph{Why is error attribution necessary?}
When proof verification fails, $V_{\mathrm{p}}$ identifies where Lean failed. However, the same rejection may arise because the informal mathematics
is false or insufficient, or because a correct argument has been implemented
incorrectly in Lean. These cases require varying actions: the former requires
a new mathematical derivation, or the latter requires proof repair at fixed
mathematical content. The error judge supplies this decision:
\(
    \ell\sim J_{\mathrm{e}}(q,c,a,s,\pi,e),
\)
where $\ell\in$ \{\textsc{math},\textsc{Syntax}\}.
The resulting label determines the correction path:
\begin{equation}
    \ell=\textsc{Syntax}
    \ \Longrightarrow\
    \pi'\sim P(q,c,s,e),
    \qquad
    \ell=\textsc{math}
    \ \Longrightarrow\
    (c',a')\sim R(q,\varphi).
\end{equation}
A \textsc{Syntax} label preserves $(c,a,s)$ and uses the diagnostic to repair
the Lean proof, whereas a \textsc{math} label constructs
$\varphi\in\mathcal{F}_{\mathrm{math}}$ and returns control to the reasoner.
Without $J_{\mathrm{e}}$, the closed-loop pipeline is  underdetermined: a failed verification provides no basis for choosing between proof repair and mathematical re-derivation.

\paragraph{A mistake-bound perspective}
Given a question $q \in \mathcal{Q}$, let $\mathcal{A}(q)$ be the possible
answers (correct or not) that an LLM could provide for it.
We might ask the question, in terms of the mistake-bound
model of learning \citep{Cesa-Bianchi_Lugosi_2006}, what is the maximal number of ``mistakes'' $M(q)$ (verifier returning
0),\footnote{We note the term ``mistake'' is referring to lack of verifiability, not correctness, due to the \emph{autoformalisation gap} that remains unpreventable with current techniques}  our pipeline, or an improved one, would make before we find a verified answer? If the LLM does not loop in its answer
(and the verifier provides a significant feedback), and the context length is limited (hence, the answer too), then this ``dimension'' of $\mathcal{Q}$ (over all questions, see also the Littlestone dimension in online learning; \citealt{4568257}) is finite.

A simple result speaks about this dimension in terms of the feedback loop removing
a fraction $\gamma \in (0,1)$ of potential (mistaken) answers left at each step  shows that $M(q) \le \log_{1/(1-\gamma)} | \mathcal{A}(q) |$. For example, if at each point we remove half of the possible answers from the LLM prediction based on our verifier feedback, then $M(q) \le \log_2 | \mathcal{A}(q) |$. Given the complexity of LLM inference, it is difficult
to point and prove an exact value of $\gamma$ or $M(q)$, but we believe the mistake-bound model is a promising learning model to analyse such pipelines and we leave it for future work to estimate $\gamma$ in our case on a larger set of answers and questions. We provide further empirical analysis of the number of mistakes our pipeline makes in \S\ref{sec:results}.

\section{Experiments}
\vspace{-5pt}
\subsection{Experimental Setup}
\label{sec:exp-setup}
 
\paragraph{Models.} We vary $R$ across both open-weight and proprietary families: K2-Horizon-7B and 375B, Qwen3.8-27B and GPT-5.6-Sol (Codex). All other components are held fixed unless stated otherwise: Goedel-Formaliser-32B as $F$, DeepSeek-V4-Flash for both judge roles, and Leanstral-1.5 as the prover $P$. This provides an entirely open weight pipeline. \S\ref{sec:scaling} substitutes GPT-5.6-Sol for $F$ and for $P$. Proofs are checked with Lean \texttt{v4.29.1} and \textsc{Mathlib} under SafeVerify; decoding settings and the budgets $T$, $M$, $K$ are provided in Appendix~\ref{app:hyperparameters}.
 
\noindent{\bf Datasets.} We evaluate on AIME~2025, AIME~2026 and HMMT February~2026: natural-language competition problems with a numerical answer and no accompanying formal statement. We additionally evaluate our smallest reasoner \textsc{K2-Horizon-7B} with \method on IMO~2026 problems.

\noindent{\bf Baselines.} We compare against state-of-the-art baselines: Claude Opus~5, Gemini~3.7 Flash, and Kimi K3. 
We let each baseline only generate one answer (i.e., pass@1 rate).

\vspace{-5pt}
\subsection{Experimental Results}
\label{sec:results}

\begin{table}[t]
\centering
\caption{Accuracy (\%) on competition-level mathematics benchmarks, covering
93 problems in total. Light-grey rows denote results obtained with \method.
Green values in parentheses report absolute improvements over the corresponding
base reasoner in percentage points.}
\label{tab:math-benchmarks}
\small
\renewcommand{\arraystretch}{1.10}

\resizebox{0.95\linewidth}{!}{%
\begin{tabular}{lrrrr}
\toprule
\textbf{Model}
& \multicolumn{1}{l}{\textbf{AIME 2025}}
& \multicolumn{1}{l}{\textbf{AIME 2026}}
& \multicolumn{1}{l}{\textbf{HMMT Feb. 2026}}
& \multicolumn{1}{l}{\textbf{Overall}} \\
\midrule

Claude Opus 5
& 100.00 \pnum{}
& 100.00 \pnum{}
& 96.97 \pnum{}
& 98.92 \pnum{} \\

Gemini 3.7 Flash
& 100.00 \pnum{}
& 100.00 \pnum{}
& 90.91 \pnum{}
& 96.77 \pnum{} \\

Kimi K3
& 93.33 \pnum{}
& 90.00 \pnum{}
& 78.79 \pnum{}
& 87.10 \pnum{} \\

\midrule

K2-Horizon-7B
& 83.33 \pnum{}
& 80.00 \pnum{}
& 60.61 \pnum{}
& 74.19 \pnum{} \\

\rowcolor{gray!15}
\quad $+$ \method
& \textbf{100.00}
  \pnum{\textcolor{green!50!black}{\scriptsize($\uparrow 16.67$)}}
& \textbf{100.00}
  \pnum{\textcolor{green!50!black}{\scriptsize($\uparrow 20.00$)}}
& \textbf{100.00}
  \pnum{\textcolor{green!50!black}{\scriptsize($\uparrow 39.39$)}}
& \textbf{100.00}
  \pnum{\textcolor{green!50!black}{\scriptsize($\uparrow 25.81$)}} \\

\addlinespace[2pt]

K2-Horizon-375B
& 93.33 \pnum{}
& 90.00 \pnum{}
& 72.73 \pnum{}
& 84.95 \pnum{} \\

\rowcolor{gray!15}
\quad $+$ \method
& \textbf{100.00}
\pnum{\textcolor{green!50!black}{\scriptsize($\uparrow 6.67$)}}
& \textbf{100.00}
  \pnum{\textcolor{green!50!black}{\scriptsize($\uparrow 10.00$)}}
& \textbf{100.00}
  \pnum{\textcolor{green!50!black}{\scriptsize($\uparrow 27.27$)}}
& \textbf{100.00}
  \pnum{\textcolor{green!50!black}{\scriptsize($\uparrow 15.05$)}} \\

\addlinespace[2pt]

Qwen3.8-27B
& \textbf{100.00} \pnum{}
& 86.67 \pnum{}
& 87.88 \pnum{}
& 91.40 \pnum{} \\

\rowcolor{gray!15}
\quad $+$ \method
& \textbf{100.00}
  \pnum{\textcolor{green!50!black}{\scriptsize }}
& \textbf{100.00}
  \pnum{\textcolor{green!50!black}{\scriptsize($\uparrow 13.33$)}}
& \textbf{100.00}
  \pnum{\textcolor{green!50!black}{\scriptsize($\uparrow 12.12$)}}
& \textbf{100.00}
  \pnum{\textcolor{green!50!black}{\scriptsize($\uparrow 8.60$)}} \\

\addlinespace[2pt]

GPT-5.6-Sol (Codex)
& 90.00 \pnum{}
& 86.67 \pnum{}
& 78.79 \pnum{}
& 84.95 \pnum{} \\

\rowcolor{gray!15}
\quad $+$ \method
& \textbf{100.00}
  \pnum{\textcolor{green!50!black}{\scriptsize($\uparrow 10.00$)}}
& \textbf{100.00}
  \pnum{\textcolor{green!50!black}{\scriptsize($\uparrow 13.33$)}}
& \textbf{100.00}
  \pnum{\textcolor{green!50!black}{\scriptsize($\uparrow 21.21$)}}
& \textbf{100.00}
  \pnum{\textcolor{green!50!black}{\scriptsize($\uparrow 15.05$)}} \\

\bottomrule
\end{tabular}%
}
\vspace{-10pt}
\end{table}

\paragraph{\method is robust to model choice.} 
As shown in Table~\ref{tab:math-benchmarks}, \method's performance is  not tied to a particular base model used as reasoner. The average gain across the three tasks over the corresponding standalone reasoner ranges from $8.6$ points for Qwen3.8-27B to $25.8$ points for \textsc{K2-Horizon-7B}. In order to show that generalist closed models can be used instead of specialised open source models as formaliser and prover, we replace both of them with GPT-5.6-Sol (Codex). This once again yields the perfect scores shown in Table~\ref{tab:math-benchmarks}. Consistency across reasoner, formaliser and prover configurations suggests the gains arise from the verification-guided pipeline rather than from any single model; \S\ref{sec:scaling} analyses what these configurations cost.

\paragraph{\textsc{K2-Horizon-7B}$+$\method solves all IMO~2026 problems.}
Coupled with \method, \textsc{K2-Horizon-7B} solves all six IMO~2026 problems. To our knowledge, this is the smallest reported reasoner to score perfectly on this task. Unlike answer-only benchmarks, IMO solutions are inherently multi-step: they require answering a sequence of intermediate subproblems, whose results are then composed to obtain the final answer.
Despite perfect final answer accuracy and formal verification we observe a residual gap between the informal explanation and the formal statement and proof. 
We therefore append a final alignment judge, instantiated with \textsc{K2-Horizon-375B}, which returns \texttt{incomplete\_proof} when the two do not support the same conclusion (Appendix~\ref{app:imo}); its feedback yields substantially more complete reasoning chains in our qualitative analysis. 
Formal correctness is thus attainable with a compact reasoner, whereas a fully aligned, communicable informal derivation remains a separate challenge.

\begin{wrapfigure}[12]{r}{0.43\textwidth}
    \centering
    \vspace{-1.5em}
    \includegraphics[width=\linewidth]{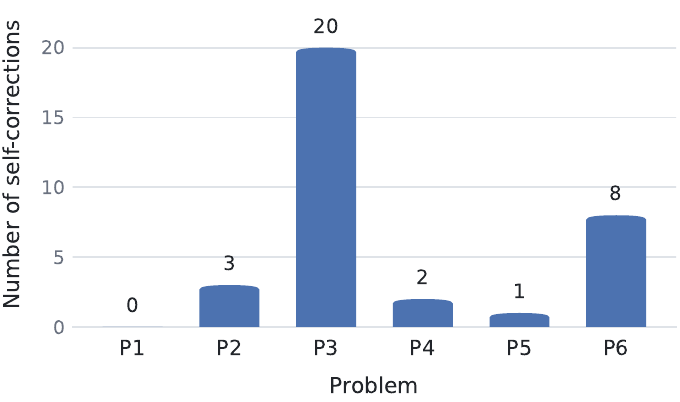}
    \caption{Self-corrections required by \textsc{K2-Horizon-7B}$+$\method on IMO~2026.    }
    \label{fig:imo-corrections}
    \vspace{-0.8em}
\end{wrapfigure}
\paragraph{Verification-guided correction trades test-time computation for model scale.} 
Most reasoners do not solve every problem first time, but their verified solve rates rise consistently across correction rounds. On AIME~2026, \textsc{K2-Horizon-7B} needs an average of $5$ correction rounds to reach a verified solution against $2$ for \textsc{K2-Horizon-375B}. 
Both ultimately attain perfect accuracy, but the larger reasoner gets there with fewer feedback cycles while the 7B model closes the gap by spending additional verification-guided computation. On the more complex IMO~2026, \textsc{K2-Horizon-7B} needs an average of $5.7$ rounds, with per-problem trajectories in Figure~\ref{fig:imo-corrections}. These averages count only revisions triggered by mathematical errors; compiler-level errors require further repair iterations (\S\ref{sec:scaling}).

\section{Analysis}
\label{sec:analysis}

\vspace{-5pt}
\subsection{Scaling Analysis}
\label{sec:scaling}

We analyse how test-time computation is distributed across the two stages of \method. We first examine how the choice of formaliser affects the number of samples required to obtain an accepted statement, and then compare the cost of different proof backends.

\subsubsection{Formal Statement Generation}
\label{sec:formalisation}

The formaliser governs how quickly the pipeline reaches a faithful, provable statement. \textsc{Codex} succeeds on its first call for approximately $66\%$ of AIME~2026 problems against $42\%$ for Goedel, and covers the benchmark within $6$ calls rather than approximately $120$ (Figure~\ref{fig:formal-statement}); at a six-call budget Goedel covers only about $68\%$. The distributions differ in shape, not merely location: 
\begin{wrapfigure}[12]{r}{0.41\textwidth}
    \centering
    \includegraphics[width=\linewidth]{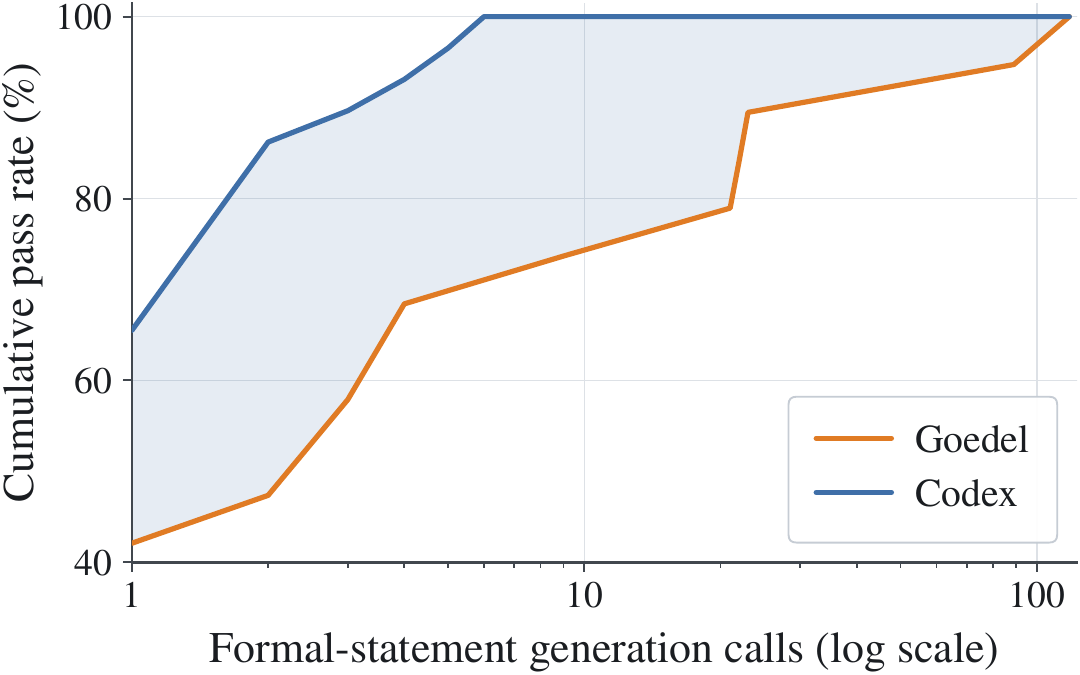}
    \caption{Cumulative pass rate against formal-statement generation calls on AIME~2026. 
    }
    \label{fig:formal-statement}
    \vspace{-3em}
\end{wrapfigure}
\textsc{Codex} is compact, and indeed $M=6$ represents a very small number of calls, whereas the Goedel curve spans two orders of magnitude and any practical $M$ truncates it. 
Truncation is not a correctness failure---adjudication rejects unfaithful statements before proof effort is spent---but an exhausted statement budget forfeits the problem, thus a long formalisation tail converts directly into lost coverage.

\subsubsection{Formal Proof Generation}
Both prover backends verify $100\%$ of AIME~2026, thus accuracy by itself cannot separate them. Reporting accuracy as a function of a budget rather than at a single operating point is standard practice in agent evaluation~\citep{kapoor2025ai} and in the test-time-scaling literature~\citep{muennighoff2025s1,snell2025scaling}. 
To make claims about self-repair meaningful, we report the cumulative pass rate:
\begin{equation}
    \mathrm{Pass}_{u}(\beta)
    =
    \frac{1}{|\mathcal{D}|}
    \sum_{q\in\mathcal{D}}
    \mathbf{1}\!\left[\mathrm{Ver}(q)=1 \wedge u(q)\le \beta\right],
    \label{eq:pass-curve}
\vspace{-1.5ex}
\end{equation}

\begin{wrapfigure}{r}{0.42\textwidth}
    \centering
    \vspace{-1em}
    \includegraphics[width=\linewidth]{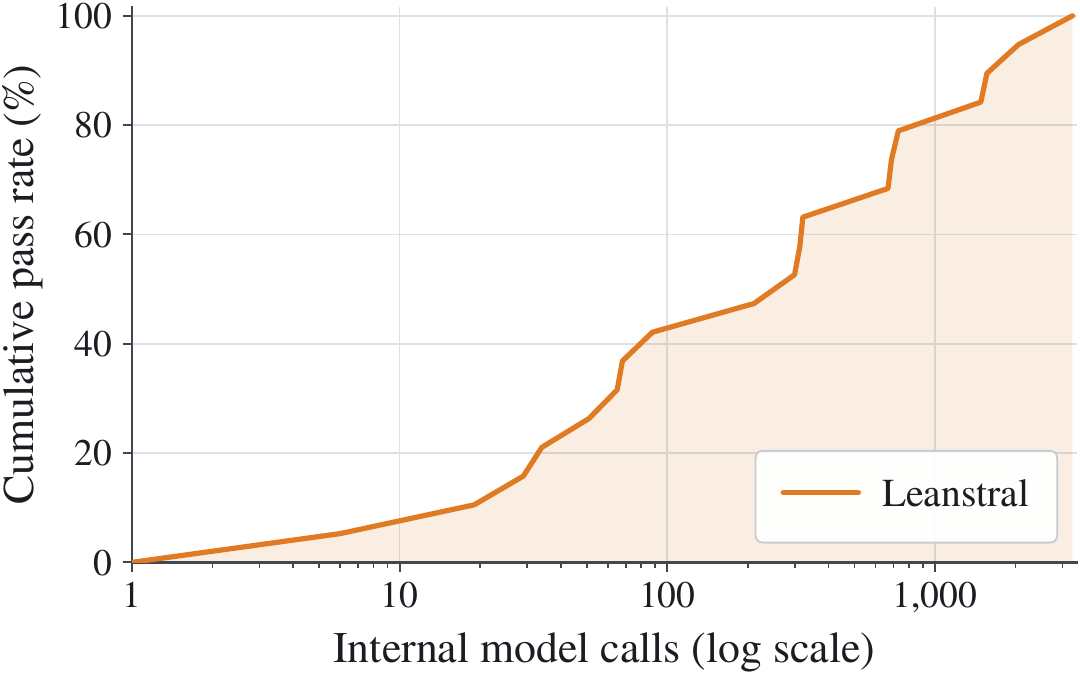}
    \caption{Internal \textsc{Leanstral} agent calls on the winning trajectory. This measure has no \textsc{Codex} counterpart, as its CLI does not expose per-invocation turns.}
    \label{fig:internal-calls}
    \vspace{-1em}
\end{wrapfigure}
where $u(q)$ measures measures the number of tokens necessary to produce the proof, $\operatorname{Ver}(q)$ indicates whether the proposed resolution of $q$ passes independent verification, and $\beta$ represents the tokens budget.
A single cost axis is not well defined across backends: \textsc{Leanstral} runs as a multi-turn agent, thus one invocation can hide hundreds of compiler queries, while a \textsc{Codex} CLI call reports one call regardless of what it does internally. We therefore count a cost as \emph{external} when a failure survives the prover's own repair attempts and returns to the error judge, and \emph{internal} when incurred inside a single invocation. External costs are defined identically for both backends; internal costs are observable only for \textsc{Leanstral} (Figure~\ref{fig:internal-calls}), and are paid on local hardware rather than billed per request.

\begin{figure}[t]
\centering
\begin{minipage}{\textwidth}
    \centering
    \begin{subfigure}[b]{0.49\linewidth}
        \includegraphics[width=\linewidth]{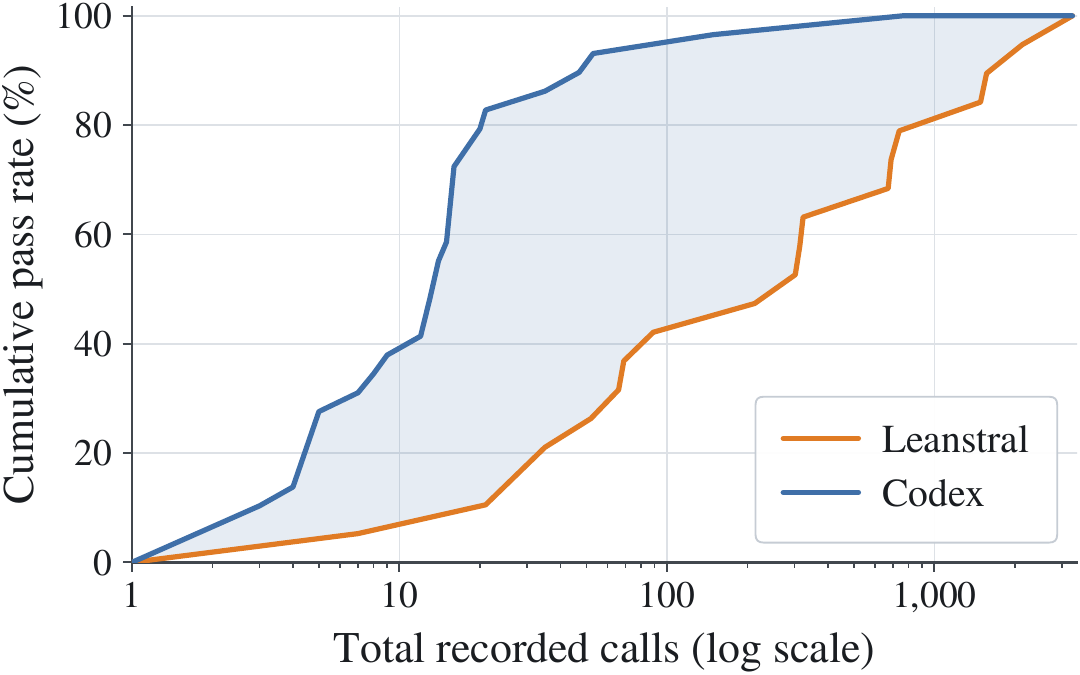}
        \caption{Total recorded calls}
        \label{fig:external-rounds}
    \end{subfigure}\hfill
    \begin{subfigure}[b]{0.49\linewidth}
        \includegraphics[width=\linewidth]{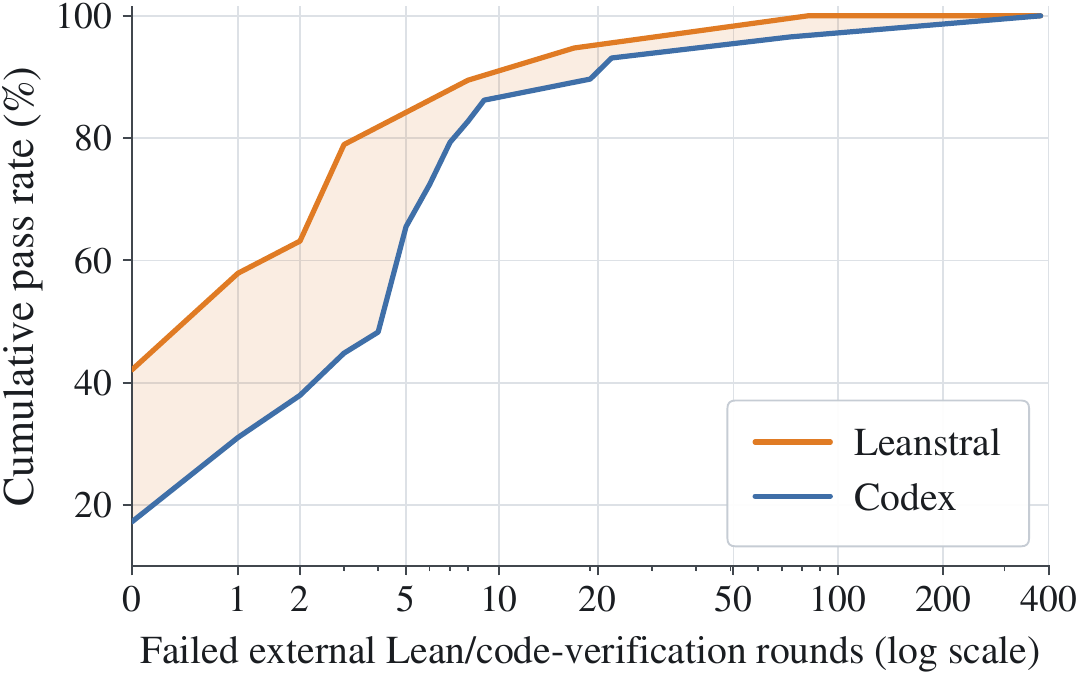}
        \caption{Failed external Lean verification rounds}
        \label{fig:failure-rounds}
    \end{subfigure}

    \vspace{0.4em}

    \begin{subfigure}[b]{0.49\linewidth}
        \includegraphics[width=\linewidth]{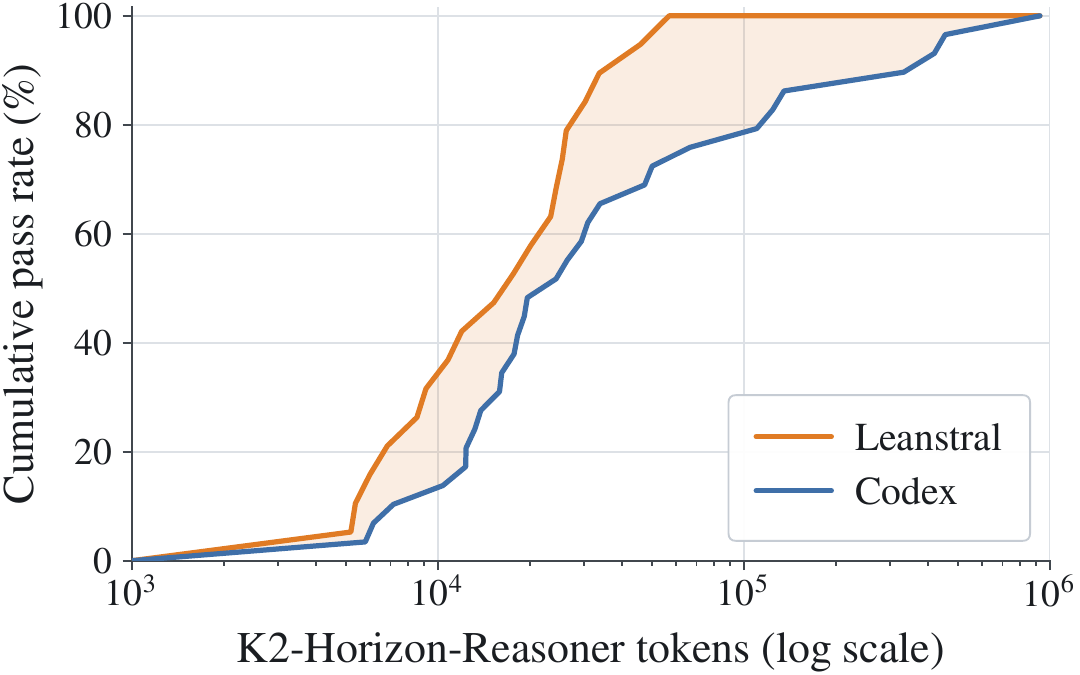}
        \caption{Reasoner (\textsc{K2-Horizon-7B}) tokens}
        \label{fig:reasoner-tokens}
    \end{subfigure}\hfill
    \begin{subfigure}[b]{0.49\linewidth}
        \includegraphics[width=\linewidth]{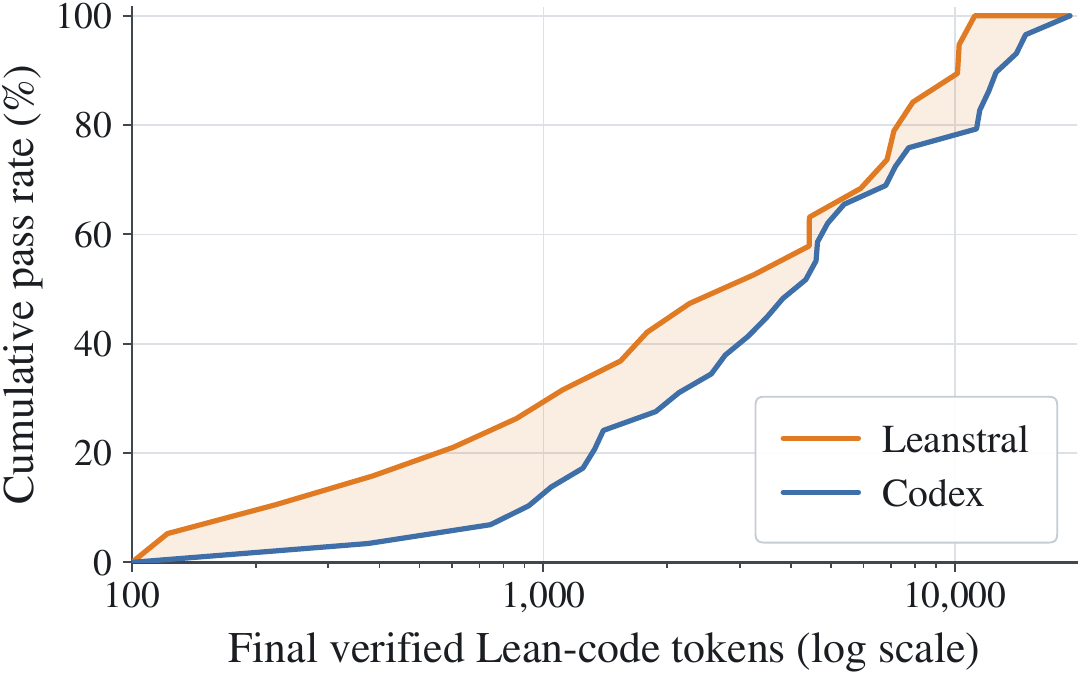}
        \caption{Final verified Lean-code length}
        \label{fig:lean-tokens}
    \end{subfigure}
\end{minipage}

\caption{Cumulative pass rate against four cost measures on AIME~2026
(Equation~\ref{eq:pass-curve}). Both backends reach $100\%$, so the curves
separate on budget, not on ceiling. Panels (b)--(d) are \emph{external}
measures, defined identically for both backends, and \textsc{Leanstral}
is left-shifted on each.}
\label{fig:codex-vs-leanstral}
\vspace{-2ex}
\end{figure}

Measured externally, \textsc{Leanstral} reaches the shared ceiling more cheaply: it solves $42\%$ of problems with no failed external verification round against $17\%$ for \textsc{Codex} (Figure~\ref{fig:failure-rounds}), reaches full coverage on approximately $6\times10^{4}$ reasoner tokens versus $9\times10^{5}$ (Figure~\ref{fig:reasoner-tokens}), and produces verified proofs roughly half as long, $1.1\times10^{4}$ against $2\times10^{4}$ tokens (Figure~\ref{fig:lean-tokens}).

This efficiency is bought with computation inside \textsc{Leanstral}'s agentic loop: its winning trajectories consume approximately $3\times10^{2}$ internal calls at median coverage and up to $3\times10^{3}$ at full coverage (Figure~\ref{fig:internal-calls}), and once these are counted the ordering reverses (Figure~\ref{fig:external-rounds}). The claim is therefore not that the open-weight prover is cheaper outright, but that at equal verified accuracy it moves cost from metered rounds into local computation one controls. Since \textsc{Codex} performs an unknown amount of internal work, its external measurements is only a lower-bound on its true compute. Attribution matters mainly in the remaining tail: an external failure reaches the router on $58\%$ of problems with \textsc{Leanstral} against $83\%$ with \textsc{Codex}.

\subsection{Robustness to Answer-Preserving Paraphrase}
\label{sec:paraphrase}

\begin{wraptable}[9]{r}{0.45\textwidth}
    \centering
    \vspace{-1.2em}
    \caption{Accuracy (\%) on original and paraphrased AIME~2026. $\Delta$ is the \% decrease.}
    \vspace{-0.5em}
    \label{tab:swapping}
    \small
    \setlength{\tabcolsep}{5pt}
\resizebox{.45\textwidth}{!}{%
    \begin{tabular}{lrrr}
    \toprule
    \textbf{Model} & \textbf{Original} & \textbf{Paraphrased} & $\bm{\Delta}$ \\
    \midrule
    K2-Horizon-375B  & 90.0  & 86.7  & -3.3 \\
    \rowcolor{gray!15}
    \quad + \method  & 100.0 & 100.0 & 0.0 \\
    \midrule
    K2-Horizon-7B    & 80.0  & 70.0  & -10.0 \\
    \rowcolor{gray!15}
    \quad + \method  & 100.0 & 100.0 & 0.0 \\
    \bottomrule
    \end{tabular}}
    \vspace{-0.2em}
\end{wraptable}

To control for possible data leakage, we use DeepSeek-V4-Flash to paraphrase every AIME~2026 problem while preserving its quantities, hypotheses, goal and reference answer, check each paraphrase before use using {\sc Codex}, and measure the robustness gap $\Delta$ between original and paraphrased accuracy.
Prior work~\citep{DBLP:conf/iclr/MirzadehASTBF25} shows that standalone reasoners are sensitive to such answer-preserving changes, 
and Table~\ref{tab:swapping} reproduces this: paraphrasing costs \textsc{K2-Horizon-375B} $3.3$ points and \textsc{K2-Horizon-7B} $10.0$ points, whereas both retain $100\%$ with \method ($\Delta=0$). 
Perturbing the numerical quantities, so that the reference answer itself changes, would be the stronger test; we therefore read this as necessary rather than sufficient evidence against contamination.

\subsection{Ablation Study}
\label{sec:ablations}

We ablate the two mechanisms that distinguish \method from a generic retry loop: the statement judge $J_{\mathrm{s}}$, which decides what is proved, and the error judge $J_{\mathrm{e}}$, which decides how a failed attempt is retried. Both rerun the same harness with one component disabled, and the rest as in \S\ref{sec:exp-setup}.

\paragraph{Statement adjudication.}
\begin{wraptable}{r}{0.50\textwidth}
    \centering
    \vspace{-1.2em}
    \caption{Effect of statement adjudication with Goedel formaliser on AIME~2026.}
    \label{tab:ablation-judge}
    \small
    \setlength{\tabcolsep}{4pt}
    \begin{tabular}{lrrr}
        \toprule
        \textbf{Setting} & $\bm{\mathrm{Ver}}\uparrow$ & $\bm{\mathrm{VerCor}}\uparrow$
        & $\bm{\mathrm{FCR}}\downarrow$ \\
        \midrule
        \rowcolor{gray!15}
        With $J_{\mathrm{s}}$
        & $63.3\%$ & $63.3\%$ & $0.0\%$ \\
        Without $J_{\mathrm{s}}$
        & $36.7\%$ & $20.0\%$ & $45.5\%$ \\
        \bottomrule
    \end{tabular}
    \vspace{-0.8em}
\end{wraptable}
A certificate is meaningful only when the proved statement preserves the original problem, so we separate two outcomes. Let $\mathrm{Ver}$ be the fraction of problems for which the pipeline returns a verified proof and $\mathrm{VerCor}$ the fraction for which it does so \emph{and} the certified answer matches the reference. The false certification rate $\mathrm{FCR}=1-\mathrm{VerCor}/\mathrm{Ver}$ is then the fraction of returned certificates whose answer is nonetheless wrong. Since the verifier $V$ is sound, $\mathrm{FCR}>0$ is possible only through an unfaithful formalisation, which makes it a direct measurement of the residual gap identified in \S\ref{sec:scope}. We disable $J_{\mathrm{s}}$ by accepting the first elaborated statement unconditionally.
Table~\ref{tab:ablation-judge} reports the result.

Without the judge $\mathrm{Ver}$ falls from $63.3\%$ to $36.7\%$ even though every first-sampled formalisation is now accepted. The judge, therefore, acts as an additional filter on the formal statement beyond semantically aligning it with the informal one. The effect on correctness is larger still: $\mathrm{VerCor}$ falls $43.3$ points to $20.0\%$ and $\mathrm{FCR}$ rises to $45.5\%$, so nearly half the surviving certificates prove something other than the problem asked, but not necessarily contradictory (Appendix~\ref{app:formal-statement-errors}).

\paragraph{Resampling versus diagnostic-conditioned self-correction.} We next replace diagnostic-conditioned self-correction with independent resampling, which discards compiler feedback and draws a fresh proof each attempt. This baseline is stronger than it may appear: for a problem with single-shot success probability $p>0$, after $n$ parallelisable attempts of using $V$ as a sound selector, the probability is increased to $1-(1-p)^{n} \ge 1-e^{-np}$. On AIME~2026 resampling with \textsc{K2-Horizon-7B} solves $29/30$ ($96.7\%$) against $30/30$ for self-correction; in the single failure the initial answer is wrong, so the statement is unprovable and further proof samples cannot recover it.
 
The gap widens on IMO~2026, where, despite $8$--$16$ independent generations per problem, resampling verifies only $1/6$ ($16.7\%$) while self-correction verifies all six, so $p$ is effectively zero under our budget for most IMO problems. Conditioning on a diagnostic moves probability mass onto proofs the unconditioned distribution does not reach.
The benefit in error attribution (\textsc{Math} versus \textsc{Syntax}) is routing the next attempt to the right component based on the previous attempt.

\section{Related Work}

\paragraph{LLMs for Mathematical Reasoning.} LLMs have achieved strong performance in competition mathematics through CoT reasoning~\citep{wei2022chain} and test-time scaling~\citep{albalak2025bigmathlargescalehighqualitymath,DBLP:journals/corr/abs-2402-03300,zhang2025survey}. As plausible reasoning traces may still lead to incorrect solution, prior work applies LLM judges~\citep{yosef2026rethinking}, process reward models~\citep{liu2024skywork}, and related scoring or selection methods~\citep{ospanov2026hermes,leang-etal-2026-picsar}. These signals remain learned proxies that may share the models' failure modes~\citep{leang-etal-2026-picsar,leang-etal-2025-comat}. \method instead uses Lean as a deterministic verifier of the generated formal proof.

\paragraph{Theorem Proving and Autoformalisation.} LLM-based theorem proving has advanced rapidly, with specialised proving models such as Pythagoras-Prover~\citep{leang2026pythagoras}, Goedel-Prover~\citep{lin2025goedelproverv2scalingformaltheorem}, and others~\citep{ma2026oprover,wang2026longcat}, together with agentic closed-source models~\citep{liu2026numinaleanagent,xu2026aoa}, have rapidly advanced LLM theorem proving, albeit often through substantial test-time scaling~\citep{varambally2026hilbert,klingner2026evaluation,chen2025seed,lin2025goedelproverv2scalingformaltheorem}. However, current benchmarking assume an accurate formal statement as the input. Producing faithful formal statements remains challenging: an incorrect formalisation may admit a valid proof without certifying the original problem~\citep{cornish2026faithformbench}.

\paragraph{Formal--Informal Distillation.} Previous work has explored connections between informal reasoning and formal proof~\citep{varambally2026hilbert, quan2025faithful}, but often relying solely on closed-source models, and underperforms state-of-the-art mathematical reasoners~\citep{DBLP:conf/acl/00010YXXCWSL025, DBLP:conf/iclr/ZhouSLSWW24, DBLP:conf/emnlp/YaoWZ25, zhang2025autoformalization}. Carefully curated formal--informal pipelines can improve reliability~\citep{DBLP:journals/corr/abs-2602-20770}, while Theorem Prover as a Judge uses GPT-4o and formal verification to validate informal traces during synthetic-data generation~\citep{leang-etal-2025-theorem}. \citet{tsoukalas2026advancing}, on the other hand, applies formal-informal distillation on solving mathematical research problems, but rely heavily on closed-source models with large parameter scale. Often, current methods assume that a formal goal is already available~\citep{DBLP:conf/emnlp/YaoWZ25,varambally2026hilbert,DBLP:journals/corr/abs-2605-28365}. 
These systems use formal verification as a selection or certification signal, rather than for error attribution.

\section{Conclusion}

We introduced \method, a training-free agentic pipeline that couples informal mathematical reasoning with formal verification. Given only a natural-language problem, it produces a reasoning chain and candidate answer, formalises them as a Lean~4 statement, and generates a checked proof, with a statement judge deciding whether the formalisation preserves the original problem and an error judge routing failed attempts to re-derivation or local repair. \method verifies every problem on AIME~2025, AIME~2026 and HMMT February~2026 with four reasoners, and with \textsc{K2-Horizon-7B} solves all six IMO~2026 problems. Our ablation experimental results show that adjudication is what keeps a certificate meaningful and that diagnostic-conditioned correction reaches proofs independent resampling does not. Because the pipeline writes the statement it proves, verification guarantees correctness only relative to the generated formalisation, not unconditionally with respect to the original problem. Closing that gap is a question of better autoformalisation rather than better proof search, which we regard as the main remaining obstacle to Lean-guided natural language reasoning.

\bibliography{iclr2027_conference}

\begin{thebibliography}{52}
\providecommand{\natexlab}[1]{#1}
\providecommand{\url}[1]{\texttt{#1}}
\expandafter\ifx\csname urlstyle\endcsname\relax
  \providecommand{\doi}[1]{doi: #1}\else
  \providecommand{\doi}{doi: \begingroup \urlstyle{rm}\Url}\fi

\bibitem[Albalak et~al.(2025)Albalak, Phung, Lile, Rafailov, Gandhi, Castricato, Singh, Blagden, Xiang, Mahan, and Haber]{albalak2025bigmathlargescalehighqualitymath}
Alon Albalak, Duy Phung, Nathan Lile, Rafael Rafailov, Kanishk Gandhi, Louis Castricato, Anikait Singh, Chase Blagden, Violet Xiang, Dakota Mahan, and Nick Haber.
\newblock Big-math: A large-scale, high-quality math dataset for reinforcement learning in language models, 2025.
\newblock URL \url{https://arxiv.org/abs/2502.17387}.

\bibitem[{Anthropic}(2026)]{anthropic2026claude46systemcard}
{Anthropic}.
\newblock {S}ystem {C}ard: {C}laude {O}pus 4.6.
\newblock \url{https://anthropic.com/claude-opus-4-6-system-card}, 2026.

\bibitem[Bourigault et~al.(2026)Bourigault, Ji, Zimmer, Tutunov, and Ammar]{DBLP:journals/corr/abs-2605-28365}
Pauline Bourigault, Xiaotong Ji, Matthieu Zimmer, Rasul Tutunov, and Haitham~Bou Ammar.
\newblock Risk-controlled lean-as-judge for natural-language mathematical reasoning, 2026.
\newblock URL \url{https://arxiv.org/abs/2605.28365}.

\bibitem[Cesa-Bianchi \& Lugosi(2006)Cesa-Bianchi and Lugosi]{Cesa-Bianchi_Lugosi_2006}
Nicolo Cesa-Bianchi and Gabor Lugosi.
\newblock \emph{Prediction, Learning, and Games}.
\newblock Cambridge University Press, 2006.

\bibitem[Chen et~al.(2025)Chen, Chen, Du, Hu, Jiang, Jie, Jin, Jin, Li, Shi, et~al.]{chen2025seed}
Jiangjie Chen, Wenxiang Chen, Jiacheng Du, Jinyi Hu, Zhicheng Jiang, Allan Jie, Xiaoran Jin, Xing Jin, Chenggang Li, Wenlei Shi, et~al.
\newblock Seed-prover 1.5: Mastering undergraduate-level theorem proving via learning from experience.
\newblock \emph{arXiv preprint arXiv:2512.17260}, 2025.

\bibitem[Chen \& Jiang(2026)Chen and Jiang]{DBLP:journals/corr/abs-2606-10806}
Xiaoyang Chen and Xiang Jiang.
\newblock Moonshine: An autonomous mathematical research agent centered on conjecture generation, 2026.
\newblock URL \url{https://arxiv.org/abs/2606.10806}.

\bibitem[Coquand \& Huet(1988)Coquand and Huet]{CoquandHuet1988CoC}
Thierry Coquand and G{\'e}rard Huet.
\newblock {The Calculus of Constructions}.
\newblock \emph{Information and Computation}, 76\penalty0 (2--3):\penalty0 95--120, 1988.
\newblock URL \url{https://doi.org/10.1016/0890-5401(88)90005-3}.

\bibitem[Cornish et~al.(2026)Cornish, Ghinassi, Yeh, Liu, Xu, Yin, Wagner, Li, Teh, and Ong]{cornish2026faithformbench}
Rob Cornish, Iacopo Ghinassi, Po-Hung Yeh, Shuqi Liu, Qiyuan Xu, Haoxuan Yin, Dominik Wagner, Wenda Li, Yee~Whye Teh, and Luke Ong.
\newblock Faithformbench: Benchmarking faithfulness of mathematical chain-of-thought autoformalisation.
\newblock \emph{arXiv preprint arXiv:2608.10916}, 2026.

\bibitem[Guan et~al.(2025)Guan, Zhang, Liu, Shang, Sun, Zhu, Yang, and Yang]{pmlr-v267-guan25f}
Xinyu Guan, Li~Lyna Zhang, Yifei Liu, Ning Shang, Youran Sun, Yi~Zhu, Fan Yang, and Mao Yang.
\newblock r{S}tar-math: Small {LLM}s can master math reasoning with self-evolved deep thinking.
\newblock In Aarti Singh, Maryam Fazel, Daniel Hsu, Simon Lacoste-Julien, Felix Berkenkamp, Tegan Maharaj, Kiri Wagstaff, and Jerry Zhu (eds.), \emph{Proceedings of the 42nd International Conference on Machine Learning}, volume 267 of \emph{Proceedings of Machine Learning Research}, pp.\  20640--20661. PMLR, 13--19 Jul 2025.
\newblock URL \url{https://proceedings.mlr.press/v267/guan25f.html}.

\bibitem[Huang \& Yang(2025)Huang and Yang]{huang2025gemini}
Yichen Huang and Lin~F. Yang.
\newblock Winning gold at imo 2025 with a model-agnostic verification-and-refinement pipeline, 2025.
\newblock URL \url{https://arxiv.org/abs/2507.15855}.

\bibitem[Jiang et~al.(2026)Jiang, Liang, Zhang, Wan, Li, Deng, Taylor, Baker, Raghavan, Zhang, Wu, Bertozzi, Chang, Meka, Sottile, Peng, Sahai, Tao, and Wang]{DBLP:journals/corr/abs-2607-07779}
Eric Jiang, Xiao Liang, Yikai Zhang, Yingjia Wan, Mengting Li, Haikang Deng, Alexander~K. Taylor, Justin Baker, Rushil Raghavan, Junyi Zhang, Ying~Nian Wu, Andrea~L. Bertozzi, Kai-Wei Chang, Raghu Meka, Matthew Sottile, Nanyun Peng, Amit Sahai, Terence Tao, and Wei Wang.
\newblock From solvers to research: Large language model-driven formal mathematics at the research frontier, 2026.
\newblock URL \url{https://arxiv.org/abs/2607.07779}.

\bibitem[Kapoor et~al.(2025)Kapoor, Stroebl, Siegel, Nadgir, and Narayanan]{kapoor2025ai}
Sayash Kapoor, Benedikt Stroebl, Zachary~S Siegel, Nitya Nadgir, and Arvind Narayanan.
\newblock {AI} agents that matter.
\newblock \emph{Transactions on Machine Learning Research}, 2025.
\newblock ISSN 2835-8856.
\newblock URL \url{https://openreview.net/forum?id=Zy4uFzMviZ}.

\bibitem[Klingner et~al.(2026)Klingner, Bladek, Crawford, Chen, Fu, Nair, Alper, Inchiostro, and Ilin]{klingner2026evaluation}
Tyson Klingner, Drew Bladek, Escher Crawford, Bohao Chen, Ariel Fu, Kaira Nair, Jarod Alper, Giovanni Inchiostro, and Vasily Ilin.
\newblock Evaluation of {LLMs} for mathematical formalization in {Lean}, 2026.
\newblock URL \url{https://arxiv.org/abs/2606.05632}.

\bibitem[Leang et~al.(2025{\natexlab{a}})Leang, Gema, and Cohen]{leang-etal-2025-comat}
Joshua Ong~Jun Leang, Aryo~Pradipta Gema, and Shay~B Cohen.
\newblock {C}o{MAT}: {C}hain of {M}athematically {A}nnotated {T}hought {I}mproves {M}athematical {R}easoning.
\newblock In Christos Christodoulopoulos, Tanmoy Chakraborty, Carolyn Rose, and Violet Peng (eds.), \emph{Proceedings of the 2025 Conference on Empirical Methods in Natural Language Processing}, pp.\  20245--20274, Suzhou, China, November 2025{\natexlab{a}}. Association for Computational Linguistics.
\newblock ISBN 979-8-89176-332-6.
\newblock \doi{10.18653/v1/2025.emnlp-main.1024}.
\newblock URL \url{https://aclanthology.org/2025.emnlp-main.1024/}.

\bibitem[Leang et~al.(2025{\natexlab{b}})Leang, Hong, Li, and Cohen]{leang-etal-2025-theorem}
Joshua Ong~Jun Leang, Giwon Hong, Wenda Li, and Shay~B Cohen.
\newblock {T}heorem {P}rover as a {J}udge for {S}ynthetic {D}ata {G}eneration.
\newblock In Wanxiang Che, Joyce Nabende, Ekaterina Shutova, and Mohammad~Taher Pilehvar (eds.), \emph{Proceedings of the 63rd Annual Meeting of the Association for Computational Linguistics (Volume 1: Long Papers)}, pp.\  29941--29977, Vienna, Austria, July 2025{\natexlab{b}}. Association for Computational Linguistics.
\newblock ISBN 979-8-89176-251-0.
\newblock \doi{10.18653/v1/2025.acl-long.1448}.
\newblock URL \url{https://aclanthology.org/2025.acl-long.1448/}.

\bibitem[Leang et~al.(2026{\natexlab{a}})Leang, Zhao, Gema, Yang, Kwan, He, Li, Minervini, Giunchiglia, and Cohen]{leang-etal-2026-picsar}
Joshua Ong~Jun Leang, Zheng Zhao, Aryo~Pradipta Gema, Sohee Yang, Wai-Chung Kwan, Xuanli He, Wenda Li, Pasquale Minervini, Eleonora Giunchiglia, and Shay~B Cohen.
\newblock {P}i{CSAR}: Probabilistic confidence selection and ranking for reasoning chains.
\newblock In Maria Liakata, Viviane~P. Moreira, Jiajun Zhang, and David Jurgens (eds.), \emph{Findings of the {A}ssociation for {C}omputational {L}inguistics: {ACL} 2026}, pp.\  31511--31544, San Diego, California, United States, July 2026{\natexlab{a}}. Association for Computational Linguistics.
\newblock ISBN 979-8-89176-395-1.
\newblock \doi{10.18653/v1/2026.findings-acl.1577}.
\newblock URL \url{https://aclanthology.org/2026.findings-acl.1577/}.

\bibitem[Leang et~al.(2026{\natexlab{b}})Leang, Zhao, Stoian, Xu, Li, Li, Cohen, and Giunchiglia]{leang2026pythagoras}
Joshua Ong~Jun Leang, Zheng Zhao, Mihaela~Cătălina Stoian, Qiyuan Xu, Haonan Li, Wenda Li, Shay~B. Cohen, and Eleonora Giunchiglia.
\newblock Pythagoras-prover: Advancing efficient formal proving via augmented lean formalisation, 2026{\natexlab{b}}.
\newblock URL \url{https://arxiv.org/abs/2606.12594}.

\bibitem[Li et~al.(2026)Li, Zhang, Li, Ji, Zeng, Cheng, Zhu, Wang, Wang, Xiao, and He]{li2026rethinking}
Zhuochun Li, Yong Zhang, Ming Li, Yuelyu Ji, Yiming Zeng, Ning Cheng, Yun Zhu, Yanmeng Wang, Shaojun Wang, Jing Xiao, and Daqing He.
\newblock Rethinking {LLM}-as-a-judge: Representation-as-a-judge with small language models via semantic capacity asymmetry.
\newblock In \emph{The Fourteenth International Conference on Learning Representations}, 2026.
\newblock URL \url{https://openreview.net/forum?id=VAISvCsrvG}.

\bibitem[Lin et~al.(2026)Lin, Tang, Lyu, Yang, Chung, Zhao, Jiang, Geng, Ge, Sun, Wu, Gesi, Lu, Acuna, Yang, Lin, Choi, Chen, Arora, and Jin]{lin2025goedelproverv2scalingformaltheorem}
Yong Lin, Shange Tang, Bohan Lyu, Ziran Yang, Jui-Hui Chung, Haoyu Zhao, Lai Jiang, Yihan Geng, Jiawei Ge, Jingruo Sun, Jiayun Wu, Jiri Gesi, Ximing Lu, David Acuna, Kaiyu Yang, Hongzhou Lin, Yejin Choi, Danqi Chen, Sanjeev Arora, and Chi Jin.
\newblock {G}oedel-{P}rover-{V}2: {S}caling {F}ormal {T}heorem {P}roving with {S}caffolded {D}ata {S}ynthesis and {S}elf-{C}orrection.
\newblock In \emph{The Fourteenth International Conference on Learning Representations}, 2026.
\newblock URL \url{https://openreview.net/forum?id=j4C0nALrgK}.

\bibitem[Littlestone(1987)]{4568257}
Nick Littlestone.
\newblock Learning quickly when irrelevant attributes abound: A new linear-threshold algorithm.
\newblock In \emph{28th Annual Symposium on Foundations of Computer Science (sfcs 1987)}, pp.\  68--77, 1987.
\newblock \doi{10.1109/SFCS.1987.37}.

\bibitem[Liu et~al.(2025)Liu, Yuan, Yin, Xu, Xu, Chen, Wang, Shang, Liu, and Zhang]{DBLP:conf/acl/00010YXXCWSL025}
Chengwu Liu, Ye~Yuan, Yichun Yin, Yan Xu, Xin Xu, Zaoyu Chen, Yasheng Wang, Lifeng Shang, Qun Liu, and Ming Zhang.
\newblock Safe: Enhancing mathematical reasoning in large language models via retrospective step-aware formal verification.
\newblock In Wanxiang Che, Joyce Nabende, Ekaterina Shutova, and Mohammad~Taher Pilehvar (eds.), \emph{Proceedings of the 63rd Annual Meeting of the Association for Computational Linguistics (Volume 1: Long Papers)}, pp.\  12171--12186, Vienna, Austria, July 2025. Association for Computational Linguistics.
\newblock ISBN 979-8-89176-251-0.
\newblock \doi{10.18653/v1/2025.acl-long.594}.
\newblock URL \url{https://aclanthology.org/2025.acl-long.594/}.

\bibitem[Liu et~al.(2024)Liu, Zeng, Liu, Yan, He, Wang, Yan, Liu, and Zhou]{liu2024skywork}
Chris~Yuhao Liu, Liang Zeng, Jiacai Liu, Rui Yan, Jujie He, Chaojie Wang, Shuicheng Yan, Yang Liu, and Yahui Zhou.
\newblock Skywork-reward: Bag of tricks for reward modeling in {LLMs}, 2024.
\newblock URL \url{https://arxiv.org/abs/2410.18451}.

\bibitem[Liu et~al.(2026)Liu, Zhou, Zhu, Santos, He, jiawei liu, Xie, Zhao, Wang, Zhi, LI, and Li]{liu2026numinaleanagent}
Junqi Liu, Zihao Zhou, Zekai Zhu, Marco~Dos Santos, Weikun He, jiawei liu, Yunzhou Xie, Junqiao Zhao, Qiufeng Wang, Lihong Zhi, Jia LI, and Wenda Li.
\newblock Numina-lean-agent: An open and general agentic reasoning system for formal mathematics.
\newblock In \emph{Forty-third International Conference on Machine Learning}, 2026.
\newblock URL \url{https://openreview.net/forum?id=0bTEd4LpQr}.

\bibitem[Lyu et~al.(2023)Lyu, Havaldar, Stein, Zhang, Rao, Wong, Apidianaki, and Callison-Burch]{lyu-etal-2023-faithful}
Qing Lyu, Shreya Havaldar, Adam Stein, Li~Zhang, Delip Rao, Eric Wong, Marianna Apidianaki, and Chris Callison-Burch.
\newblock {F}aithful {C}hain-of-{T}hought {R}easoning.
\newblock In Jong~C. Park, Yuki Arase, Baotian Hu, Wei Lu, Derry Wijaya, Ayu Purwarianti, and Adila~Alfa Krisnadhi (eds.), \emph{Proceedings of the 13th International Joint Conference on Natural Language Processing and the 3rd Conference of the Asia-Pacific Chapter of the Association for Computational Linguistics (Volume 1: Long Papers)}, pp.\  305--329, Nusa Dua, Bali, November 2023. Association for Computational Linguistics.
\newblock \doi{10.18653/v1/2023.ijcnlp-main.20}.
\newblock URL \url{https://aclanthology.org/2023.ijcnlp-main.20/}.

\bibitem[Ma et~al.(2026)Ma, Ma, Guo, Shi, Zhao, Shi, Zhang, Cheung, Liu, and Wang]{ma2026oprover}
David Ma, Kaijing Ma, Shawn Guo, Yunfeng Shi, Enduo Zhao, Jiajun Shi, Zhaoxiang Zhang, Gavin Cheung, Jiaheng Liu, and Zili Wang.
\newblock {OProver}: A unified framework for agentic formal theorem proving, 2026.
\newblock URL \url{https://arxiv.org/abs/2605.17283}.

\bibitem[Mirzadeh et~al.(2025)Mirzadeh, Alizadeh, Shahrokhi, Tuzel, Bengio, and Farajtabar]{DBLP:conf/iclr/MirzadehASTBF25}
Iman Mirzadeh, Keivan Alizadeh, Hooman Shahrokhi, Oncel Tuzel, Samy Bengio, and Mehrdad Farajtabar.
\newblock {GSM}-{S}ymbolic: {U}nderstanding the {L}imitations of {M}athematical {R}easoning in {L}arge {L}anguage {M}odels.
\newblock In \emph{The Thirteenth International Conference on Learning Representations, {ICLR} 2025, Singapore, April 24-28, 2025}. OpenReview.net, 2025.
\newblock URL \url{https://openreview.net/forum?id=AjXkRZIvjB}.

\bibitem[Moura \& Ullrich(2021)Moura and Ullrich]{10.1007/978-3-030-79876-5_37}
Leonardo~de Moura and Sebastian Ullrich.
\newblock The lean 4 theorem prover and programming language.
\newblock In \emph{Automated Deduction – CADE 28: 28th International Conference on Automated Deduction, Virtual Event, July 12–15, 2021, Proceedings}, pp.\  625–635, Berlin, Heidelberg, 2021. Springer-Verlag.
\newblock ISBN 978-3-030-79875-8.
\newblock \doi{10.1007/978-3-030-79876-5_37}.
\newblock URL \url{https://doi.org/10.1007/978-3-030-79876-5_37}.

\bibitem[Muennighoff et~al.(2025)Muennighoff, Yang, Shi, Li, Fei-Fei, Hajishirzi, Zettlemoyer, Liang, Cand{\`e}s, and Hashimoto]{muennighoff2025s1}
Niklas Muennighoff, Zitong Yang, Weijia Shi, Xiang~Lisa Li, Li~Fei-Fei, Hannaneh Hajishirzi, Luke Zettlemoyer, Percy Liang, Emmanuel Cand{\`e}s, and Tatsunori Hashimoto.
\newblock s1: {S}imple test-time scaling.
\newblock In Christos Christodoulopoulos, Tanmoy Chakraborty, Carolyn Rose, and Violet Peng (eds.), \emph{Proceedings of the 2025 Conference on Empirical Methods in Natural Language Processing}, pp.\  20275--20321, Suzhou, China, November 2025. Association for Computational Linguistics.
\newblock ISBN 979-8-89176-332-6.
\newblock \doi{10.18653/v1/2025.emnlp-main.1025}.
\newblock URL \url{https://aclanthology.org/2025.emnlp-main.1025/}.

\bibitem[{OpenAI}(2026)]{DBLP:journals/corr/abs-2601-03267}
{OpenAI}.
\newblock {O}pen{AI} {GPT-5} {S}ystem {C}ard, 2026.
\newblock URL \url{https://arxiv.org/abs/2601.03267}.

\bibitem[Ospanov et~al.(2026)Ospanov, Feng, Sun, Bai, XIN, and Farnia]{ospanov2026hermes}
Azim Ospanov, Zijin Feng, Jiacheng Sun, Haoli Bai, SHEN XIN, and Farzan Farnia.
\newblock {HERMES}: Towards efficient and verifiable mathematical reasoning in {LLM}s.
\newblock In \emph{Forty-third International Conference on Machine Learning}, 2026.
\newblock URL \url{https://openreview.net/forum?id=w7BZcUc0fJ}.

\bibitem[Paulson(1994)]{DBLP:books/sp/Paulson94}
Lawrence~C. Paulson.
\newblock \emph{{I}sabelle - {A} {G}eneric {T}heorem {P}rover (with a contribution by {T}. {N}ipkow)}, volume 828 of \emph{Lecture Notes in Computer Science}.
\newblock Springer, 1994.
\newblock ISBN 3-540-58244-4.
\newblock \doi{10.1007/BFB0030541}.
\newblock URL \url{https://doi.org/10.1007/BFb0030541}.

\bibitem[Quan et~al.(2025)Quan, Valentino, Dennis, and Freitas]{quan2025faithful}
Xin Quan, Marco Valentino, Louise~A. Dennis, and Andre Freitas.
\newblock Faithful and robust {LLM}-driven theorem proving for {NLI} explanations.
\newblock In Wanxiang Che, Joyce Nabende, Ekaterina Shutova, and Mohammad~Taher Pilehvar (eds.), \emph{Proceedings of the 63rd Annual Meeting of the Association for Computational Linguistics (Volume 1: Long Papers)}, pp.\  17734--17755, Vienna, Austria, July 2025. Association for Computational Linguistics.
\newblock ISBN 979-8-89176-251-0.
\newblock \doi{10.18653/v1/2025.acl-long.867}.
\newblock URL \url{https://aclanthology.org/2025.acl-long.867/}.

\bibitem[{Qwen Team}(2026)]{qwen3.6-27b}
{Qwen Team}.
\newblock {Qwen3.6-27B}: {F}lagship-{L}evel {C}oding in a {27B} {D}ense {M}odel, April 2026.
\newblock URL \url{https://qwen.ai/blog?id=qwen3.6-27b}.

\bibitem[Sahoo et~al.(2024)Sahoo, Arriola, Schiff, Gokaslan, Marroquin, Chiu, Rush, and Kuleshov]{sahoo2024simple}
Subham~Sekhar Sahoo, Marianne Arriola, Yair Schiff, Aaron Gokaslan, Edgar Marroquin, Justin~T Chiu, Alexander Rush, and Volodymyr Kuleshov.
\newblock {S}imple and {E}ffective {M}asked {D}iffusion {L}anguage {M}odels.
\newblock In A.~Globerson, L.~Mackey, D.~Belgrave, A.~Fan, U.~Paquet, J.~Tomczak, and C.~Zhang (eds.), \emph{Advances in Neural Information Processing Systems}, volume~37, pp.\  130136--130184. Curran Associates, Inc., 2024.
\newblock \doi{10.52202/079017-4135}.
\newblock URL \url{https://proceedings.neurips.cc/paper_files/paper/2024/file/eb0b13cc515724ab8015bc978fdde0ad-Paper-Conference.pdf}.

\bibitem[Sazonova et~al.(2026)Sazonova, Shmelkin, Kikot, and Motolygin]{DBLP:journals/corr/abs-2602-20770}
Varvara Sazonova, Dmitri Shmelkin, Stanislav Kikot, and Vasily Motolygin.
\newblock Pipeline for verifying {LLM}-generated mathematical solutions, 2026.
\newblock URL \url{https://arxiv.org/abs/2602.20770}.

\bibitem[Shao et~al.(2024)Shao, Wang, Zhu, Xu, Song, Bi, Zhang, Zhang, Li, Wu, and Guo]{DBLP:journals/corr/abs-2402-03300}
Zhihong Shao, Peiyi Wang, Qihao Zhu, Runxin Xu, Junxiao Song, Xiao Bi, Haowei Zhang, Mingchuan Zhang, Y.~K. Li, Y.~Wu, and Daya Guo.
\newblock {DeepSeekMath}: Pushing the limits of mathematical reasoning in open language models, 2024.
\newblock URL \url{https://arxiv.org/abs/2402.03300}.

\bibitem[Snell et~al.(2025)Snell, Lee, Xu, and Kumar]{snell2025scaling}
Charlie~Victor Snell, Jaehoon Lee, Kelvin Xu, and Aviral Kumar.
\newblock Scaling {LLM} test-time compute optimally can be more effective than scaling parameters for reasoning.
\newblock In \emph{The Thirteenth International Conference on Learning Representations}, 2025.
\newblock URL \url{https://openreview.net/forum?id=4FWAwZtd2n}.

\bibitem[Team(2026)]{DBLP:journals/corr/abs-2607-24653}
Kimi Team.
\newblock Kimi {K3:} open frontier intelligence, 2026.
\newblock URL \url{https://arxiv.org/abs/2607.24653}.

\bibitem[Tsoukalas et~al.(2024)Tsoukalas, Lee, Jennings, Xin, Ding, Jennings, Thakur, and Chaudhuri]{tsoukalas2024putnambenchevaluatingneuraltheoremprovers}
George Tsoukalas, Jasper Lee, John Jennings, Jimmy Xin, Michelle Ding, Michael Jennings, Amitayush Thakur, and Swarat Chaudhuri.
\newblock {P}utnam{B}ench: {E}valuating {N}eural {T}heorem-{P}rovers on the {P}utnam {M}athematical {C}ompetition.
\newblock In A.~Globerson, L.~Mackey, D.~Belgrave, A.~Fan, U.~Paquet, J.~Tomczak, and C.~Zhang (eds.), \emph{Advances in Neural Information Processing Systems}, volume~37, pp.\  11545--11569. Curran Associates, Inc., 2024.
\newblock \doi{10.52202/079017-0368}.
\newblock URL \url{https://proceedings.neurips.cc/paper_files/paper/2024/file/1582eaf9e0cf349e1e5a6ee453100aa1-Paper-Datasets_and_Benchmarks_Track.pdf}.

\bibitem[Tsoukalas et~al.(2026)Tsoukalas, Kovsharov, Shirobokov, Surina, Firsching, B{\'e}rczi, Ruiz, Suggala, Wagner, Wieser, et~al.]{tsoukalas2026advancing}
George Tsoukalas, Anton Kovsharov, Sergey Shirobokov, Anja Surina, Moritz Firsching, Gergely B{\'e}rczi, Francisco~JR Ruiz, Arun Suggala, Adam~Zsolt Wagner, Eric Wieser, et~al.
\newblock Advancing mathematics research with ai-driven formal proof search.
\newblock \emph{arXiv preprint arXiv:2605.22763}, 2026.

\bibitem[Varambally et~al.(2026)Varambally, Voice, Sun, Chen, Yu, and Ye]{varambally2026hilbert}
Sumanth Varambally, Thomas Voice, Yanchao Sun, Zhifeng Chen, Rose Yu, and Ke~Ye.
\newblock Hilbert: Recursively building formal proofs with informal reasoning.
\newblock In \emph{The Fourteenth International Conference on Learning Representations}, 2026.
\newblock URL \url{https://openreview.net/forum?id=GN8OdkTo3B}.

\bibitem[Wang et~al.(2026)Wang, Zhang, Guo, Guo, Li, Zhang, Peng, Wang, Zhao, Shi, Wang, et~al.]{wang2026longcat}
Jianing Wang, Jianfei Zhang, Qi~Guo, Linsen Guo, Rumei Li, Chao Zhang, Chong Peng, Cunguang Wang, Dengchang Zhao, Jiarong Shi, Jingang Wang, et~al.
\newblock Longcat-flash-prover: Advancing native formal reasoning via agentic tool-integrated reinforcement learning, 2026.
\newblock URL \url{https://arxiv.org/abs/2603.21065}.

\bibitem[Wei et~al.(2022)Wei, Wang, Schuurmans, Bosma, ichter, Xia, Chi, Le, and Zhou]{wei2022chain}
Jason Wei, Xuezhi Wang, Dale Schuurmans, Maarten Bosma, brian ichter, Fei Xia, Ed~Chi, Quoc~V Le, and Denny Zhou.
\newblock {C}hain-of-{T}hought {P}rompting {E}licits {R}easoning in {L}arge {L}anguage {M}odels.
\newblock In S.~Koyejo, S.~Mohamed, A.~Agarwal, D.~Belgrave, K.~Cho, and A.~Oh (eds.), \emph{Advances in Neural Information Processing Systems}, volume~35, pp.\  24824--24837. Curran Associates, Inc., 2022.
\newblock URL \url{https://proceedings.neurips.cc/paper_files/paper/2022/file/9d5609613524ecf4f15af0f7b31abca4-Paper-Conference.pdf}.

\bibitem[Xin et~al.(2024)Xin, Guo, Shao, Ren, Zhu, Liu, Ruan, Li, and Liang]{DBLP:journals/corr/abs-2405-14333}
Huajian Xin, Daya Guo, Zhihong Shao, Zhizhou Ren, Qihao Zhu, Bo~Liu, Chong Ruan, Wenda Li, and Xiaodan Liang.
\newblock {D}eep{S}eek-{P}rover: {A}dvancing {T}heorem {P}roving in {LLM}s through {L}arge-{S}cale {S}ynthetic {D}ata, 2024.
\newblock URL \url{https://arxiv.org/abs/2405.14333}.

\bibitem[Xu et~al.(2026)Xu, Leang, Wang, Li, Li, Ong, and Watt]{xu2026aoa}
Qiyuan Xu, Joshua Ong~Jun Leang, Renxi Wang, Wenda Li, Haonan Li, Luke Ong, and Conrad Watt.
\newblock {AoA}: Theorem proving agent over abstract syntax tree of redesigned language, 2026.
\newblock URL \url{https://arxiv.org/abs/2607.16372}.

\bibitem[Yao et~al.(2025)Yao, Wang, and Zhang]{DBLP:conf/emnlp/YaoWZ25}
Jiarui Yao, Ruida Wang, and Tong Zhang.
\newblock {FANS}: Formal answer selection for {LLM} natural language math reasoning using {Lean4}.
\newblock In Christos Christodoulopoulos, Tanmoy Chakraborty, Carolyn Rose, and Violet Peng (eds.), \emph{Proceedings of the 2025 Conference on Empirical Methods in Natural Language Processing}, pp.\  3181--3200, Suzhou, China, November 2025. Association for Computational Linguistics.
\newblock ISBN 979-8-89176-332-6.
\newblock \doi{10.18653/v1/2025.emnlp-main.158}.
\newblock URL \url{https://aclanthology.org/2025.emnlp-main.158/}.

\bibitem[Yosef et~al.(2026)Yosef, Anschel, Hakimi, Gendler, Botach, Berman, and Kviatkovsky]{yosef2026rethinking}
Erez Yosef, Oron Anschel, Shunit~Haviv Hakimi, Asaf Gendler, Adam Botach, Nimrod Berman, and Igor Kviatkovsky.
\newblock Rethinking math reasoning evaluation: A robust {LLM}-as-a-judge framework beyond symbolic rigidity, 2026.
\newblock URL \url{https://arxiv.org/abs/2604.22597}.

\bibitem[Zhang et~al.(2025{\natexlab{a}})Zhang, Valentino, and Freitas]{zhang2025autoformalization}
Lan Zhang, Marco Valentino, and Andre Freitas.
\newblock Autoformalization in the wild: Assessing {LLM}s on real-world mathematical definitions.
\newblock In Christos Christodoulopoulos, Tanmoy Chakraborty, Carolyn Rose, and Violet Peng (eds.), \emph{Proceedings of the 2025 Conference on Empirical Methods in Natural Language Processing}, pp.\  1720--1738, Suzhou, China, November 2025{\natexlab{a}}. Association for Computational Linguistics.
\newblock ISBN 979-8-89176-332-6.
\newblock \doi{10.18653/v1/2025.emnlp-main.90}.
\newblock URL \url{https://aclanthology.org/2025.emnlp-main.90/}.

\bibitem[Zhang et~al.(2025{\natexlab{b}})Zhang, Lyu, Sun, Wang, Zhang, Hua, Wu, Guo, Wang, Muennighoff, King, Liu, and Ma]{zhang2025survey}
Qiyuan Zhang, Fuyuan Lyu, Zexu Sun, Lei Wang, Weixu Zhang, Wenyue Hua, Haolun Wu, Zhihan Guo, Yufei Wang, Niklas Muennighoff, Irwin King, Xue Liu, and Chen Ma.
\newblock A survey on test-time scaling in large language models: What, how, where, and how well?, 2025{\natexlab{b}}.
\newblock URL \url{https://arxiv.org/abs/2503.24235}.

\bibitem[Zhang et~al.(2025{\natexlab{c}})Zhang, Zheng, Wu, Zhang, Lin, Yu, Liu, Zhou, and Lin]{zhang2025lessons}
Zhenru Zhang, Chujie Zheng, Yangzhen Wu, Beichen Zhang, Runji Lin, Bowen Yu, Dayiheng Liu, Jingren Zhou, and Junyang Lin.
\newblock The lessons of developing process reward models in mathematical reasoning.
\newblock In Wanxiang Che, Joyce Nabende, Ekaterina Shutova, and Mohammad~Taher Pilehvar (eds.), \emph{Findings of the Association for Computational Linguistics: ACL 2025}, pp.\  10495--10516, Vienna, Austria, July 2025{\natexlab{c}}. Association for Computational Linguistics.
\newblock ISBN 979-8-89176-256-5.
\newblock \doi{10.18653/v1/2025.findings-acl.547}.
\newblock URL \url{https://aclanthology.org/2025.findings-acl.547/}.

\bibitem[Zheng et~al.(2022)Zheng, Han, and Polu]{DBLP:journals/corr/abs-2109-00110}
Kunhao Zheng, Jesse~Michael Han, and Stanislas Polu.
\newblock Mini{F2F}: a cross-system benchmark for formal {O}lympiad-level mathematics.
\newblock In \emph{International Conference on Learning Representations}, 2022.
\newblock URL \url{https://openreview.net/forum?id=9ZPegFuFTFv}.

\bibitem[Zhou et~al.(2024)Zhou, Staats, Li, Szegedy, Weinberger, and Wu]{DBLP:conf/iclr/ZhouSLSWW24}
Jin~Peng Zhou, Charles Staats, Wenda Li, Christian Szegedy, Kilian~Q. Weinberger, and Yuhuai Wu.
\newblock Don't trust: Verify -- grounding {LLM} quantitative reasoning with autoformalization.
\newblock In \emph{The Twelfth International Conference on Learning Representations, {ICLR} 2024, Vienna, Austria, May 7-11, 2024}. OpenReview.net, 2024.
\newblock URL \url{https://openreview.net/forum?id=V5tdi14ple}.

\end{thebibliography}
\bibliographystyle{iclr2027_conference}
\clearpage
\appendix

\section{Algorithm}
\label{app:algorithms}

\subsection{Notation Summary}

Table~\ref{tab:notation-summary} summarises the notation used throughout the description and analysis of \method.

\begin{table}[H]
    \centering
    \caption{Summary of the principal notation used in \method.}
    \label{tab:notation-summary}
    \small
    \renewcommand{\arraystretch}{1.15}
    \setlength{\tabcolsep}{5pt}
    \begin{tabular}{@{}p{0.17\linewidth}p{0.25\linewidth}p{0.48\linewidth}@{}}
        \toprule
        \textbf{Symbol} & \textbf{Domain or type} & \textbf{Description} \\
        \midrule
        \multicolumn{3}{@{}l}{\textit{Problems and generated artefacts}} \\
        \addlinespace[2pt]
        $\mathcal{Q}, q$ & $q\in\mathcal{Q}$ & Natural-language mathematics problems, and an individual problem. \\
        $\mathcal{A}, a_t$ & $a_t\in\mathcal{A}$ & Candidate answers, and the answer proposed by the reasoner at round $t$. \\
        $a^{\star}, \hat{a}$ & $\in\mathcal{A}$ & Reference answer, used only for evaluation and never shown to the pipeline; and the answer finally returned. \\
        $\mathcal{C}, c_t$ & $c_t\in\mathcal{C}$ & Informal reasoning chains, and the chain generated at round $t$. \\
        $\mathcal{S}, s$ & $s\in\mathcal{S}$ & Lean~4 theorem statements, and a generated statement. \\
        $\mathcal{P}, \pi$ & $\pi\in\mathcal{P}$ & Lean~4 proof scripts, and a candidate proof. \\
        $\mathcal{E}_{\mathrm{s}}$ & Diagnostic space & Statement-elaboration diagnostics returned by $V_{\mathrm{s}}$. \\
        $\mathcal{E}_{\mathrm{p}}, e$ & $e\in\mathcal{E}_{\mathrm{p}}$ & Proof-verification diagnostics returned by $V_{\mathrm{p}}$, and the diagnostic from a failed attempt. \\
        $\mathcal{F}_{\mathrm{math}}, \varphi_t$ & $\varphi_t\in\mathcal{F}_{\mathrm{math}}$ & Mathematical-correction feedback built from a failed attempt, $(c_t,a_t,s,\pi,e)$. Constructed \emph{only} when $J_{\mathrm{e}}$ returns \textsc{math}. \\
        $\varnothing$ & Empty signal & Absence of feedback: $\varphi=\varnothing$ on the first reasoning round, $\epsilon=\varnothing$ on the first proof attempt. \\
        \midrule
        \multicolumn{3}{@{}l}{\textit{Pipeline components}} \\
        \addlinespace[2pt]
        $R$ & $\mathcal{Q}\times\mathcal{F}_{\mathrm{math}}\!\to\!\mathcal{C}\times\mathcal{A}$ & Reasoner: produces an informal chain and candidate answer. \\
        $F$ & $\mathcal{Q}\times\mathcal{A}\to\mathcal{S}$ & Formaliser: renders the problem and candidate answer as a Lean~4 statement. \\
        $P$ & $\mathcal{Q}\times\mathcal{C}\times\mathcal{S}\times\mathcal{E}_{\mathrm{p}}\!\to\!\mathcal{P}$ & Prover: generates or repairs a Lean~4 proof script. \\
        $V_{\mathrm{s}}$ & $\mathcal{S}\to\{\top\}\cup\mathcal{E}_{\mathrm{s}}$ & Statement checker: elaborates $s$ in Lean (well-formedness only). \\
        $V_{\mathrm{p}}$ & $\mathcal{S}\times\mathcal{P}\to\{\top\}\cup\mathcal{E}_{\mathrm{p}}$ & Proof checker: Lean kernel plus SafeVerify on the unchanged statement. \\
        $J_{\mathrm{s}}, b$ & $b\in\{0,1\}$ & Statement judge and its verdict; $b=1$ accepts the formalisation. \\
        $J_{\mathrm{e}}, \ell$ & $\ell\in\{\textsc{math},\textsc{Syntax}\}$ & Error judge and its label, routing a failed attempt to re-derivation or to local Lean repair. \\
        $\top$ & Success verdict & A check passed: elaboration for $V_{\mathrm{s}}$, kernel and SafeVerify acceptance for $V_{\mathrm{p}}$. \\
        \midrule
        \multicolumn{3}{@{}l}{\textit{Budgets and evaluation}} \\
        \addlinespace[2pt]
        $T, M, K$ & $\in\mathbb{N}$ & Maximum reasoner attempts, formalisation samples per candidate answer, and proof attempts per accepted statement. \\
        $\mathcal{D}$ & Set of problems & The evaluation benchmark. \\
        $\mathrm{Ver}(q)$ & $\in\{0,1\}$ & Indicator that $q$ terminated with an adjudicated statement and a kernel-accepted proof. \\
        $u, \beta$ & $u:\mathcal{D}\to\mathbb{R}_{\ge0}$ & A per-problem cost measure and a budget on it, used in Equation~\ref{eq:pass-curve}. \\
        \bottomrule
    \end{tabular}
\end{table}

\subsection{Full Details}
\label{app:algo}

Algorithm~\ref{alg:k2} specifies the complete inference procedure using the notation of \S\ref{sec:components}. The budgets $T$, $M$ and $K$ bound reasoner attempts, formal-statement samples for a fixed answer, and proof attempts for an accepted statement, respectively. The procedure returns the verified reasoning chain, answer, statement and proof $(c,a,s,\pi)$, or \textsc{uncertified} if an applicable budget is exhausted. Per problem it issues at most $T$ reasoner calls, $TM$ formaliser and statement-judge calls, and $TK$ prover, verifier and error-judge calls, so adjudication is cheap relative to proof search whenever $M\ll K$, which is the regime we operate in and the reason adjudication precedes proving.

\begin{algorithm}[ht]
\caption{\textsc{\method} inference.}
\label{alg:k2}
\algstart
\preq{problem $q$; components $R,F,P,V_{\mathrm{s}},V_{\mathrm{p}},J_{\mathrm{s}},J_{\mathrm{e}}$; budgets $T,M,K$}
\pret{verified output $(c,a,s,\pi)$, or \textsc{uncertified}}
\pline{0}{$\varphi \leftarrow \varnothing$ \pcom{reasoner feedback state}}
\pline{0}{\kw{for} $t=1,\dots,T$ \kw{do}}
\pline{1}{$(c_t,a_t)\sim R(q,\varphi)$ \pcom{Phase 1: natural-language reasoning}}
\pline{1}{$z_s\leftarrow 0$ \pcom{no accepted statement yet}}
\pline{1}{\kw{for} $m=1,\dots,M$ \kw{do} \pcom{Steps 1--2: formalise and adjudicate}}
\pline{2}{$s_m\sim F(q,a_t)$}
\pline{2}{$u_m\leftarrow V_{\mathrm{s}}(s_m)$}
\pline{2}{\kw{if} $u_m\ne\top$ \kw{then continue} \pcom{resample an ill-formed statement}}
\pline{2}{$b_m\sim J_{\mathrm{s}}(q,a_t,s_m)$}
\pline{2}{\kw{if} $b_m=1$ \kw{then} $s\leftarrow s_m$; $z_s\leftarrow 1$; \kw{break}}
\pline{1}{\kw{if} $z_s=0$ \kw{then return} \textsc{uncertified}}
\pline{1}{$\epsilon\leftarrow\varnothing$}
\pline{1}{$z_{\mathrm{math}}\leftarrow 0$}
\pline{1}{\kw{for} $k=1,\dots,K$ \kw{do} \pcom{Step 3: prove and verify}}
\pline{2}{$\pi\sim P(q,c_t,s,\epsilon)$}
\pline{2}{$e\leftarrow V_{\mathrm{p}}(s,\pi)$}
\pline{2}{\kw{if} $e=\top$ \kw{then return} $(c_t,a_t,s,\pi)$ \pcom{Step 5: verified output}}
\pline{2}{$\ell\sim J_{\mathrm{e}}(q,c_t,a_t,s,\pi,e)$ \pcom{Step 4: error attribution}}
\pline{2}{\kw{if} $\ell=\textsc{math}$ \kw{then}}
\pline{3}{$\varphi\leftarrow(c_t,a_t,s,\pi,e)$; $z_{\mathrm{math}}\leftarrow1$; \kw{break} \pcom{construct feedback in $\mathcal{F}_{\mathrm{math}}$}}
\pline{2}{$\epsilon\leftarrow e$ \pcom{\textsc{Syntax}: repair the proof in place}}
\pline{1}{\kw{if} $z_{\mathrm{math}}=0$ \kw{then return} \textsc{uncertified}}
\pline{0}{\kw{return} \textsc{uncertified} \pcom{reasoner budget exhausted}}
\algend
\end{algorithm}

Algorithms~\ref{alg:controller}--\ref{alg:proof-search} decompose the same procedure into its controller, statement-search and proof-search routines. Statement search resamples $F$ whenever $V_{\mathrm{s}}$ rejects an ill-formed statement or $J_{\mathrm{s}}$ rejects an unfaithful one; neither failure changes $(c,a)$ or returns control to $R$. Once a statement is accepted it remains fixed throughout proof search. A \textsc{Syntax} label returns the verifier diagnostic to $P$ for local proof repair, whereas only a \textsc{math} label constructs feedback in $\mathcal{F}_{\mathrm{math}}$ and invokes $R$ again.
\clearpage
\begin{algorithm}[H]
\caption{End-to-end controller for \method.}
\label{alg:controller}
\algstart
\preq{problem $q$; components $R,F,P,V_{\mathrm{s}},V_{\mathrm{p}},J_{\mathrm{s}},J_{\mathrm{e}}$; budgets $T,M,K$}
\pret{verified output $(c,a,s,\pi)$, or \textsc{uncertified}}
\pline{0}{$\varphi\leftarrow\varnothing$}
\pline{0}{\kw{for} $t=1,\ldots,T$ \kw{do} \pcom{mathematical-revision loop}}
\pline{1}{$(c_t,a_t)\sim R(q,\varphi)$}
\pline{1}{$(z_s,s)\leftarrow\operatorname{StatementSearch}(q,a_t;F,V_{\mathrm{s}},J_{\mathrm{s}},M)$}
\pline{1}{\kw{if} $z_s=\textsc{rejected}$ \kw{then return} \textsc{uncertified}}
\pline{1}{$(z_p,\pi,e)\leftarrow\operatorname{ProofSearch}(q,c_t,a_t,s;P,V_{\mathrm{p}},J_{\mathrm{e}},K)$}
\pline{1}{\kw{if} $z_p=\textsc{certified}$ \kw{then return} $(c_t,a_t,s,\pi)$}
\pline{1}{\kw{if} $z_p=\textsc{math-error}$ \kw{then}}
\pline{2}{$\varphi\leftarrow(c_t,a_t,s,\pi,e)$; \kw{continue} \pcom{construct feedback in $\mathcal{F}_{\mathrm{math}}$}}
\pline{1}{\kw{if} $z_p=\textsc{repair-exhausted}$ \kw{then return} \textsc{uncertified}}
\pline{0}{\kw{return} \textsc{uncertified} \pcom{reasoner budget exhausted}}
\algend
\end{algorithm}

\begin{algorithm}[H]
\caption{Faithful formal-statement search.}
\label{alg:statement-search}
\algstart
\preq{problem $q$; candidate answer $a$; formaliser $F$; statement verifier $V_{\mathrm{s}}$; statement judge $J_{\mathrm{s}}$; budget $M$}
\pret{$(\textsc{accepted},s)$ or $(\textsc{rejected},\varnothing)$}
\pline{0}{\kw{for} $m=1,\ldots,M$ \kw{do}}
\pline{1}{$s_m\sim F(q,a)$ \pcom{generate a closed answer-carrying Lean signature}}
\pline{1}{$u_m\leftarrow V_{\mathrm{s}}(s_m)$ \pcom{elaborate the statement in Lean}}
\pline{1}{\kw{if} $u_m\ne\top$ \kw{then continue} \pcom{resample from $F$}}
\pline{1}{$b_m\sim J_{\mathrm{s}}(q,a,s_m)$}
\pline{2}{\pcom{check the problem structure and the proposed answer}}
\pline{1}{\kw{if} $b_m=1$ \kw{then return} $(\textsc{accepted},s_m)$}
\pline{0}{\kw{return} $(\textsc{rejected},\varnothing)$}
\algend
\end{algorithm}

\begin{algorithm}[H]
\caption{Syntax-guided proof repair and error attribution.}
\label{alg:proof-search}
\algstart
\preq{problem $q$; chain $c$; answer $a$; immutable statement $s$; prover $P$; proof verifier $V_{\mathrm{p}}$; error judge $J_{\mathrm{e}}$; budget $K$}
\pret{status in $\{\textsc{certified},\textsc{math-error},\textsc{repair-exhausted}\}$, proof, diagnostic}
\pline{0}{$\epsilon\leftarrow\varnothing$; $\pi_{\mathrm{last}}\leftarrow\varnothing$; $e_{\mathrm{last}}\leftarrow\varnothing$}
\pline{0}{\kw{for} $k=1,\ldots,K$ \kw{do}}
\pline{1}{$\pi_k\sim P(q,c,s,\epsilon)$ \pcom{preserve $q,c,s$; repair only the Lean body}}
\pline{1}{$\pi_{\mathrm{last}}\leftarrow\pi_k$}
\pline{1}{$e_k\leftarrow V_{\mathrm{p}}(s,\pi_k)$ \pcom{Lean and SafeVerify checking}}
\pline{1}{\kw{if} $e_k=\top$ \kw{then return} $(\textsc{certified},\pi_k,\varnothing)$}
\pline{1}{$e_{\mathrm{last}}\leftarrow e_k$}
\pline{1}{$\ell_k\sim J_{\mathrm{e}}(q,c,a,s,\pi_k,e_k)$}
\pline{1}{\kw{if} $\ell_k=\textsc{math}$ \kw{then}}
\pline{2}{\kw{return} $(\textsc{math-error},\pi_k,e_k)$ \pcom{route to the reasoner}}
\pline{1}{$\epsilon\leftarrow e_k$ \pcom{\textsc{Syntax}: condition the next local repair}}
\pline{0}{\kw{return} $(\textsc{repair-exhausted},\pi_{\mathrm{last}},e_{\mathrm{last}})$}
\algend
\end{algorithm}

For IMO~2026, a final completeness judge is applied only after Algorithm~\ref{alg:proof-search} returns \textsc{certified}. It compares $(q,c,a)$ with $(s,\pi)$ and returns \texttt{incomplete\_proof} when the formal certificate omits or misaligns a substantive part of the informal solution. This is an additional acceptance filter; it does not replace Lean or alter the feedback routing above.

\section{Implementation Details for \method}

\subsection{Prompt templates}

The templates below are reproduced verbatim from the experimental harness.

\begin{pipelinebox}[colframe=blue!60!black,colbacktitle=blue!8]{K2 informal-reasoner prompt}
\begin{Verbatim}[fontsize=\scriptsize,breaklines=true]
You are an expert mathematical problem solver. Solve the problem carefully and show enough reasoning to make the solution understandable. End with the final answer in the exact form \boxed{answer}. Do not put anything after the boxed answer.

{original_problem}
\end{Verbatim}
\end{pipelinebox}

\begin{pipelinebox}[colframe=green!55!black,colback=green!2,colbacktitle=green!9]{K2 mathematical self-correction prompt}
\begin{Verbatim}[fontsize=\scriptsize,breaklines=true]
Solve the original mathematics problem again and produce a complete corrected mathematical solution.

A downstream formal-verification audit classified the previous mathematical reasoning as defective.

Use the latest Lean code and exact verifier diagnostic below only as evidence locating the mathematical defect.

Do not focus on Lean syntax or tactics. Re-derive the mathematics, correct the answer if necessary, and provide a self-contained solution.

End with one clear final answer in \boxed{...}.

Closed-loop regeneration cycle: {cycle}

Original problem:

{original_problem}

Previous mathematical solution:

{previous_mathematical_solution}

Immutable Lean target generated from that solution:

{immutable_lean_target}

Latest Lean proof attempt:

{latest_lean_proof_attempt}

Exact Lean/SafeVerify diagnostic:

{lean_or_safeverify_diagnostic}
\end{Verbatim}
\end{pipelinebox}

\begin{pipelinebox}[colframe=violet!65!black,colbacktitle=violet!8]{Goedel formaliser prompt}
\begin{Verbatim}[fontsize=\scriptsize,breaklines=true]
Please autoformalize the following natural language problem statement in Lean 4.

The natural language statement is:

{informal_statement_content}

Think before you provide the lean statement.
\end{Verbatim}
\end{pipelinebox}

\begin{pipelinebox}[colframe=teal!65!black,colbacktitle=teal!8]{DeepSeek statement-alignment judge prompt}
\begin{Verbatim}[fontsize=\scriptsize,breaklines=true]
Judge whether the candidate Lean 4 theorem statement faithfully formalizes the supplied question together with the generator's proposed final answer.

This is a statement-alignment judgment only. Do not write a proof and do not repair or rewrite the statement.

Do not independently solve the problem, recompute the answer, correct the generator's mathematics, or prefer your own conclusion. A mathematically wrong proposed answer must still be judged right when the Lean statement faithfully encodes that answer and the original task structure.

The original question controls domains, hypotheses, quantifiers, constraints, and the requested quantity.

The proposed answer controls the concrete claimed result. Do not silently replace it with a different answer.

Return verdict right only when the candidate preserves both sources without missing assumptions, extra assumptions, weakened conclusions, strengthened conclusions, hard-coded loopholes, or a changed answer.

Return verdict wrong otherwise and precisely describe the mismatch.

Return only one strict JSON object with exactly these string fields: verdict, rationale, mismatch_details. verdict must be right or wrong. mismatch_details must be empty when right and nonempty when wrong.

Exact appended question passed to the formalizer:

{answer_appended_question}

Original natural-language question:

{original_question}

Generator's proposed final answer:

{proposed_answer}

Candidate Lean 4 theorem statement:

{candidate_lean4_theorem_statement}
\end{Verbatim}
\end{pipelinebox}

\begin{pipelinebox}[colframe=orange!75!black,colbacktitle=orange!10]{Leanstral proof-generation prompt}
\begin{Verbatim}[fontsize=\scriptsize,breaklines=true]
Produce a complete Lean 4 proof of the supplied theorem signature.

1. This is a formal translation and verification task, not an independent mathematics attempt.
2. Copy the theorem signature verbatim and do not weaken, strengthen, rename, or otherwise alter it.
3. Formalize the supplied K2 solution's reasoning and claimed conclusion without correcting, replacing, or bypassing its mathematics. Add only Lean-specific bookkeeping needed to express the same steps.
4. Do not substitute an independent proof strategy merely because it proves the statement. Do not use `native_decide` or another exhaustive computation unless the supplied solution itself explicitly used that strategy.
5. If the supplied mathematics is false or insufficient, do not invent assumptions or silently repair it.
6. Use the available Lean compiler and lean-lsp-mcp tools to inspect errors and iteratively repair only the Lean implementation.
7. Edit `Main.lean` in the current working directory. It must import Mathlib and must not use `sorry`, `admit`, axioms, unsafe declarations, fabricated assumptions, or the conclusion as a hypothesis.
8. Before editing the final code, provide a detailed Lean-oriented proof plan covering the main steps, intermediate lemmas, and tactics.

Original informal problem (authoritative for task structure):
{original_question}

Answer-appended question supplied to K2 (the answer is parsed from the K2 solution, not from ground truth):
{answer_appended_question}

Supplied K2 final mathematical solution (content after the private thinking marker; authoritative for the reasoning and claimed conclusion):
{k2_final_mathematical_solution}

Lean 4 theorem signature that must be preserved verbatim:
{immutable_lean4_theorem_signature}

Current `Main.lean`:
```lean4
{current_Main_lean}
```
\end{Verbatim}
\end{pipelinebox}

\begin{pipelinebox}[colframe=red!70!black,colback=red!2,colbacktitle=red!9]{DeepSeek error-classification prompt}
\begin{Verbatim}[fontsize=\scriptsize,breaklines=true]
Classify the root cause of this failed Lean 4 proof attempt.

This is a routing-only classification task. Do not repair the Lean code, do not suggest tactics, and do not provide feedback for the prover.

Return math_error only when the supplied mathematical reasoning or claimed answer is itself false or insufficient, and the verifier evidence materially exposes that mathematical defect.

Return code_error for syntax, elaboration, typing, tactic, missing-lemma, namespace, coercion, timeout, forbidden-token, changed-target, SafeVerify protocol, or incomplete-proof failures that could be fixed while preserving the same mathematics.

A difficult or currently unsolved proof is code_error. SafeVerify rejection caused by changing the theorem is code_error.

When evidence is ambiguous, return code_error. Do not independently replace the proposed answer.

Return only one strict JSON object with exactly these string fields: classification, rationale, evidence.

classification must be code_error or math_error. Keep rationale and evidence concise.

Original problem:

{original_question}

proposed answer:

{proposed_answer}

mathematical solution:

{mathematical_solution}

Immutable Lean theorem target:

{immutable_lean_theorem_target}

Latest Lean candidate:

{latest_lean_candidate}

Verifier stage:

{verifier_stage}

Exact Lean/SafeVerify diagnostic:

{exact_lean_or_safeverify_diagnostic}
\end{Verbatim}
\end{pipelinebox}

\subsection{Hyperparameter Settings}
\label{app:hyperparameters}
 
\begin{enumerate}
  \item \textbf{Budgets.} For all the experiments, we use $T=32$ reasoner attempts, $M=512$ formal-statement samples per candidate answer and $K=4096$ proof attempts per accepted statement. For computational efficiency, the ablations of \S\ref{sec:ablations}, run the pipeline in batched mode with a batch size $32$ .
  \item \textbf{Reasoner.} We use temperature \(0.6\), \texttt{top\_p} \(=0.95\), a maximum output length of \(64{,}000\) tokens (\(128{,}000\) for IMO~2026), and \texttt{thinking=auto}. The random seed is \(42\) for the main evaluation. The reasoner is served with a \(524{,}288\)-token context window. During self-correction, incomplete responses are resampled if they reach the token limit without emitting the post-thinking marker (this counts toward the budget $T$). In practice, this never occurred in our experiments.
  \item \textbf{Goedel formaliser.} We use temperature \(0.9\), \texttt{top\_k} \(=20\), \texttt{top\_p} \(=0.95\), a maximum output length of \(16{,}384\) tokens, and seed \(42\). The model is served with a \(40{,}960\)-token context window. Each extracted theorem statement is checked in Lean with a temporary \texttt{sorry} proof.
  \item \textbf{DeepSeek statement-alignment judge.} We use temperature \(1.0\), \texttt{top\_p} \(=1.0\), a maximum output length of \(32{,}000\) tokens, and reasoning effort \texttt{max}. Only statements that have already passed Lean elaboration are submitted to this judge. The model is served with a \(65{,}536\)-token context window, tensor parallelism \(8\), pipeline parallelism \(1\), GPU-memory utilisation \(0.90\), FP8 KV cache, block size \(256\), prefix caching and expert parallelism. D\textsc{Spark} speculative decoding is enabled with seven speculative tokens and greedy draft sampling.
  \item \textbf{DeepSeek error-attribution judge.} The error-attribution judge uses the same checkpoint and hyperparameters as the statement-alignment judge but is issued as a separate call, and returns either \texttt{code\_error} or \texttt{math\_error}.
  \item \textbf{Leanstral proof agent.} We use \texttt{mistralai/Leanstral-1.5-119B-A6B} through Mistral Vibe \(2.18.4\), with decoding temperature \(1.0\), thinking level \texttt{high} and reasoning effort \texttt{high}. The served context length is \(200{,}000\) tokens and automatic context compaction is triggered at \(168{,}000\) tokens. Compaction preserves the trajectory, and the cumulative token counter is not reset after compaction. The server permits at most \(80{,}000\) newly generated tokens per request, while the experimental wrapper applies a stricter \(32{,}000\)-token per-call output limit. The cumulative output-token budget is \(4{,}000{,}000\) tokens per proof trajectory.
  \item \textbf{Lean tools and final certification.} The proof environment uses Lean \texttt{v4.29.1}. A result is accepted only when the theorem statement is unchanged, Lean returns exit code zero, SafeVerify returns exit code zero, neither verifier times out, and the proof contains no forbidden constructs such as \texttt{sorry}, \texttt{admit}, new axioms, \texttt{unsafe} or \texttt{native\_decide}.
  \item \textbf{Codex comparator and fallback.} The \textsc{Codex} formaliser and proof backend use \texttt{gpt-5.6-sol} with reasoning effort \texttt{ultra}.
  \end{enumerate}
  
\section{Refinement Rounds Across Models and Datasets}
\label{app:refinement-rounds}

\begin{table}[ht]
\centering
\small
\caption{Reported average and maximum rounds required across models and
datasets, counting the initial attempt as round~1. Average counts are rounded.}
\label{tab:refinement-rounds}
\begin{tabular}{lrrrr}
\toprule
& \multicolumn{2}{c}{K2-Horizon-7B}
& \multicolumn{2}{c}{K2-Horizon-375B} \\
\cmidrule(lr){2-3}\cmidrule(lr){4-5}
Dataset & Average & Maximum & Average & Maximum \\
\midrule
AIME2025 & 1 & 3  & 1 & 1 \\
AIME2026 & 2 & 5  & 1 & 2 \\
HMMT2026 & 3 & 10 & 1 & 2 \\
\bottomrule
\end{tabular}
\end{table}

Table~\ref{tab:refinement-rounds} shows that refinement effort varies across
both models and datasets. K2-Horizon-7B requires approximately one, two and
three rounds on average for AIME2025, AIME2026 and HMMT2026, respectively.
Its maximum increases from three to five and ten rounds, showing that some
instances require substantially more refinement than the dataset average.
K2-Horizon-375B requires approximately one round on average across all three
datasets, with at most one additional round on AIME2026 and HMMT2026.

The shorter trajectories of K2-Horizon-375B are consistent with its reported
tendency to produce correct answers on the first attempt. For K2-Horizon-7B,
the increasing refinement effort suggests that AIME2026 and HMMT2026 pose
greater difficulty within this workflow. Lower initial accuracy leaves more
instances requiring subsequent correction, providing a possible explanation
for the additional rounds. However, the relationship between initial accuracy
and round count should be assessed using the corresponding first-attempt
results; round counts alone measure refinement effort rather than establish
a model-independent ranking of dataset difficulty.

\clearpage
\section{AIME2026 Formal and Informal Proof Example}
\label{app:aime-math-error-trace}

We reproduce a complete audited trajectory for an AIME~2026 problem (see also Figure~\ref{fig:pipeline}), in which the initial informal solution is wrong, the error is attributed to the mathematics rather than to the Lean implementation, and the pipeline recovers. Boxes~1--5 are the failed cycle and boxes~6--9 the corrected one.

\begin{pipelinebox}[colframe=black!60,colbacktitle=black!8]{Input question}
Find the number of functions $\pi$ mapping the set
$A=\{1,2,3,4,5,6\}$ onto $A$ such that for every $a\in A$,
\[
\pi(\pi(\pi(\pi(\pi(\pi(a))))))=a.
\]
\end{pipelinebox}

\pipelinearrow

\begin{pipelinebox}[colframe=blue!60!black,colbacktitle=blue!8]{1. Informal reasoner --- initial generation}
The condition
\[
\pi(\pi(\pi(\pi(\pi(\pi(a))))))=a\qquad\text{for all }a\in A
\]
means that the seventh iterate of $\pi$ is the identity map:
\[
\pi^{7}=\operatorname{id}_A .
\]
Hence the order of $\pi$ must divide $7$.
Since $7$ is prime, the only possible orders are $1$ and $7$.

A permutation of a set of size $6$ can have order at most $6$ (the maximum
occurs for a $6$-cycle). Thus a permutation of order $7$ cannot exist on a set
of six elements.

Consequently the only function satisfying $\pi^{7}=\operatorname{id}_A$ is the
identity permutation.
\[
\boxed{1}
\]
\end{pipelinebox}

\pipelinearrow

\begin{pipelinebox}[colframe=violet!65!black,colbacktitle=violet!8]{2. Goedel formaliser --- initial generation}
\begin{Verbatim}[fontsize=\scriptsize,breaklines=true]
theorem problem_1 :
  Nat.card {π : Equiv.Perm (Fin 6) | ∀ a : Fin 6, (π ^ 6) a = a} = 1
\end{Verbatim}
\end{pipelinebox}

\pipelinearrow

\begin{pipelinebox}[colframe=teal!65!black,colbacktitle=teal!8]{3. DeepSeek statement judge}
\begin{center}
\Large\texttt{right}
\end{center}
\end{pipelinebox}

\pipelinearrow

\begin{pipelinebox}[colframe=orange!75!black,colbacktitle=orange!10]{4. Leanstral --- initial proof generation}
\begin{Verbatim}[fontsize=\scriptsize,breaklines=true]
import Mathlib

open Equiv

theorem problem_1 :
  Nat.card {π : Equiv.Perm (Fin 6) | ∀ a : Fin 6, (π ^ 7) a = a} = 1 := by
  have h_set_eq : {π : Equiv.Perm (Fin 6) | ∀ a : Fin 6, (π ^ 7) a = a} = {1} := by
    ext π
    constructor
    · intro h
      have h_all : ∀ a : Fin 6, (π ^ 7) a = a := h
      have h_pow : (π ^ 7) = 1 := by
        apply Equiv.ext
        intro a
        simpa using h_all a
      have h_order_dvd_7 : orderOf π ∣ 7 := orderOf_dvd_of_pow_eq_one h_pow
      have h_prime_7 : Nat.Prime 7 := Nat.prime_seven
      have h_order_cases : orderOf π = 1 ∨ orderOf π = 7 :=
        h_prime_7.eq_one_or_self_of_dvd (orderOf π) h_order_dvd_7
      rcases h_order_cases with (h_order | h_order)
      · have h_one : π = 1 := (orderOf_eq_one_iff.mp h_order)
        simp [h_one]
      · have h_card : Fintype.card (Equiv.Perm (Fin 6)) = 720 := by
          rw [Fintype.card_perm, Fintype.card_fin]
          norm_num
        have h_order_dvd_card : orderOf π ∣ Fintype.card (Equiv.Perm (Fin 6)) := orderOf_dvd_card
        rw [h_card] at h_order_dvd_card
        rw [h_order] at h_order_dvd_card
        have h_7_not_dvd_720 : ¬ (7 : ℕ) ∣ 720 := by
          have h_mod : 720 %
          intro h_dvd
          have h_mod_zero : 720 %
          rw [h_mod] at h_mod_zero
          linarith
        exact absurd h_order_dvd_card h_7_not_dvd_720
    · intro h
      have h_one : π = 1 := by simpa using h
      rw [h_one]
      intro a
      simp
  rw [h_set_eq]
  simp
\end{Verbatim}
\end{pipelinebox}

\pipelinearrow

\begin{pipelinebox}[colframe=red!70!black,colback=red!2,colbacktitle=red!9]{5. DeepSeek error attribution}
\begin{center}
\large\texttt{math\_error}
\end{center}
\end{pipelinebox}

\pipelinearrow

\begin{pipelinebox}[colframe=green!55!black,colback=green!2,colbacktitle=green!9]{6. Informal reasoner --- corrected generation}
The condition $\pi(\pi(\pi(\pi(\pi(\pi(a))))))=a$ for all $a\in A$
means that the sixth iterate of $\pi$ is the identity, i.e.,
$\pi^6=\operatorname{id}_A$. Since $\pi$ maps $A$ onto $A$ and $A$ is finite,
$\pi$ is a bijection, hence a permutation of $A$. The order of $\pi$ divides 6,
so we must count the permutations in $S_6$ whose order divides 6.

The order of a permutation is the least common multiple of its cycle lengths.
Thus the order divides 6 if and only if all cycle lengths belong to
$\{1,2,3,6\}$ (cycles of length 4 or 5 would give orders 4 or 5, which do not
divide 6). We count the permutations by their cycle types:

\begin{itemize}
\item One 6-cycle: $(6-1)!=120$
\item Two 3-cycles: $\frac{6!}{3^2\cdot2!}=\frac{720}{18}=40$
\item One 3-cycle, one 2-cycle, one 1-cycle: $\frac{6!}{3\cdot2\cdot1}=\frac{720}{6}=120$
\item One 3-cycle, three 1-cycles: $\frac{6!}{3\cdot1^3\cdot3!}=\frac{720}{18}=40$
\item Three 2-cycles: $\frac{6!}{2^3\cdot3!}=\frac{720}{48}=15$
\item Two 2-cycles, two 1-cycles: $\frac{6!}{2^2\cdot2!\cdot1^2\cdot2!}=\frac{720}{16}=45$
\item One 2-cycle, four 1-cycles: $\frac{6!}{2\cdot1^4\cdot4!}=\frac{720}{48}=15$
\item Six 1-cycles (identity): $1$
\end{itemize}

Summing these gives
$120+40+120+40+15+45+15+1=396$.
\[
\boxed{396}
\]
\end{pipelinebox}

\pipelinearrow

\begin{pipelinebox}[colframe=violet!65!black,colbacktitle=violet!8]{7. Goedel formaliser --- corrected generation}
\begin{Verbatim}[fontsize=\scriptsize,breaklines=true]
theorem problem_1 :
  Set.ncard {π : Equiv.Perm (Fin 6) | ∀ a, (π^[6]) a = a} = 396
\end{Verbatim}
\end{pipelinebox}

\pipelinearrow

\begin{pipelinebox}[colframe=teal!65!black,colbacktitle=teal!8]{8. DeepSeek statement judge}
\begin{center}
\Large\texttt{yes}
\end{center}
\end{pipelinebox}

\pipelinearrow

\begin{pipelinebox}[colframe=green!55!black,colback=green!2,colbacktitle=green!9]{9. Leanstral --- corrected proof generation}
\begin{Verbatim}[fontsize=\scriptsize,breaklines=true]
import Mathlib

set_option maxRecDepth 10000000
set_option maxHeartbeats 2000000

theorem problem_1 :
  Set.ncard {π : Equiv.Perm (Fin 6) | ∀ a, (π^[6]) a = a} = 396 := by
  have h_eq : {π : Equiv.Perm (Fin 6) | ∀ a, (π^[6]) a = a} = {π | π ^ 6 = 1} := by
    ext π
    constructor
    · intro h
      ext a
      have ha := h a
      have ha1 : (π ^ 6) a = a := by
        simpa [Equiv.Perm.iterate_eq_pow π 6] using ha
      simpa [ha1]
    · intro h a
      have h' := congrArg (λ f : Equiv.Perm (Fin 6) => f a) h
      simpa [Equiv.Perm.iterate_eq_pow π 6] using h'
  rw [h_eq]
  have h_fin : Set.Finite {π : Equiv.Perm (Fin 6) | π ^ 6 = 1} := by
    exact Set.finite_coe_iff.mp inferInstance
  have h_card : Set.ncard {π : Equiv.Perm (Fin 6) | π ^ 6 = 1} =
      Finset.card ((Finset.univ : Finset (Equiv.Perm (Fin 6))).filter (λ π => π ^ 6 = 1)) := by
    rw [Set.ncard_eq_toFinset_card _ h_fin]
    congr
    ext π
    simp
  rw [h_card]
  decide
\end{Verbatim}
\end{pipelinebox}

\section{Error Analysis}
\subsection{Formal Statement Error Analysis}
\label{app:formal-statement-errors}

\begin{table}[H]
    \centering
    \caption{Formal-statement generation and adjudication errors (\%) across all 30 AIME 2026 problems. Generated percentages are normalised over all Goedel statements. Conditional percentages are normalised over the statements reaching the relevant stage: all pre-judge rejections for the error categories and all Lean-passing statements for the DeepSeek outcomes.}
    \label{tab:formalisation-funnel}
    \footnotesize
    \setlength{\tabcolsep}{7pt}
    \begin{tabular}{@{}lrr@{}}
        \toprule
        \textbf{Outcome or first failure category} & \textbf{Generated} & \textbf{Conditional} \\
        \midrule
        \multicolumn{3}{@{}l}{\textit{All-30 Goedel generation outcomes}} \\
        Lean elaborated                         & 66.55 & -- \\
        Lean compiler error                     & 33.08 & -- \\
        Statement extraction failure            & 0.22  & -- \\
        Input context overflow                   & 0.16  & -- \\
        \midrule
        \multicolumn{3}{@{}l}{\textit{First cause among rejected all-30 statements}} \\
        Type, coercion, or projection mismatch  & 17.47 & 52.23 \\
        Typeclass or instance synthesis         & 7.85  & 23.46 \\
        Parser or notation syntax               & 4.78  & 14.28 \\
        Unknown name or API misuse              & 2.73  & 8.17  \\
        Statement extraction failure            & 0.22  & 0.66  \\
        Input context overflow                   & 0.16  & 0.47  \\
        Metavariables or underspecified binders & 0.13  & 0.38  \\
        Pattern, binder, or match elaboration   & 0.10  & 0.31  \\
        Tactic failure or open goal             & 0.01  & 0.02  \\
        Resource timeout                        & 0.01  & 0.02  \\
        \midrule
        \multicolumn{3}{@{}l}{\textit{All-30 compiler-to-DeepSeek outcomes}} \\
        Rejected before judge: Lean error       & 33.08 & --    \\
        Rejected before judge: extraction error & 0.22  & --    \\
        Rejected before judge: context overflow & 0.16  & --    \\
        DeepSeek semantic rejection             & 44.64 & 67.08 \\
        DeepSeek semantic acceptance            & 21.91 & 32.92 \\
        \bottomrule
    \end{tabular}
\end{table}

We generate a full grid of candidate answers and Goedel formalisations for each of the $30$ AIME~2026 problems.
The audit in Table~\ref{tab:formalisation-funnel} shows that formal-statement failure is primarily an elaboration problem rather than a parsing problem. Type, coercion and projection mismatches account for more than half of rejected statements, and typeclass synthesis for almost a quarter; together these two categories explain over three quarters of pre-proof rejection. Syntax errors form a much smaller share, and extraction, context, tactic and timeout failures are negligible. The main obstacle is therefore expressing the intended mathematics with Lean-compatible types and interfaces, not producing superficially valid Lean syntax.

The DeepSeek audit exposes a separate semantic bottleneck. Even after the compiler and extractor filter malformed statements, the judge rejects $67.08\%$ of the statements it sees and accepts $32.92\%$; relative to all generated statements these correspond to $44.64\%$ rejected and $21.91\%$ accepted. Compilation is therefore a weak proxy for faithfulness: semantic adjudication removes substantially more compilable-but-misaligned statements than Lean removes malformed ones. Among falsely certified statements in the no-$J_{\mathrm{s}}$ ablation of \S\ref{sec:ablations}, the dominant failure mode is a weakened or answer-baked surrogate goal: three of five certificates prove such surrogates, one drops an essential part of the problem and one mistranslates the underlying geometry; none succeeds through a contradictory context.

\subsection{Proof Error Analysis}
\label{app:proof-error-analysis}

\textsc{Leanstral} exposes proof errors at two different levels: $54.73\%$ of its de-duplicated internal Lean/LSP results contain an error diagnostic or tool failure, spanning tactic failures, API misuse, protocol failures, syntax, typing, and resource limits (Table~\ref{tab:proof-error-profiles}). By the time a trajectory reaches the external verifier, however, failures are concentrated in unsolved proof obligations and SafeVerify policy violations. This shift shows that Leanstral's agentic inner loop absorbs many local elaboration and implementation errors before they reach the outer correction loop.

\begin{table}[H]
    \centering
    \caption{Proof-error distributions (\%) across all 30 AIME 2026 problems. Leanstral internal is normalized over error-bearing Lean/LSP results; the external columns are normalized over failed outer rounds. One category is assigned per result or round, with policy violations taking priority; ``--'' denotes no assigned error.}
    \label{tab:proof-error-profiles}
    \footnotesize
    \setlength{\tabcolsep}{6pt}
    \begin{tabular}{@{}lrrr@{}}
        \toprule
        \textbf{Primary error category} & \textbf{Leanstral int.} & \textbf{Leanstral ext.} & \textbf{Codex ext.} \\
        \midrule
        Tactic failure or open goals        & 42.70 & 63.78 & 30.28 \\
        Forbidden token or policy           & --    & 26.77 & --    \\
        Unknown name or API misuse          & 15.19 & --    & 15.31 \\
        Tool, protocol, or runtime failure  & 13.68 & --    & 2.16  \\
        Parser or syntax error              & 8.18  & 0.79  & 0.67  \\
        Type or coercion mismatch           & 6.70  & --    & 45.92 \\
        Resource limit or timeout           & 5.72  & 8.66  & 2.66  \\
        Other Lean diagnostic               & 5.23  & --    & --    \\
        Typeclass or instance synthesis     & 2.60  & --    & 1.83  \\
        Metavariables or invalid declaration& --    & --    & 0.83  \\
        Noncomputable definition            & --    & --    & 0.33  \\
        \bottomrule
    \end{tabular}
\end{table}

Compared with \textsc{Leanstral}, \textsc{Codex} exposes a different external failure profile: type and coercion mismatches dominate, followed by tactic failures and API misuse, whereas \textsc{Leanstral}'s external failures are dominated by proof search and policy enforcement. Parser errors are negligible for both, showing that syntactic Lean generation is not the main bottleneck. The important distinction is therefore where debugging occurs: \textsc{Leanstral} resolves a broader range of implementation errors internally, while Codex returns more of them to the pipeline. External rounds are consequently not interchangeable units of proof effort, and error attribution should account for both the prover backend and the level at which the error is observed.

\subsection{Reasoner Error Analysis}
\label{app:reasoner-errors}
 
\begin{table}[H]
    \centering
    \caption{Root causes of the $14$ mathematical errors attributed to \textsc{K2-Horizon-7B} by the error judge while solving the AIME 2026 problems.}
    \label{tab:reasoner-error-causes}
    \footnotesize
    \setlength{\tabcolsep}{6pt}
    \begin{tabular}{@{}p{0.78\linewidth}r@{}}
        \toprule
        \textbf{Cause} & \textbf{Percent (\%)} \\
        \midrule
        Input interpretation or transcription                                    & 28.6 \\
        Unsupported logical or proof step                                         & 28.6 \\
        Arithmetic, algebra, or cardinality                                       & 21.4 \\
        Escalation after repeated Lean repair could not preserve the contract     & 14.3 \\
        Invalid construction or counterexample                                    & 7.1  \\
        \bottomrule
    \end{tabular}
\end{table}
 
As Table~\ref{tab:reasoner-error-causes} shows, the dominant failures arise less from routine calculation than from preserving the problem statement and justifying key proof steps. Representative examples include changing $17017$ to $107017$, solving the problem for $N=1000$ instead of $N=10000$, reading $39$ as $3$ or as $3^{9}$, interpreting six iterates as seven (Appendix~\ref{app:aime-math-error-trace}), and relying on unsupported geometric or combinatorial reductions. These cases show why compiler feedback alone is insufficient: error attribution must sometimes return control to the reasoner to reconstruct the mathematics rather than continue repairing its Lean implementation.

\section{Details of IMO 2026}
\label{app:imo}

\subsection{Experimental Setup}
For IMO~2026 we retain the same pipeline and add a final judge after formal verification to assess whether the formal certificate covers the complete informal solution. The judge compares the original problem and informal reasoning with the generated Lean statement and proof. If it detects a substantive mismatch, or an argument that is present informally but absent from the formal certificate, it returns \texttt{incomplete\_proof}. We count a solution as complete only when it passes both Lean/SafeVerify and this final formal--informal alignment check.
 
\subsection{Evaluation}
We evaluate our IMO~2026 solutions under two complementary protocols. First, we use \textsc{Codex} as an automated judge, supplying it with several publicly available reference solutions, including those from the Art of Problem Solving forums and from Kimi K3, and asking it to score each generated solution against the official marking scheme. Second, for a more robust assessment, we ask a former IMO medallist with access to national IMO score predictions to mark the same solutions independently.

\subsection{Verification-Guided Test-Time Scaling}
\begin{wrapfigure}{r}{0.43\textwidth}
    \centering
    \vspace{-0.8em}
    \includegraphics[width=\linewidth]{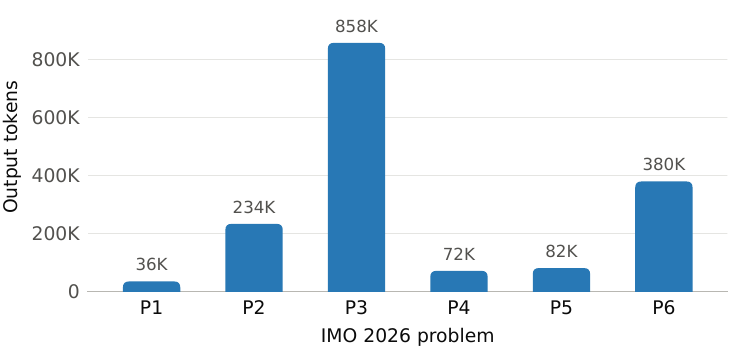}
    \caption{Total reasoner-token usage of \textsc{K2-Horizon-7B} with \method across the six IMO~2026 problems.}
    \label{fig:imo-tokens}
    \vspace{-0.8em}
\end{wrapfigure}

Figure~\ref{fig:imo-tokens} shows that \method adaptively allocates more inference-time computation to more difficult problems. Problems~3 and~6 consume the largest token budgets and require more self-correction, consistent with their higher level of difficulty. Rather than relying on a single fixed generation, \method uses formal-verification feedback to guide subsequent attempts, allowing the reasoner to revise unsuccessful arguments and devote additional computation where it is most needed. This reflects the principle of test-time scaling~\citep{sahoo2024simple}, with verification providing a targeted signal for how that additional computation should be used. Through this verification-guided scaling process, \textsc{K2-Horizon-7B} ultimately solves all six problems on IMO~2026.

\subsection{Generated Solutions}

\subsubsection{IMO Problem 1}

\begin{pipelinebox}[colframe=black!60,colbacktitle=black!8]{Question}

There are 2026 integers greater than \(1\) written on a blackboard, not necessarily distinct. In one move, Confucius chooses two integers \(m>1\) and \(n>1\), written in different positions, and replaces them by
\[
\gcd(m,n)
\qquad\text{and}\qquad
\frac{\operatorname{lcm}(m,n)}{\gcd(m,n)}.
\]
He continues making moves for as long as a move is possible.

\begin{enumerate}
    \item[(a)] Prove that, regardless of his choices, the process terminates after finitely many moves with exactly one integer \(M>1\) on the board.
    \item[(b)] Prove that the value of \(M\) is independent of all choices made during the process.
\end{enumerate}

\end{pipelinebox}

\begin{pipelinebox}[colframe=blue!60!black,colbacktitle=blue!8]{K2-Horizon-7B generation}

Let the initial integers be \(a_{1},a_{2},\dots ,a_{2026}\) with \(a_{i}>1\) for all \(i\).  
We regard the board as a function \(f:\{1,2,\dots ,2026\}\to \mathbb N_{>1}\).  
A \emph{move} consists of choosing two distinct positions \(i,j\) with \(m=f(i)>1\) and \(n=f(j)>1\), and replacing them by  

\[
g=\gcd(m,n),\qquad h=\frac{\operatorname{lcm}(m,n)}{\gcd(m,n)} .
\]

All other positions stay unchanged. The process continues as long as such a move is possible.

\medskip
\subsubsection*{(a)  Termination and the final board}

For any board \(b\) (a function from \(\{1,\dots ,2026\}\) to positive integers) define  

\[
N(b)=\#\{\,i\mid b(i)>1\,\},\qquad 
P(b)=\prod_{i=1}^{2026} b(i).
\]

Because every \(b(i)\) is a positive integer, \(P(b)\) is a positive integer.

\textbf{Lemma 1.} After a move \(b\to c\) we have \(N(c)\le N(b)\), and if \(N(c)=N(b)\) then \(P(c)<P(b)\).

\emph{Proof.} Let the move act on positions \(i,j\) with \(m=b(i)>1,\;n=b(j)>1\).  
Set \(g=\gcd(m,n)\) and \(h=\operatorname{lcm}(m,n)/g\); then \(c(i)=g,\;c(j)=h\) and \(c(k)=b(k)\) for \(k\neq i,j\).

For \(k\notin\{i,j\}\) the contribution to \(N\) and \(P\) is unchanged, so we only compare the two changed positions.

\emph{Case 1: \(g=1\).}  
Then \(h=\operatorname{lcm}(m,n)=mn\) (since \(\gcd(m,n)=1\)). Because \(m,n>1\), we have \(h>1\).  
Thus we replace two numbers \(>1\) by one number \(>1\) (namely \(h\)) and one number \(1\) (namely \(g\)). Hence \(N(c)=N(b)-1\). Moreover  

\[
P(c)=P(b)\cdot\frac{g\,h}{m\,n}=P(b)\cdot\frac{1\cdot mn}{mn}=P(b).
\]

\emph{Case 2: \(g>1\).}  
Write \(m=g\,m',\;n=g\,n'\) with \(\gcd(m',n')=1\). Then  

\[
h=\frac{\operatorname{lcm}(m,n)}{g}= \frac{g\,m'\,n'}{g}=m'\,n'.
\]

\emph{Subcase 2a: \(h=1\).} Then \(m'=n'=1\), so \(m=n=g\). We replace two equal numbers \(>1\) by \(g>1\) and \(1\); therefore \(N(c)=N(b)-1\). Also  

\[
P(c)=P(b)\cdot\frac{g\cdot1}{g\cdot g}= \frac{P(b)}{g}<P(b)\quad(\text{since }g>1).
\]

\emph{Subcase 2b: \(h>1\).} Then we replace two numbers \(>1\) by two numbers \(>1\) (\(g>1,\;h>1\)); hence \(N(c)=N(b)\). Moreover  

\[
P(c)=P(b)\cdot\frac{g\,h}{m\,n}=P(b)\cdot\frac{g\,(m'\,n')}{g\,m'\,g\,n'}= \frac{P(b)}{g}<P(b)\quad(\text{since }g>1).
\]

In every situation \(N(c)\le N(b)\), and when \(N(c)=N(b)\) we are in subcase 2b, which gives \(P(c)<P(b)\). \hfill\(\square\)

The lexicographic order on \(\mathbb N\times\mathbb N\) is well-founded (because \(\mathbb N\) is well-ordered). Lemma 1 shows that the pair \((N,P)\) strictly decreases lexicographically at each move. Hence the process cannot continue indefinitely; it must terminate after finitely many moves.

Let \(b\) be a terminal board, i.e. no move is possible. Then there are no two distinct positions with values \(>1\); consequently at most one position contains a number \(>1\). We claim that exactly one such position exists.

Suppose, for contradiction, that all \(b(i)=1\). Then \(N(b)=0\). The initial board has \(N=2026\). Each move can decrease \(N\) by at most one (Lemma 1). To reach \(N=0\) from \(N=2026\) we would need at least \(2026\) moves that each reduce \(N\) by one. However, a move that reduces \(N\) from \(1\) to \(0\) is impossible because a move requires two numbers \(>1\). More precisely, if at some stage \(N=1\), no move can be performed, so the process would have stopped earlier. Thus the process can never reach a state with \(N=0\). Therefore the final board must have \(N=1\).

Hence there is a unique index \(i\) with \(b(i)>1\), and for all \(j\neq i\) we have \(b(j)=1\). Set \(M=b(i)\); then \(M>1\). This proves part (a).

\medskip
\subsubsection*{(b)  Independence of the final value \(M\)}

For a prime \(p\) and a positive integer \(x\) let \(v_{p}(x)\) be the exponent of \(p\) in the prime factorisation of \(x\) (with \(v_{p}(1)=0\)).  
For a board \(b\) define  

\[
G_{p}(b)=\gcd\{\,v_{p}(b(i))\mid i=1,\dots ,2026\,\}.
\]

(If the set consists only of zeros we take the gcd to be \(0\).)

\textbf{Lemma 2.} \(G_{p}(b)\) is invariant under moves.

\emph{Proof.} Consider a move on positions \(i,j\) with \(m=b(i),\;n=b(j)\). Put \(a=v_{p}(m),\;b=v_{p}(n)\). Then  

\[
v_{p}(\gcd(m,n))=\min(a,b),\qquad 
v_{p}(\operatorname{lcm}(m,n))=\max(a,b).
\]

Since \(h=\operatorname{lcm}(m,n)/\gcd(m,n)\),

\[
v_{p}(h)=v_{p}(\operatorname{lcm}(m,n))-v_{p}(\gcd(m,n))
      =\max(a,b)-\min(a,b)=|a-b|.
\]

Thus the multiset of \(p\)-adic valuations changes from \(\{a,b\}\) to \(\{\min(a,b),|a-b|\}\).  
Let \(R\) be the multiset of valuations of the other \(2024\) positions. Then

\[
\gcd\bigl(\{a,b\}\cup R\bigr)=\gcd\bigl(\gcd(a,b),\gcd(R)\bigr),
\]
\[
\gcd\bigl(\{\min(a,b),|a-b|\}\cup R\bigr)=\gcd\bigl(\gcd(\min(a,b),|a-b|),\gcd(R)\bigr).
\]

Hence it suffices to prove  

\[
\gcd(\min(a,b),|a-b|)=\gcd(a,b). \tag{*}
\]

Assume without loss of generality that \(a\ge b\). Then \(\min(a,b)=b\) and \(|a-b|=a-b\).  
If \(d\) divides both \(b\) and \(a-b\), then \(d\) divides \(b+(a-b)=a\); conversely, if \(d\) divides both \(b\) and \(a\), then \(d\) divides \(a-b\). Thus the sets of common divisors of \(\{b,a-b\}\) and \(\{a,b\}\) coincide, so their greatest common divisors are equal. Hence \((*)\) holds, and the gcd of the whole multiset is unchanged. \hfill\(\square\)

Now look at the terminal board: it contains one number \(M>1\) and the remaining \(2025\) numbers equal to \(1\). The multiset of valuations is \(\{v_{p}(M)\}\cup\{0,0,\dots ,0\}\) (2025 zeros). The gcd of this set is \(v_{p}(M)\) because \(\gcd(v_{p}(M),0)=v_{p}(M)\). Therefore  

\[
G_{p}(\text{final})=v_{p}(M).
\]

By Lemma 2, \(G_{p}(\text{final})=G_{p}(\text{initial})\) for every prime \(p\). Consequently  

\[
v_{p}(M)=G_{p}(\text{initial})\qquad\text{for all primes }p.
\]

The prime factorisation of an integer is unique, so  

\[
M=\prod_{p\text{ prime}} p^{\,G_{p}(\text{initial})}.
\]

Since the right-hand side depends only on the initial multiset, the value of \(M\) is independent of the choices made during the process. \hfill\(\square\)

 \[
  \boxed{
  \begin{minipage}{0.88\textwidth}
  \centering
  The process terminates with exactly one integer \(M>1\) on the board, and
  \[
  M=\prod_p p^{\,g_p},
  \qquad
  g_p=\gcd\!\bigl(v_p(a_1),\ldots,v_p(a_{2026})\bigr),
  \]
  which is independent of the choices.
  \end{minipage}
  }
  \]
\end{pipelinebox}

\clearpage
\subsubsection{IMO Problem 2}

\begin{pipelinebox}[colframe=black!60,colbacktitle=black!8]{Question}

Let \(ABC\) be a triangle, and let \(M\) and \(N\) be the midpoints of \(AB\) and \(AC\), respectively. Let \(K\) and \(L\) be chosen strictly inside triangles \(BMC\) and \(BNC\), respectively, such that \(K\) lies strictly inside \(\angle LBA\), \(L\) lies strictly inside \(\angle ACK\), and
\[
\angle KBA=\angle ACL,\qquad
\angle LBK=\angle LNC,\qquad
\angle LCK=\angle BMK.
\]
If \(O\) is the circumcentre of triangle \(AKL\), prove that \(OM=ON\).

\end{pipelinebox}

\begin{pipelinebox}[colframe=blue!60!black,colbacktitle=blue!8]{K2-Horizon-7B generation}

Place triangle \(ABC\) in the plane with \(A\) at the origin.  Let \(B=(b_{1},b_{2})\) and \(C=(c_{1},c_{2})\).  Since \(\triangle ABC\) is non-degenerate, the vectors \(B\) and \(C\) are linearly independent.  Define  

\[
\begin{aligned}
D&=\det(B,C)=b_{1}c_{2}-b_{2}c_{1}\neq 0,
&\qquad r&=|B|^{2}=b_{1}^{2}+b_{2}^{2},\\
s&=B\!\cdot\!C=b_{1}c_{1}+b_{2}c_{2},
&\qquad t&=|C|^{2}=c_{1}^{2}+c_{2}^{2}.
\end{aligned}
\]

For any points \(P,Q\) we write \(P-Q\) for the vector from \(Q\) to \(P\), and for vectors \(u,v\) we denote by \(\det(u,v)\) the determinant (cross product) and by \(u\!\cdot\!v\) the dot product.

\medskip

\subsubsection*{1.  Coordinates of \(K\) and \(L\)}

Because \(K\) lies inside \(\triangle BMC\) there exist positive numbers \(\alpha,\beta,\gamma\) with \(\alpha+\beta+\gamma=1\) such that  

\[
K=\alpha B+\beta M+\gamma C .
\]

The midpoint \(M\) of \(AB\) is \(M=\frac12(A+B)=\frac12B\); hence  

\[
K=\Bigl(\alpha+\frac{\beta}{2}\Bigr)B+\gamma C .
\]

Set  

\[
x=\alpha+\frac{\beta}{2},\qquad y=\gamma .
\]

Then \(K=xB+yC\).  From \(\alpha,\beta,\gamma>0\) and \(\alpha+\beta+\gamma=1\) we obtain  

\[
0<x<1,\qquad 0<y<1,\qquad x+y=\alpha+\frac{\beta}{2}+\gamma=1-\frac{\beta}{2}<1 .
\]

Similarly, \(L\) lies inside \(\triangle BNC\).  Its midpoint \(N\) is \(N=\frac12C\); thus there exist positive \(\alpha',\beta',\gamma'\) with \(\alpha'+\beta'+\gamma'=1\) such that  

\[
L=\alpha' B+\beta' N+\gamma' C=\alpha' B+\Bigl(\frac{\beta'}{2}+\gamma'\Bigr)C .
\]

Set  

\[
p=\alpha',\qquad q=\frac{\beta'}{2}+\gamma' .
\]

Then \(L=pB+qC\) and  

\[
0<p<1,\qquad 0<q<1,\qquad p+q=1-\frac{\beta'}{2}<1 .
\]

\medskip

\subsubsection*{2.  The conditions ``\(K\) inside \(\angle LBA\)'' and ``\(L\) inside \(\angle ACK\)''}

\(K\) inside \(\angle LBA\) means that the ray \(BK\) lies between \(BA\) and \(BL\).  With \(A=0\) this is equivalent to the existence of positive numbers \(u,v\) such that  

\[
K-B=u(L-B)+v(A-B)=u(L-B)-vB. \tag{*}
\]

Write \(K-B=(x-1)B+yC\) and \(L-B=(p-1)B+qC\).  Substituting into \((*)\) gives  

\[
(x-1)B+yC=u\bigl[(p-1)B+qC\bigr]-vB
      =\bigl[u(p-1)-v\bigr]B+uqC .
\]

Because \(B\) and \(C\) are independent, we equate coefficients:

\[
y=uq,\qquad x-1=u(p-1)-v. \tag{1}
\]

From the first equation \(u=y/q>0\).  Then  

\[
v=u(p-1)-(x-1)=\frac{y}{q}(p-1)-(x-1)
   =\frac{y(p-1)-q(x-1)}{q}.
\]

Define  

\[
d=q(1-x)-y(1-p).
\]

A short calculation shows \(y(p-1)-q(x-1)=d\); hence \(v=d/q\).  Since \(v>0\) and \(q>0\),

\[
d>0. \tag{2}
\]

Similarly, \(L\) inside \(\angle ACK\) means there exist positive \(u',v'\) such that  

\[
L-C=u'(A-C)+v'(K-C)=-u'C+v'(K-C). \tag{**}
\]

Write \(L-C=pB+(q-1)C\) and \(K-C=xB+(y-1)C\).  Substituting into \((**)\) gives  

\[
pB+(q-1)C=v'\bigl[xB+(y-1)C\bigr]-u'C
      =(v'x)B+\bigl[v'(y-1)-u'\bigr]C .
\]

Equating coefficients:

\[
p=v'x,\qquad q-1=v'(y-1)-u'. \tag{3}
\]

From the first equation \(v'=p/x>0\).  Then  

\[
u'=v'(y-1)-(q-1)=\frac{p}{x}(y-1)-(q-1)
   =\frac{p(y-1)-x(q-1)}{x}.
\]

Define  

\[
e=x(1-q)-p(1-y).
\]

Again \(p(y-1)-x(q-1)=e\); hence \(u'=e/x\).  Since \(u'>0\) and \(x>0\),

\[
e>0. \tag{4}
\]

\medskip

\subsubsection*{3.  A lemma on angle equalities}

\textbf{Lemma.}  Let \(u,v,u',v'\) be non-zero vectors in the plane.  If the (undirected) angle between \(u\) and \(v\) equals the angle between \(u'\) and \(v'\) and \(\det(u,v)\det(u',v')>0\), then  

\[
\det(u,v)\,(u'\!\cdot\!v')=\det(u',v')\,(u\!\cdot\!v). \tag{*}
\]

\emph{Proof.}  Equality of the undirected angles gives equality of the cosines:

\[
\frac{u\!\cdot\!v}{|u||v|}=\frac{u'\!\cdot\!v'}{|u'||v'|}.
\]

The absolute values of the determinants satisfy  

\[
\frac{|\det(u,v)|}{|u||v|}=\frac{|\det(u',v')|}{|u'||v'|}.
\]

Because the determinants have the same sign,  

\[
\frac{\det(u,v)}{|u||v|}=\frac{\det(u',v')}{|u'||v'|}.
\]

Multiplying the two equalities yields \(\star\). \hfill\(\square\)

\medskip

\subsubsection*{4.  Translating the angle equalities}

We apply the lemma to each given angle equality.  First we compute the relevant determinants and dot products.

\emph{\(\angle KBA\) = \(\angle ACL\).}  
Take \(u=K-B,\;v=A-B=-B,\;u'=A-C=-C,\;v'=L-C\).  Then  

\[
\det(K-B,-B)=yD,\qquad \det(-C,L-C)=pD .
\]

Since \(y>0,\;p>0\) and \(D\neq0\), the product is positive.  The dot products are  

\[
(K-B)\!\cdot\!(-B)=(1-x)r-ys,\qquad (-C)\!\cdot\!(L-C)=t(1-q)-ps .
\]

Lemma \(\star\) gives  

\[
yD\bigl(t(1-q)-ps\bigr)=pD\bigl((1-x)r-ys\bigr).
\]

Cancelling \(D\) and simplifying we obtain  

\[
rp(1-x)=ty(1-q). \tag{I}
\]

\emph{\(\angle LBK\) = \(\angle LNC\).}  
Take \(u=L-B,\;v=K-B,\;u'=L-N,\;v'=C-N\).  Then  

\[
\det(L-B,K-B)=dD,\qquad \det(L-N,C-N)=\frac{p}{2}D .
\]

Since \(d>0,\;p>0\) the product is positive.  The dot products are  

\[
\begin{aligned}
(L-B)\!\cdot\!(K-B)&=r(1-p)(1-x)\\
&\quad-s\bigl((1-p)y+q(1-x)\bigr)+tqy,\\
(L-N)\!\cdot\!(C-N)&=\frac{sp+t(q-\tfrac12)}{2}.
\end{aligned}
\]

Lemma \(\star\) yields  

\[
\begin{aligned}
dD\cdot\frac{sp+t(q-\tfrac12)}{2}
={}&\frac{p}{2}D\cdot\Bigl[r(1-p)(1-x)\\
&\quad-s\bigl((1-p)y+q(1-x)\bigr)+tqy\Bigr].
\end{aligned}
\]

Cancelling \(D\) and multiplying by \(2\) we get  

\[
\begin{aligned}
d\bigl(sp+t(q-\tfrac12)\bigr)
={}&p\Bigl[r(1-p)(1-x)\\
&\quad-s\bigl((1-p)y+q(1-x)\bigr)+tqy\Bigr].
\end{aligned}
\tag{II}
\]

\emph{\(\angle LCK\) = \(\angle BMK\).}  
Take \(u=L-C,\;v=K-C,\;u'=B-M,\;v'=K-M\).  Then  

\[
\det(L-C,K-C)=eD,\qquad \det(B-M,K-M)=\frac{y}{2}D .
\]

Since \(e>0,\;y>0\) the product is positive.  The dot products are  

\[
\begin{aligned}
(L-C)\!\cdot\!(K-C)&=rpx-s\bigl(p(1-y)+x(1-q)\bigr)\\
&\quad+t(1-q)(1-y),\\
(B-M)\!\cdot\!(K-M)&=\frac{r(x-\tfrac12)+sy}{2}.
\end{aligned}
\]

Lemma \(\star\) gives  

\[
\begin{aligned}
eD\cdot\frac{r(x-\tfrac12)+sy}{2}
={}&\frac{y}{2}D\cdot\Bigl[rpx-s\bigl(p(1-y)+x(1-q)\bigr)\\
&\quad+t(1-q)(1-y)\Bigr].
\end{aligned}
\]

Cancelling \(D\) and multiplying by \(2\) we obtain  

\[
\begin{aligned}
e\bigl(r(x-\tfrac12)+sy\bigr)
={}&y\Bigl[rpx-s\bigl(p(1-y)+x(1-q)\bigr)\\
&\quad+t(1-q)(1-y)\Bigr].
\end{aligned}
\tag{III}
\]

\medskip

\subsubsection*{5.  A crucial algebraic identity}

Set  

\[
\begin{aligned}
E_{1}&=rp(1-x)-ty(1-q)=0
&&\text{(from (I))},\\
E_{2}&=p\bigl[r(1-p)(1-x)-s\bigl((1-p)y+q(1-x)\bigr)+tqy\bigr]\\
&\quad-d\bigl(sp+t(q-\tfrac12)\bigr)=0
&&\text{(from (II))},\\
E_{3}&=y\bigl[rpx-s\bigl(p(1-y)+x(1-q)\bigr)+t(1-q)(1-y)\bigr]\\
&\quad-e\bigl(r(x-\tfrac12)+sy\bigr)=0
&&\text{(from (III))}.
\end{aligned}
\]

Define  

\[
c_{1}=(x+y)(p+2q-1)-(p+q)(1-x).
\]

Consider the linear combination  

\[
F=c_{1}E_{1}+2(x+y)(1-q)E_{2}-2(p+q)(1-x)E_{3}.
\]

Since \(E_{1},E_{2},E_{3}\) are zero, we have \(F=0\).  Now we expand \(F\).  Substituting the expressions for \(d\) and \(e\),

\[
d=q(1-x)-y(1-p),\qquad e=x(1-q)-p(1-y),
\]

and expanding, we obtain after collecting terms:

\[
\begin{aligned}
F=(1-x)(1-q)\Bigl[{}
&r\bigl(2(p+q)x^{2}-2(x+y)p^{2}-(xq-yp)\bigr)\\
&+s\bigl(4(p+q)xy-4(x+y)pq\bigr)\\
&+t\bigl(2(p+q)y^{2}-2(x+y)q^{2}+(xq-yp)\bigr)\Bigr].
\end{aligned}
\]

Because \(0<x<1\) and \(0<q<1\), we have \((1-x)(1-q)\neq0\).  Hence the bracket must vanish, which gives

\[
\begin{aligned}
&2(p+q)(r x^{2}+2sxy+t y^{2})\\
&\quad-2(x+y)(r p^{2}+2spq+t q^{2})=(xq-yp)(r-t).
\end{aligned}
\tag{IV}
\]

\medskip

\subsubsection*{6.  The circumcenter \(O\) of \(\triangle AKL\)}

Since \(A\) is the origin, the condition \(OA=OK=OL\) is equivalent to  

\[
|O|^{2}=|O-K|^{2}=|O-L|^{2}.
\]

Expanding the squares gives  

\[
2\,O\!\cdot\!K=|K|^{2},\qquad 2\,O\!\cdot\!L=|L|^{2}.
\]

Using \(K=xB+yC\) and \(L=pB+qC\),

\begin{align}
x\,(O\!\cdot\!B)+y\,(O\!\cdot\!C)&=\frac{r x^{2}+2s x y+t y^{2}}{2}, \tag{5}\\
p\,(O\!\cdot\!B)+q\,(O\!\cdot\!C)&=\frac{r p^{2}+2s p q+t q^{2}}{2}. \tag{6}
\end{align}

Let \(u=O\!\cdot\!B\) and \(v=O\!\cdot\!C\).  Then (5) and (6) are a linear system for \(u,v\).

We must prove \(OM=ON\).  Because \(M=\frac12B\) and \(N=\frac12C\),

\[
\begin{aligned}
OM^{2}&=|O-\tfrac12B|^{2}=|O|^{2}-u+\frac{r}{4},\\
ON^{2}&=|O-\tfrac12C|^{2}=|O|^{2}-v+\frac{t}{4}.
\end{aligned}
\]

Thus \(OM=ON\) is equivalent to \(OM^{2}=ON^{2}\), i.e.  

\[
v-u=\frac{t-r}{4}. \tag{7}
\]

To obtain (7) from (5) and (6), eliminate \(u\) and \(v\).  Multiply (5) by \(p+q\) and (6) by \(x+y\) and subtract:

\[
\begin{aligned}
&(p+q)(xu+yv)-(x+y)(pu+qv)\\
&\quad=\frac12\Bigl[(p+q)(r x^{2}+2sxy+t y^{2})\\
&\qquad\quad-(x+y)(r p^{2}+2spq+t q^{2})\Bigr].
\end{aligned}
\]

The left-hand side simplifies to  

\[
(p y-x q)(v-u).
\]

Hence  

\[
\begin{aligned}
(py-xq)(v-u)
={}&\frac12\Bigl[(p+q)(r x^{2}+2sxy+t y^{2})\\
&\qquad-(x+y)(r p^{2}+2spq+t q^{2})\Bigr].
\end{aligned}
\tag{8}
\]

Now use (IV).  Dividing (IV) by \(2\) gives  

\[
\begin{aligned}
&(p+q)(r x^{2}+2sxy+t y^{2})\\
&\quad-(x+y)(r p^{2}+2spq+t q^{2})
=\frac{(xq-yp)(r-t)}{2}.
\end{aligned}
\]

Substituting this into (8) yields  

\[
(p y-x q)(v-u)=\frac{(x q-y p)(r-t)}{4}.
\]

Since \(p y-x q=-(x q-y p)\), we have  

\[
-(x q-y p)(v-u)=\frac{(x q-y p)(r-t)}{4}.
\]

The triangle \(AKL\) is non-degenerate (its circumcenter \(O\) exists), therefore \(K\) and \(L\) are not collinear with \(A\).  In our coordinates this means  

\[
\det(K,L)=(x q-y p)D\neq0,
\]

and because \(D\neq0\) we have \(x q-y p\neq0\).  Cancelling this non-zero factor gives  

\[
-(v-u)=\frac{r-t}{4}\quad\Longrightarrow\quad v-u=\frac{t-r}{4},
\]

which is exactly (7).  Hence \(OM^{2}=ON^{2}\), and since distances are non-negative, \(OM=ON\).

\hfill\(\square\)
\end{pipelinebox}

\clearpage
\subsubsection{IMO Problem 3}
\begin{pipelinebox}[colframe=black!60,colbacktitle=black!8]{Question}
Let \(n\) be a positive integer. Liu Bang and Xiang Yu have a stick of length \(1\). First Liu Bang marks at most \(n\) points on the stick, and then Xiang Yu marks at most \(n\) further points; all marked points are distinct. The stick is cut at every marked point. The players then alternately claim unclaimed pieces, with Liu Bang moving first, and each player seeks to maximize the total length of the pieces he claims. Determine the largest number \(c\) that Liu Bang can guarantee, regardless of Xiang Yu's play.
\end{pipelinebox}
\begin{pipelinebox}[colframe=blue!60!black,colbacktitle=blue!8]{K2-Horizon-7B generation}
Let the stick be the interval \([0,1]\). Liu marks at most \(n\) points, then Xiang marks at most \(n\) further points; all marked points are distinct. The stick is cut at every marked point, producing pieces. The players then alternately claim unclaimed pieces, Liu moving first, each maximizing total length.

\textbf{Lemma 1 (Optimal play).}  
Let the pieces be sorted in non-increasing order \(y_1\ge y_2\ge\cdots\ge y_m>0\). Under optimal play Liu obtains the pieces at odd positions and Xiang the pieces at even positions. Consequently Liu's total is \(\frac{1+D}{2}\) where  
\[
D = \sum_{i\text{ odd}} y_i - \sum_{i\text{ even}} y_i .
\]
Liu wants to maximise \(D\), Xiang to minimise it.

\emph{Proof.} We prove by induction on \(m\) that the player whose turn it is always takes the largest remaining piece. For \(m=1\) this is trivial. Assume the statement holds for \(m-1\) pieces. For \(m\) pieces, suppose it is the current player's turn. If the player takes \(y_j\) (\(j>1\)), the remaining pieces are sorted descending with \(y_j\) removed. By induction the opponent will obtain the odd-indexed pieces in that sorted list. Let \(G_j\) be the opponent's total if the current player takes \(y_j\). For \(j=1\), \(G_1 = y_2+y_4+\cdots\). For \(j>1\), removing \(y_j\) shifts the parity of indices for elements after \(j\); the opponent takes \(y_i\) for \(i<j\) with \(i\) odd, and for \(i>j\) with \(i\) even. Hence  
\[
G_j = \sum_{\substack{i<j\\ i\text{ odd}}} y_i \;+\; \sum_{\substack{i>j\\ i\text{ even}}} y_i .
\]
Then  
\[
G_j - G_1 = \Bigl(\sum_{\substack{i<j\\ i\text{ odd}}} y_i + \sum_{\substack{i>j\\ i\text{ even}}} y_i\Bigr) - \sum_{i\text{ even}} y_i .
\]
Write \(\sum_{i\text{ even}} y_i = \sum_{\substack{i<j\\ i\text{ even}}} y_i + (y_j\text{ if }j\text{ is even else }0) + \sum_{\substack{i>j\\ i\text{ even}}} y_i\). Thus  
\[
G_j - G_1 = \sum_{\substack{i<j\\ i\text{ odd}}} y_i - \sum_{\substack{i<j\\ i\text{ even}}} y_i - (y_j\text{ if }j\text{ is even else }0)
= \sum_{\substack{i<j\\ i\text{ odd}}} y_i - \sum_{\substack{i\le j\\ i\text{ even}}} y_i .
\]
Now consider the cases.  
\noindent\textbullet\ If \(j = 2h\) is even, then  
\(\sum_{i<2h, i\text{ odd}} y_i = y_1+y_3+\cdots+y_{2h-1} = \sum_{r=1}^h y_{2r-1}\),  
\(\sum_{i\le 2h, i\text{ even}} y_i = y_2+y_4+\cdots+y_{2h} = \sum_{r=1}^h y_{2r}\).  
Hence \(G_j - G_1 = \sum_{r=1}^h (y_{2r-1}-y_{2r}) \ge 0\).  
\noindent\textbullet\ If \(j = 2h+1\) is odd, then  
\(\sum_{i<2h+1, i\text{ odd}} y_i = y_1+y_3+\cdots+y_{2h-1} = \sum_{r=1}^h y_{2r-1}\),  
\(\sum_{i\le 2h+1, i\text{ even}} y_i = y_2+y_4+\cdots+y_{2h} = \sum_{r=1}^h y_{2r}\).  
Hence \(G_j - G_1 = \sum_{r=1}^h (y_{2r-1}-y_{2r}) \ge 0\).  
In both cases \(G_j \ge G_1\), so taking \(y_1\) minimises the opponent's total and maximises the current player's total. \hfill\(\square\)

\textbf{Lemma 2 (Pairs-plus-one).}  
Let \(S\) be a multiset of positive real numbers. Suppose \(S\) consists of some number of equal pairs (each pair consists of two equal numbers) plus possibly one extra element \(r>0\). (If \(r=0\) we may ignore it.) Then when \(S\) is sorted in non-increasing order, the alternating sum  
\[
D = \sum_{\text{odd } i} s_i - \sum_{\text{even } i} s_i
\]
equals \(r\).

\emph{Proof.} Sort \(S\) descending. Group the elements by their values. Since all numbers except possibly \(r\) appear in pairs, every value block (except possibly the block containing \(r\)) has even size. The block containing \(r\) has odd size. The number of elements before this block is the sum of the sizes of the previous blocks, which are all even, so it is even. In the alternating sum, each block of even size contributes \(0\) (the alternating sum of an even number of equal terms cancels). The block containing \(r\) starts at an odd position (because the number of preceding elements is even), so its terms alternate signs starting with \(+\). The sum of an odd number of equal terms with alternating signs starting with \(+\) is the value itself. Hence the contribution of this block is \(r\). All other blocks contribute \(0\). Thus \(D=r\). If \(r=0\) then \(D=0\). \hfill\(\square\)

\medskip

\subsubsection*{1. Lower bound -- Liu can guarantee \(D\ge\delta\)}

Set \(\delta = \dfrac{1}{2^{\,n+1}-1}\).  
Liu marks the points \((2^k-1)\delta\) for \(k=1,2,\dots,n\). The stick is divided into \(n+1\) intervals  
\[
I_k = \bigl[(2^{k-1}-1)\delta,\;(2^k-1)\delta\bigr],\qquad k=1,\dots,n+1,
\]
with lengths \(l_k = 2^{\,k-1}\delta\). Their sum is \(\delta(2^{n+1}-1)=1\).

Xiang marks at most \(n\) further points. The stick is cut at every marked point, producing pieces. Let \(m\) be the number of positive pieces. Since there are at most \(2n\) marks, \(m\le 2n+1\). Sort the pieces in non-increasing order and pad with zeros to obtain a list of length \(2n+2\) (if \(m<2n+1\) add zeros; if \(m=2n+1\) add one zero). Pair the list as \((a_1,b_1),\dots,(a_{n+1},b_{n+1})\) with \(a_i\ge b_i\ge 0\). Associate each piece with its parent interval \(I_k\); zero pieces are associated with a dummy vertex \(D\).

Construct a directed multigraph \(G\): vertices are \(I_1,\dots,I_{n+1}\) and \(D\). For each pair \((a_i,b_i)\) draw a directed edge from the parent of \(a_i\) to the parent of \(b_i\). Thus \(G\) has \(V=n+2\) vertices and \(E=n+1\) edges. Every \(I_k\) contains at least one piece, and \(D\) is incident to at least one zero piece (since exactly \(2n+2-m\ge 1\) zeros were added), so every vertex has degree \(\ge 1\).

Consider the undirected version of \(G\). For any connected component \(C\), we have \(E_C\ge V_C-1\) (a connected graph has at least \(V_C-1\) edges). Summing over all components, \(E\ge V-c\), where \(c\) is the number of components. Since \(E=V-1\), we get \(V-1\ge V-c\), hence \(c\ge 1\). If every component satisfied \(E_C\ge V_C\), then \(E\ge V\), contradicting \(E=V-1\). Therefore there exists at least one component \(T\) with \(E_T=V_T-1\). Because every vertex has degree \(\ge 1\), \(T\) contains at least two vertices. A connected multigraph with \(E_T=V_T-1\) has no cycles (any cycle requires at least \(V_T\) edges), so \(T\) is a tree (acyclic, simple). Bipartition this tree \(T\) into \(U\cup W\). Both \(U\) and \(W\) are non-empty.

Define \(P=\{k:I_k\in U\}\) and \(Q=\{k:I_k\in W\}\). The tree has at least two vertices and only one can be the dummy \(D\), so \(P\cup Q\) is nonempty. For a directed edge \(e=(A,B)\) in \(T\) (where \(A\) is the parent of \(a_i\), \(B\) is the parent of \(b_i\)), define \(\operatorname{sgn}(e)=+1\) if \(A\in U,\,B\in W\) and \(-1\) if \(A\in W,\,B\in U\). Let \(d_e = a_i-b_i\) (so \(d_e\ge 0\)). For a vertex \(v\) let \(L(v)\) be the total length of pieces from \(v\) (all pieces from \(v\) are incident to edges in \(T\), because \(v\) belongs to \(T\)).

We claim  
\[
\sum_{e\in T}\operatorname{sgn}(e)\,d_e = \sum_{v\in U}L(v) - \sum_{v\in W}L(v).
\]
Indeed, for an edge \(e=(A,B)\) the contribution is \(\operatorname{sgn}(e)(a_i-b_i)\). If \(A\in U,\,B\in W\) this is \(a_i-b_i\); if \(A\in W,\,B\in U\) it is \(b_i-a_i\). In both cases the piece from the \(U\)-endpoint receives \(+1\) and the piece from the \(W\)-endpoint receives \(-1\). Summing over all edges in \(T\), each piece from \(v\in U\) is counted \(+1\) and each from \(v\in W\) is counted \(-1\). Hence the identity holds.

Now \(L(I_k)=l_k = 2^{\,k-1}\delta\) for \(k=1,\dots,n+1\) and \(L(D)=0\). The difference \(\sum_U L - \sum_W L\) is a sum of some \(l_k\) with coefficients \(\pm1\). Since the \(l_k\) are distinct powers of two times \(\delta\), any such sum is a multiple of \(\delta\). Moreover, \(P\) and \(Q\) are disjoint and \(P\cup Q\) is nonempty. If \(\sum_{k\in P}2^{k-1}=\sum_{k\in Q}2^{k-1}\), uniqueness of binary expansion gives \(P=Q\); disjointness then gives \(P=Q=\varnothing\), contradicting \(P\cup Q\) nonempty. Therefore the signed sum is nonzero and its absolute value is at least \(\delta\).

Now \(D = \sum_{i=1}^{n+1} (a_i-b_i) = \sum_{e\in E} d_e\). Since \(d_e\ge 0\), we have  
\[
D \ge \sum_{e\in T} d_e \ge \Bigl|\sum_{e\in T}\operatorname{sgn}(e)\,d_e\Bigr|
= \Bigl|\sum_{U}L - \sum_{W}L\Bigr| \ge \delta.
\]

Thus Liu's total is at least \(\frac{1+\delta}{2}\).

\medskip

\subsubsection*{2. Upper bound -- Xiang can force \(D\le\delta\)}

If Liu marks fewer than \(n\) points, he creates at most \(n\) intervals. Xiang bisects each interval (using at most \(n\) cuts); then all pieces come in equal pairs, so \(D=0\le\delta\).

Assume Liu marks exactly \(n\) points, creating \(n+1\) intervals of lengths \(a_0,a_1,\dots,a_n>0\) with \(\sum a_i = 1\).

Consider all subsets of \(\{0,1,\dots,n\}\). There are \(2^{\,n+1}\) subset sums (counted with multiplicity), all in \([0,1]\), including \(0\) and \(1\). Sort them: \(s_1\le s_2\le\cdots\le s_{2^{\,n+1}}\). The \(2^{\,n+1}-1\) consecutive gaps are non-negative and telescope to \(1\), so one gap is at most \(\frac{1}{2^{\,n+1}-1}=\delta\). The two entries bounding this gap come from two distinct subsets (if the gap is zero the subsets are still distinct). Hence there exist distinct subsets \(X,Y\subseteq\{0,\dots,n\}\) with \(|\sum X - \sum Y|\le\delta\).

Among all such pairs, choose one that minimises \(|X|+|Y|\). Let \(A=X,\;B=Y\) with \(\sum A \ge \sum B\) (swap if necessary) and set \(r = \sum A - \sum B\); then \(0\le r\le\delta\).

\textbf{Claim 1.} \(A\cap B = \varnothing\).  
\emph{Proof.} If \(Z=A\cap B\neq\varnothing\), let \(A'=A\setminus Z,\;B'=B\setminus Z\). Then \(A',B'\) are distinct, \(|\sum A'-\sum B'| = r\le\delta\), and \(|A'|+|B'| = |A|+|B|-2|Z| < |A|+|B|\), contradicting minimality. \hfill\(\square\)

Thus \(A,B\) are disjoint. Let \(p=|A|,\;q=|B|\).

\textbf{Case 1: \(B = \varnothing\).}  
Then \(A\) is non-empty and \(\sum A = r\le\delta\). Minimality forces \(|A|=1\). Indeed, if \(|A|\ge2\), choose a singleton \(\{i\}\subset A\). Then \(\sum\{i\} = a_i \le r\le\delta\) and \(|\{i\}|\!+\!|\varnothing| = 1 < |A|\), contradicting minimality. Thus \(A=\{k\}\) for some \(k\) and \(r=a_k\le\delta\).  
Xiang leaves \(I_k\) uncut and bisects all other \(n\) intervals (using \(n\) cuts). The pieces are one piece of length \(r\) and \(2n\) pieces forming \(n\) equal pairs. By Lemma 2, \(D = r \le \delta\).

\textbf{Case 2: \(B \neq \varnothing\).}  
Then \(q\ge 1\). We have \(A,B\) disjoint. Order the intervals in \(A\) by their positions on the stick: lengths \(\alpha_1,\dots,\alpha_p\), cumulative sums \(A_0=0,A_1=\alpha_1,\dots,A_p=\sum A\). Similarly for \(B\): \(\beta_1,\dots,\beta_q\), \(B_0=0,\dots,B_q=\sum B\).

\textbf{Claim 2.} For all \(k\) with \(1\le k < p\), \(A_k \le \sum B\).  
\emph{Proof.} Suppose \(\sum B < A_k < \sum A\) for some \(k<p\). Let \(T\) be the set of indices of \(A\) corresponding to \(\{\alpha_{k+1},\dots,\alpha_p\}\). Then \(\sum T = \sum A - A_k < r\le\delta\) and \(|T| = p-k < p+q\) (since \(q\ge1\)). The pair \((T,\varnothing)\) is disjoint, \(|\sum T-\sum\varnothing| = \sum T \le\delta\), and \(|T|+0 = p-k < p+q\), contradicting minimality. \hfill\(\square\)

Hence \(A_{p-1}\le\sum B\le A_p\).

Now we construct Xiang's marks. The cumulative numbers \(A_j\) and \(B_j\) are coordinates in two separate abstract concatenations, not positions on the original stick. Order the actual selected \(A\)-intervals left-to-right and translate them end-to-end, preserving lengths, to virtual \([0,\sum A]\), where actual interval \(\alpha_j\) represents virtual \([A_{j-1},A_j]\). Separately translate the actual \(B\)-intervals to virtual \([0,\sum B]\).

Retain Claim 2 and the common sorted boundary union \(0=t_0<t_1<\dots<t_s=\sum B\), where  
\[
S = \{0\} \cup \{A_1,\dots,A_{p-1}\} \cup \{B_1,\dots,B_q\}.
\]
(Recall \(A_{p-1}\le\sum B\), so all \(A_j\) for \(j<p\) are \(\le\sum B\), and \(B_q=\sum B\).)

For each \(B\)-boundary \(B_j\) (\(1\le j\le q\)) strictly inside a virtual \(A\)-cell \([A_{k-1},A_k]\), mark the corresponding \textbf{actual} \(A\)-interval at distance \(B_j-A_{k-1}\) from its actual left endpoint. For each \(A\)-boundary \(A_j\) (\(1\le j\le p-1\)) strictly inside a virtual \(B\)-cell \([B_{k-1},B_k]\), mark the corresponding \textbf{actual} \(B\)-interval at distance \(A_j-B_{k-1}\) from its actual left endpoint. Equal boundaries need no cuts. Thus overlay cuts \(\le p+q-1\), and all marks are legal and distinct.

Because \(A\) and \(B\) are disjoint, the \(A\)-intervals and \(B\)-intervals are distinct intervals on the stick. The two abstract concatenations now have the same refinement boundary list up to \(\sum B\), so their resulting \textbf{physical} pieces have paired equal lengths \(t_i-t_{i-1}\); they need not be at the same physical locations.

Because \(A_{p-1}\le\sum B\le A_p\), the virtual tail \([\sum B,\sum A]\) corresponds to one tail in the last actual \(A\)-interval, of length \(r\); when \(r=0\) it is absent.

Bisect all intervals not in \(A\) or \(B\); there are \(n+1-p-q\) of them, using \(n+1-p-q\) cuts. Total cuts used is at most \((p+q-1)+(n+1-p-q) = n\). Hence Xiang can use at most \(n\) cuts.

The final pieces consist of:
\noindent\textbullet\ equal pairs from the overlay construction,
\noindent\textbullet\ equal pairs from the bisected intervals,
\noindent\textbullet\ one extra piece of length \(r\) (if \(r>0\)).

By Lemma 2, the alternating sum \(D\) equals \(r\) (or \(0\) if \(r=0\)). Since \(r\le\delta\), we have \(D\le\delta\).

\medskip

\subsubsection*{3. Conclusion}

Liu can guarantee at least \(\frac{1+\delta}{2}\) and Xiang can force at most \(\frac{1+\delta}{2}\). Therefore the largest number \(c\) Liu Bang can guarantee is  
\[
c = \frac{1+\delta}{2} = \frac{1+\frac{1}{2^{\,n+1}-1}}{2} = \frac{2^{\,n}}{2^{\,n+1}-1}.
\]

\[
\boxed{\dfrac{2^n}{2^{n+1}-1}}
\]
\end{pipelinebox}

\clearpage
\subsubsection{IMO Problem 4}

\begin{pipelinebox}[colframe=black!60,colbacktitle=black!8]{Question}

Shan-Yu and Mulan are playing a game. Let \(\theta\) be an angle satisfying
\[
0^\circ < \theta < 180^\circ,
\]
known to both players. Initially, Shan-Yu constructs a paper triangle \(T\) with measurements of his choice. They then repeatedly perform the following steps:

\begin{enumerate}
    \item If \(T\) has at least one angle measuring exactly \(\theta\), the game ends and Mulan wins.
    \item Otherwise, Mulan chooses a point \(P\) on the perimeter of \(T\), distinct from its three vertices.
    \item She makes a straight cut from \(P\) to the opposite vertex of \(T\), dividing \(T\) into two triangles.
    \item Shan-Yu discards one of the two triangles, and the remaining triangle becomes the new \(T\).
\end{enumerate}

For which real values of \(\theta\) can Mulan guarantee victory in finitely many steps, regardless of how Shan-Yu plays?

\end{pipelinebox}

\begin{pipelinebox}[colframe=blue!60!black,colbacktitle=blue!8]{K2-Horizon-7B generation}

Let \(\theta\) be an angle with \(0^\circ<\theta<180^\circ\).  A triangle is determined by its three positive angles \(A,B,C\) with \(A+B+C=180^\circ\).  
A cut at vertex \(A\) (with angles \(A,B,C\)) means choosing a point \(P\) on side \(BC\) (different from \(B,C\)) and drawing segment \(AP\).  This splits the triangle into two triangles with angles  

\[
(x,\;B,\;180^\circ-x-B)\qquad\text{and}\qquad(A-x,\;C,\;180^\circ-(A-x)-C)
\]

for some \(x\) with \(0<x<A\).  (The angles at \(B\) and \(C\) remain unchanged, and the new angles at \(A\) are \(x\) and \(A-x\); the third angles are forced by the sum \(180^\circ\).)

We determine for which \(\theta\) Mulan has a strategy that guarantees victory in finitely many steps, no matter how Shan-Yu plays.

\medskip
\subsubsection*{1.  Necessity:  if \(\theta\) does \textbf{not} divide \(180^\circ\), Mulan cannot force a win.}

Assume \(\theta\nmid 180^\circ\); then there is no integer \(k\) with \(180^\circ=k\theta\).

\textbf{Lemma 1.}  Let \(T\) be a triangle with angles \(A,B,C\), none of which is an integer multiple of \(\theta\).  For any cut of \(T\), at most one of the two resulting triangles has an angle that is an integer multiple of \(\theta\).

\emph{Proof.}  Suppose we cut at vertex \(A\), splitting \(A\) into \(x\) and \(A-x\) (\(0<x<A\)).  The two triangles are  

\[
T_1=(x,\;B,\;180^\circ-x-B),\qquad 
T_2=(A-x,\;C,\;180^\circ-(A-x)-C).
\]

Assume, for contradiction, that both \(T_1\) and \(T_2\) contain an angle that is an integer multiple of \(\theta\).  We examine the possibilities.

\noindent\textbullet\ If \(T_1\)'s multiple is \(B\) or \(T_2\)'s multiple is \(C\), then \(T\) itself has a multiple, contradicting the hypothesis.
\noindent\textbullet\ If \(T_1\)'s multiple is \(x\) and \(T_2\)'s multiple is \(A-x\), then \(A=x+(A-x)\) is a multiple of \(\theta\), contradiction.
\noindent\textbullet\ If \(T_1\)'s multiple is \(x\) and \(T_2\)'s multiple is the third angle \(180^\circ-(A-x)-C\), then  
  \[
  180^\circ-(A-x)-C = B+x \qquad(\text{since }A+B+C=180^\circ).
  \]
  Hence \(B+x\) is a multiple of \(\theta\); because \(x\) is a multiple of \(\theta\), \(B\) is also a multiple of \(\theta\), contradiction.
\noindent\textbullet\ If \(T_1\)'s multiple is the third angle \(180^\circ-x-B\) and \(T_2\)'s multiple is \(A-x\), then  
  \[
  180^\circ-x-B = (A-x)+C \qquad(\text{again using }A+B+C=180^\circ),
  \]
  so \(C\) is a multiple of \(\theta\), contradiction.
\noindent\textbullet\ If both multiples are the third angles, then  
  \[
  (180^\circ-x-B)+(180^\circ-(A-x)-C)=360^\circ-(A+B+C)=180^\circ,
  \]
  which gives \((p+q)\theta=180^\circ\) for some integers \(p,q\); this contradicts \(\theta\nmid 180^\circ\).

All cases lead to a contradiction, so at most one of \(T_1,T_2\) can have a multiple of \(\theta\). \hfill\(\square\)

\textbf{Shan-Yu's strategy.}  Choose an initial triangle with no angle a multiple of \(\theta\); for instance the equilateral triangle with angles \(60^\circ,60^\circ,60^\circ\).  (If \(60^\circ\) were a multiple of \(\theta\), then \(180^\circ=3\cdot60^\circ\) would be a multiple of \(\theta\), contradicting \(\theta\nmid 180^\circ\).)  

Whenever Mulan makes a cut, Shan-Yu discards the triangle that contains a multiple of \(\theta\) (if such a triangle exists); otherwise he discards arbitrarily.  By Lemma 1 the remaining triangle has no multiple of \(\theta\).  Hence the game never produces an angle equal to \(\theta\), and Mulan never wins.  Thus Mulan cannot guarantee victory when \(\theta\nmid 180^\circ\).

\medskip
\subsubsection*{2.  Sufficiency:  if \(\theta\) divides \(180^\circ\), Mulan has a winning strategy.}

Now assume \(180^\circ=m\theta\) for some integer \(m\ge 2\) (because \(\theta<180^\circ\) gives \(m>1\)).  We describe Mulan's strategy.

\textbf{Strategy.}  Given the current triangle \(T\) with angles \(A,B,C\).

\noindent\textbullet\ If any angle equals \(\theta\), Mulan wins immediately.
\noindent\textbullet\ Otherwise, we distinguish two cases.

  \textbf{Case 1.}  \(T\) has an angle that is a positive integer multiple of \(\theta\), say \(A=q\theta\) with \(q\ge 2\).  
  Mulan cuts at vertex \(A\), splitting it into \(\theta\) and \((q-1)\theta\).  The two resulting triangles are  

  \[
  T_1=(\theta,\;B,\;A+C-\theta),\qquad 
  T_2=((q-1)\theta,\;C,\;B+\theta).
  \]

  (One checks that \(A+C-\theta=180^\circ-B-\theta\) and \(B+\theta=180^\circ-C-(q-1)\theta\), so the angles sum to \(180^\circ\) in each triangle.)  
  \(T_1\) contains \(\theta\), so Shan-Yu will discard \(T_1\).  The remaining triangle is \(T_2\).  
  If \(q=2\), then \(T_2\) contains \(\theta\) and Mulan wins.  
  If \(q>2\), then \(T_2\) has an angle \((q-1)\theta\) which is a positive integer multiple of \(\theta\); we are back to Case 1 with a smaller multiple.  
  Hence after at most \(q-1\) moves Mulan wins.

  \textbf{Case 2.}  \(T\) has \textbf{no} angle that is a positive integer multiple of \(\theta\).  
  Mulan performs a special cut to create two triangles that both have a multiple of \(\theta\).

  \emph{Lemma 2.}  There exist a vertex \(i\in\{0,1,2\}\) and a positive integer \(p\) such that  

  \[
  T(i+1) < p\theta < T(i)+T(i+1) \qquad(\text{indices modulo }3).
  \]

  \emph{Proof of Lemma 2.}  Suppose first that there is a vertex \(i\) with \(T(i)>\theta\).  Let \(p\) be the smallest positive integer with \(p\theta>T(i+1)\) (such \(p\) exists because \(T(i+1)<180^\circ=m\theta\)).  Then \((p-1)\theta\le T(i+1)<p\theta\).  Since \(T(i)>\theta\),  

  \[
  p\theta \le T(i+1)+\theta < T(i+1)+T(i),
  \]
  so \(p\theta<T(i)+T(i+1)\).  The inequality \(T(i+1)<p\theta\) holds by construction.

  If, on the other hand, every angle satisfies \(T(i)\le\theta\), then because no angle equals \(\theta\) we have \(T(i)<\theta\) for all \(i\).  Pick \(i=0\).  Then \(T(1)<\theta\).  Moreover,  
  \[
  T(0)+T(1)=180^\circ-T(2) > 180^\circ-\theta = (m-1)\theta \ge \theta \quad(\text{since }m\ge2).
  \]
  Hence \(T(1)<\theta<T(0)+T(1)\), and we may take \(p=1\). \hfill\(\square\)

  Now let \(i\) and \(p\) be as in Lemma 2.  Set \(x=p\theta-T(i+1)\).  Then \(0<x<T(i)\).  Mulan cuts at vertex \(i\), splitting the angle \(T(i)\) into \(x\) and \(T(i)-x\).  The two resulting triangles are  

  \[
  L=(x,\;T(i+1),\;(m-p)\theta),\qquad 
  R=(T(i)-x,\;T(i+2),\;p\theta).
  \]

  (One verifies the third angles using \(A+B+C=180^\circ\) and \(180^\circ=m\theta\).)  

  Both \(L\) and \(R\) have an angle that is a positive integer multiple of \(\theta\).  Shan-Yu discards one of them.  The remaining triangle has an angle that is a positive integer multiple of \(\theta\).  If that multiple is \(\theta\), Mulan wins immediately.  If it is larger, say \(q\theta\) with \(q\ge2\), Mulan proceeds to Case 1, which will finish in at most \(q-1\) moves.  Because \(p\le m-1\) and the remaining multiple is either \(p\) or \(m-p\), we have \(q\le m-1\).  Thus after the initial cut Mulan wins in at most \(m-1\) further moves, i.e. in at most \(m\) moves altogether.

  This completes the description of Mulan's strategy.

\medskip
\subsubsection*{3.  Conclusion}

Mulan can guarantee her victory in finitely many steps if and only if \(\theta\) divides \(180^\circ\), i.e.  

\[
\boxed{\theta = \frac{180^\circ}{k}\quad\text{for some integer }k\ge 2}.
\]
\end{pipelinebox}

\clearpage
\subsubsection{IMO Problem 5}

\begin{pipelinebox}[colframe=black!60,colbacktitle=black!8]{Question}

Determine all functions \(f:\mathbb{R}_{>0}\to\mathbb{R}_{>0}\) such that \(\sqrt{(x^2+f(y)^2)/2}\ge (f(x)+y)/2\ge\sqrt{x f(y)}\) for every \(x,y\in\mathbb{R}_{>0}\).

\end{pipelinebox}

\begin{pipelinebox}[colframe=blue!60!black,colbacktitle=blue!8]{K2-Horizon-7B generation}

Let \(f:(0,\infty)\to(0,\infty)\) satisfy  

\[
\sqrt{\frac{x^{2}+f(y)^{2}}{2}}\;\ge\;\frac{f(x)+y}{2}\;\ge\;\sqrt{x\,f(y)}\qquad(\forall x,y>0).
\]

All quantities are positive, so squaring preserves the order.  
The left inequality becomes  

\[
\frac{x^{2}+f(y)^{2}}{2}\ge\Bigl(\frac{f(x)+y}{2}\Bigr)^{2}
\;\Longleftrightarrow\;
2x^{2}+2f(y)^{2}\ge (f(x)+y)^{2},
\]

and the right inequality becomes  

\[
\Bigl(\frac{f(x)+y}{2}\Bigr)^{2}\ge x\,f(y)
\;\Longleftrightarrow\;
(f(x)+y)^{2}\ge 4x\,f(y).
\]

Thus for all \(x,y>0\)

\[
2x^{2}+2f(y)^{2}\;\ge\;(f(x)+y)^{2}\;\ge\;4x\,f(y). \tag{1}
\]

\medskip

\subsubsection*{1.  A functional equation  }

Put \(x=f(y)\) in (1).  Because \(f(y)>0\),

\[
2f(y)^{2}+2f(y)^{2}\ge (f(f(y))+y)^{2}\ge 4f(y)^{2},
\]

hence \((f(f(y))+y)^{2}=4f(y)^{2}\).  Since \(f(f(y))+y>0\),

\[
f(f(y))+y=2f(y)\qquad(\forall y>0),
\]

or  

\[
f(f(y))=2f(y)-y. \tag{2}
\]

\medskip

\subsubsection*{2.  Introducing \(g\)}

Define \(g(y)=f(y)-y\) for \(y>0\); then \(f(y)=y+g(y)\).  
From (2),

\[
g(f(y))=f(f(y))-f(y)=(2f(y)-y)-f(y)=f(y)-y=g(y). \tag{3}
\]

\medskip

\subsubsection*{3.  Iteration and non-negativity of \(g\)}

We prove by induction that for every \(y>0\) and every \(n\in\mathbb N\),

\[
g(y+n\,g(y))=g(y)\qquad\text{and}\qquad y+n\,g(y)>0. \tag{4}
\]

\emph{Base \(n=0\):}  \(g(y)=g(y)\) and \(y>0\).  
\emph{Inductive step:}  Assume (4) holds for some \(n\).  Then  

\[
f(y+n\,g(y))=(y+n\,g(y))+g(y+n\,g(y))=y+(n+1)g(y).
\]

Because \(y+n\,g(y)>0\), the hypothesis \(f(z)>0\) for \(z>0\) gives  
\(f(y+n\,g(y))>0\); hence \(y+(n+1)g(y)>0\).  Moreover,

\[
g(y+(n+1)g(y))=g\bigl(f(y+n\,g(y))\bigr)=g(y+n\,g(y))=g(y),
\]

where we used (3) and the induction hypothesis.  Thus (4) holds for \(n+1\).

Now suppose \(g(y)<0\) for some \(y>0\).  Then \(-g(y)>0\).  Choose \(n\in\mathbb N\) with  
\(n>\dfrac{y}{-g(y)}\).  Then  

\[
y+n\,g(y)=y-n(-g(y))<0,
\]

contradicting the second part of (4).  Therefore  

\[
g(y)\ge 0\qquad(\forall y>0). \tag{5}
\]

Consequently \(f(y)=y+g(y)\ge y\) for all \(y>0\).

\medskip

\subsubsection*{4.  A key inequality  }

For arbitrary \(x,y>0\) set \(\Delta=g(x)-g(y)\).  Expanding the two parts of (1) gives  

\[
\begin{aligned}
2x^{2}+2f(y)^{2}-(f(x)+y)^{2}
&=(x-f(y))^{2}-2\Delta\,(x+f(y))-\Delta^{2},\\[2mm]
(f(x)+y)^{2}-4x\,f(y)
&=(x-f(y))^{2}+2\Delta\,(x+f(y))+\Delta^{2}.
\end{aligned}
\]

Hence (1) yields  

\[
(x-f(y))^{2}-2\Delta\,(x+f(y))-\Delta^{2}\ge 0, \tag{6}
\]
\[
(x-f(y))^{2}+2\Delta\,(x+f(y))+\Delta^{2}\ge 0. \tag{7}
\]

Now observe that  

\[
4f(y)\Delta+\bigl((x-f(y))+\Delta\bigr)^{2}
=(x-f(y))^{2}+2\Delta\,(x+f(y))+\Delta^{2}.
\]

Thus (6) and (7) are exactly the two parts of  

\[
0\le 4f(y)\bigl(g(x)-g(y)\bigr)
+\bigl((x-f(y))+(g(x)-g(y))\bigr)^{2}
\le 2(x-f(y))^{2}. \tag{8}
\]

So (8) holds for all \(x,y>0\).

\medskip

\subsubsection*{5.  Bounding the difference of \(g\)}

Let \(u,v>0\).  We prove  

\[
4\min(u,v)\,|g(u)-g(v)|\le (u-v)^{2}. \tag{9}
\]

Assume first that \(g(v)\le g(u)\).  Put \(x=f(u),\;y=v\) in (8).  Using (3),

\[
\begin{aligned}
0\le{}&4f(v)\bigl(g(f(u))-g(v)\bigr)\\
&+\bigl((f(u)-f(v))+(g(f(u))-g(v))\bigr)^{2}\\
&\le 2(f(u)-f(v))^{2}.
\end{aligned}
\]

Since \(g(f(u))=g(u)\) and \(f(v)=v+g(v)\), the right-hand inequality becomes  

\[
\begin{aligned}
&4(v+g(v))(g(u)-g(v))\\
&\quad+\bigl((f(u)-f(v))+(g(u)-g(v))\bigr)^{2}\\
&\quad\le 2(f(u)-f(v))^{2}.
\end{aligned}
\]

Now \(f(u)=u+g(u),\;f(v)=v+g(v)\), so  

\[
f(u)-f(v)=(u-v)+(g(u)-g(v)).
\]

Let \(\Delta=g(u)-g(v)\ge 0\).  Then  

\[
f(u)-f(v)=(u-v)+\Delta,\qquad
(f(u)-f(v))+\Delta=(u-v)+2\Delta.
\]

Substituting these into the inequality and expanding gives  

\[
4v\Delta+4g(v)\Delta+(u-v)^{2}+4\Delta(u-v)+4\Delta^{2}
\le 2(u-v)^{2}+4\Delta(u-v)+2\Delta^{2},
\]

which simplifies to  

\[
4v\Delta+4g(v)\Delta+2\Delta^{2}\le (u-v)^{2}.
\]

Because \(g(v)\ge 0\) and \(\Delta\ge 0\), we have  

\[
4v\Delta\le 4(v+g(v))\Delta\le (u-v)^{2}-2\Delta^{2}\le (u-v)^{2}.
\]

Thus  

\[
4v\,(g(u)-g(v))\le (u-v)^{2}. \tag{10}
\]

If instead \(g(u)\le g(v)\), we interchange the roles of \(u\) and \(v\) (use \(x=f(v),\;y=u\) in (8)) and obtain  

\[
4u\,(g(v)-g(u))\le (u-v)^{2}. \tag{11}
\]

In either case (9) follows.

\medskip

\subsubsection*{6.  \(g\) is constant  }

Suppose there exist \(p,q>0\) with \(g(p)\neq g(q)\).  Without loss of generality, assume \(p<q\).  Put \(E=|g(q)-g(p)|>0\).  Choose a natural number \(n\) such that  

\[
n>\frac{(q-p)^{2}}{4pE}. \tag{12}
\]

Set \(d=\dfrac{q-p}{n}>0\) and define \(z_{k}=p+k d\) for \(k=0,1,\dots,n\).  Then \(z_{n}=q\), and for each \(k\) we have \(z_{k}>0\) and \(\min(z_{k+1},z_{k})=z_{k}\ge p\).

Applying (9) to the pair \((z_{k+1},z_{k})\) yields  

\[
4z_{k}\,|g(z_{k+1})-g(z_{k})|\le (z_{k+1}-z_{k})^{2}=d^{2}.
\]

Since \(z_{k}\ge p\),

\[
4p\,|g(z_{k+1})-g(z_{k})|\le 4z_{k}\,|g(z_{k+1})-g(z_{k})|\le d^{2}.
\]

Summing these inequalities for \(k=0,1,\dots,n-1\) gives  

\[
4p\sum_{k=0}^{n-1}|g(z_{k+1})-g(z_{k})|\le n d^{2}=\frac{(q-p)^{2}}{n}.
\]

By the triangle inequality,

\[
\sum_{k=0}^{n-1}|g(z_{k+1})-g(z_{k})|\ge |g(z_{n})-g(z_{0})|=|g(q)-g(p)|=E.
\]

Consequently  

\[
4pE\le\frac{(q-p)^{2}}{n}.
\]

But (12) implies \(\dfrac{(q-p)^{2}}{n}<4pE\), a contradiction.  Hence our assumption was false, and \(g(p)=g(q)\) for all \(p,q>0\).  Therefore \(g\) is constant.

\medskip

\subsubsection*{7.  Determining \(f\)}

Let \(g(y)=c\) for all \(y>0\).  Then \(f(y)=y+c\).  The condition \(f(y)>0\) for all \(y>0\) forces \(c\ge 0\).  (If \(c<0\), then for \(y=-c/2>0\) we would have \(f(y)=c/2<0\), impossible.)

Conversely, if \(c\ge 0\) and \(f(x)=x+c\), then for any \(x,y>0\),

\[
\sqrt{\frac{x^{2}+(y+c)^{2}}{2}}\;\ge\;\frac{x+y+c}{2}\;\ge\;\sqrt{x(y+c)}.
\]

Indeed, the right inequality is equivalent to \((x+y+c)^{2}\ge 4x(y+c)\), which simplifies to \((x-y-c)^{2}\ge 0\); the left inequality is equivalent to \(2(x^{2}+(y+c)^{2})\ge (x+y+c)^{2}\), which simplifies to \((x-(y+c))^{2}\ge 0\).  Both are true.

\medskip

\[
\boxed{\,f(x)=x+c\quad\text{for an arbitrary constant }c\ge 0\,}
\]
\end{pipelinebox}

\clearpage
\subsubsection{IMO Problem 6}
\begin{pipelinebox}[colframe=black!60,colbacktitle=black!8]{Question}
Let \(a_1,a_2,a_3,\ldots\) be an infinite sequence of positive integers greater than \(1\). Suppose that, for every positive integer \(n\), the number \(a_{n+1}\) is the smallest integer greater than \(a_n\) such that \(\gcd(a_{n+1},a_i)>1\) for every \(1\le i\le n\). Prove that there exist positive integers \(T\) and \(L\) such that \(a_{n+T}=a_n+L\) for every positive integer \(n\).
\end{pipelinebox}
\begin{pipelinebox}[colframe=blue!60!black,colbacktitle=blue!8]{K2-Horizon-7B generation}
Let \((a_n)_{n\ge 1}\) be the given sequence of integers \(>1\).  
Define a positive integer \(x>1\) to be \textbf{good} if \(\gcd(x,a_i)>1\) for every \(i\ge 1\).

\textbf{Lemma 1.} For all \(m,n\ge 1\), \(\gcd(a_m,a_n)>1\).  
\emph{Proof.} If \(m=n\) then \(a_m>1\) so \(\gcd(a_m,a_m)=a_m>1\). If \(m<n\), then by the definition of \(a_n\) (the smallest integer \(>a_{n-1}\) with \(\gcd(a_n,a_i)>1\) for all \(i\le n-1\)) we have \(m\le n-1\) and hence \(\gcd(a_n,a_m)>1\). If \(m>n\) the same argument with roles reversed gives \(\gcd(a_m,a_n)>1\). \hfill\(\square\)  
Consequently every term \(a_n\) is good: for any \(i\), if \(i<n\) then \(\gcd(a_n,a_i)>1\) by the above; if \(i>n\) then \(\gcd(a_i,a_n)>1\); and if \(i=n\) then \(a_n>1\).

\textbf{Lemma 2.} The sequence is strictly increasing and \(a_n\ge a_1+n-1\) for all \(n\ge 1\).  
\emph{Proof.} The hypothesis gives \(a_{n+1}>a_n\). By induction, \(a_n\ge a_1+n-1\). \hfill\(\square\)

\textbf{Lemma 3.} The set of good numbers that are \(\ge a_1\) is exactly \(\{a_n : n\ge 1\}\).  
\emph{Proof.} By Lemma 1 every \(a_n\) is good and \(a_n\ge a_1\). Conversely, let \(x\) be good with \(x\ge a_1\). If \(x=a_1\) we are done. If \(x>a_1\), then by Lemma 2 the sequence is unbounded, so there exists a smallest \(N\ge 2\) such that \(a_N\ge x\). Then \(a_{N-1}<x\le a_N\). If \(x=a_N\) we are done. If \(x<a_N\), then by the definition of \(a_N\) there exists \(i\le N-1\) with \(\gcd(x,a_i)=1\), contradicting that \(x\) is good. Hence \(x=a_N\). \hfill\(\square\)

\textbf{Lemma 4.} Any two good numbers are not coprime.  
\emph{Proof.} Let \(x,y\) be good. Set \(z=y^{\,a_1+1}\). Because \(y>1\) and \(a_1\ge 2\), we have \(z>a_1\). For any \(i\), since \(y\) is good there is a prime \(p\mid y\) and \(p\mid a_i\); then \(p\mid z\). Thus \(\gcd(z,a_i)\ge p>1\) for all \(i\), so \(z\) is good. By Lemma 3, \(z=a_N\) for some \(N\). Since \(x\) is good, \(\gcd(x,a_N)>1\), hence \(\gcd(x,z)>1\). If \(\gcd(x,y)=1\), then \(\gcd(x,y^{\,a_1+1})=1\), a contradiction. Therefore \(\gcd(x,y)>1\). \hfill\(\square\)

\textbf{Lemma 5.} Let \(B = a_1^{\,a_1+1}+a_1+2\). For every good number \(x\) and every \(i\ge 1\) there exists a prime \(p<B\) such that \(p\mid x\) and \(p\mid a_i\).  
\emph{Proof.} We first establish some auxiliary facts.

\textbf{Witness Lemma.} If \(k>a_1\) is not good, then there exists an index \(j\ge 1\) with \(a_j<k\) and \(\gcd(k,a_j)=1\).  
\emph{Proof.} Since \(a_n\to\infty\), choose \(N\) with \(a_N>k\). Let \(n\) be the largest integer with \(a_n<k\) (so \(a_n<k<a_{n+1}\) or \(k=a_{n+1}\)). Because \(k\) is not good, \(k\neq a_{n+1}\); thus \(a_n<k<a_{n+1}\). By the definition of \(a_{n+1}\) there is \(i\le n\) with \(\gcd(k,a_i)=1\). Take \(j=i\); then \(a_j\le a_n<k\) and \(\gcd(k,a_j)=1\). \hfill\(\square\)

\textbf{Core numbers.} A positive integer \(d>1\) is called a \textbf{core} if \(d\) is good and for every good divisor \(f\) of \(d\) we have \(d\le f\) (i.e., \(d\) has no proper good divisor). Every good number \(x\) possesses a core divisor: let \(d\) be the smallest good divisor of \(x\). Then \(d\) is good, \(d\mid x\), and if \(f\) is a good divisor of \(d\), then \(f\) is a good divisor of \(x\), so by minimality \(d\le f\). Hence \(d\) is core.

\textbf{Claim.} For every core \(d\), all prime factors of \(d\) are \(<B\).  
\emph{Proof by strong induction on \(d\).}  
\emph{Base case:} \(d\) is prime. Since \(d\) is good, \(\gcd(d,a_1)>1\); because \(d\) is prime, \(d\mid a_1\), so \(d\le a_1<B\). The only prime factor is \(d\) itself, which is \(<B\).  
\emph{Inductive step:} Assume \(d\) is composite and the claim holds for all core \(d'<d\). Let \(p\) be a prime divisor of \(d\) and write \(d=p\,e\) with \(e>1\). Because \(d\) is core, \(e\) is not good (otherwise \(e\) would be a smaller good divisor of \(d\)). Also \(e<d\).  
We distinguish two cases.

\textbf{Case 1:} \(e>a_1\).  
Since \(e\) is not good and \(e>a_1\), the Witness Lemma gives an index \(j\) with \(a_j<e\) and \(\gcd(e,a_j)=1\).  
Let \(c\) be the smallest good divisor of \(a_j\). Then \(c\) is good, \(c\mid a_j\), \(c\le a_j<e<d\), and \(c\) is core. Moreover \(\gcd(e,c)=1\) because \(c\mid a_j\) and \(\gcd(e,a_j)=1\).  
Now \(c\) and \(d\) are both good, so by Lemma 4, \(\gcd(c,d)>1\). Choose a prime \(q\mid\gcd(c,d)\). Then \(q\mid d=p\,e\), so \(q\mid p\) or \(q\mid e\). Since \(\gcd(e,c)=1\), \(q\nmid e\); hence \(q\mid p\). As \(p\) is prime, \(q=p\). Thus \(p\mid c\).  
Now \(c\) is a core number with \(c<d\). By the induction hypothesis, all prime factors of \(c\) are \(<B\); in particular \(p<B\).

\textbf{Case 2:} \(e\le a_1\).  
Consider \(k=e^{\,a_1+1}\). Because \(e>1\) and \(a_1\ge 2\), we have \(k>a_1\) and \(k\le a_1^{\,a_1+1}<B\). Also \(e\mid k\). We claim \(k\) is not good. Indeed, if \(k\) were good then for every \(i\), \(\gcd(k,a_i)>1\); every prime divisor of \(k\) divides \(e\), so \(\gcd(e,a_i)>1\) for all \(i\), making \(e\) good -- a contradiction. Hence \(k\) is not good and \(k>a_1\).  
By the Witness Lemma there exists an index \(j\) with \(a_j<k\) and \(\gcd(k,a_j)=1\). (Note that \(k\) is not a term, so the lemma applies.)  
Let \(c\) be the smallest good divisor of \(a_j\). Then \(c\) is good, \(c\mid a_j\), \(c\le a_j<k\), and \(c\) is core. Moreover \(\gcd(e,c)=1\) because \(e\mid k\) and \(\gcd(k,a_j)=1\).  
Again, \(c\) and \(d\) are good, so \(\gcd(c,d)>1\). Let \(q\) be a prime dividing \(\gcd(c,d)\). Then \(q\mid p\,e\), and since \(\gcd(e,c)=1\), \(q\mid p\); thus \(q=p\). Hence \(p\mid c\).  
Now \(c<k\le a_1^{\,a_1+1}<B\). Consequently \(p\le c<B\).

In both cases we obtain \(p<B\). This completes the induction. \hfill\(\square\)

Now return to the original statement. Let \(x\) be good and \(i\ge 1\). Take \(d\) to be the smallest good divisor of \(x\); then \(d\) is core. By the Claim, every prime factor of \(d\) is \(<B\). Since \(d\) is good, \(\gcd(d,a_i)>1\); pick a prime \(p\mid\gcd(d,a_i)\). Then \(p\mid d\mid x\) and \(p\mid a_i\) and \(p<B\). \hfill\(\square\)

\textbf{Lemma 6.} Let \(L = B!\) (the factorial of \(B\)). For every \(x>1\) we have \(\text{good}(x)\iff\text{good}(x+L)\).  
\emph{Proof.} If \(\text{good}(x)\), then for any \(i\) Lemma 5 gives a prime \(p<B\) with \(p\mid x\) and \(p\mid a_i\). Because \(p<B\), \(p\mid L\), so \(p\mid x+L\). Hence \(\gcd(x+L,a_i)\ge p>1\) and \(x+L>1\); thus \(\text{good}(x+L)\).  
Conversely, suppose \(\text{good}(x+L)\). Then for any \(i\), Lemma 5 applied to \(x+L\) gives a prime \(p<B\) with \(p\mid x+L\) and \(p\mid a_i\). Since \(p\mid L\), we have \(p\mid (x+L)-L = x\). Thus \(p\mid x\) and \(p\mid a_i\). Also \(x>1\): if \(x\le 1\) then \(x=1\) (since \(x\) is a positive integer), but then \(p\mid 1\), impossible. Therefore \(\gcd(x,a_i)>1\) for all \(i\), so \(\text{good}(x)\). \hfill\(\square\)

\textbf{Construction of \(T\) and \(L\).}  
Since \(a_1\) is good, Lemma 6 yields \(\text{good}(a_1+L)\). By Lemma 3, \(a_1+L\ge a_1\) is a term of the sequence; let \(N\) be the index with \(a_N=a_1+L\). Because \(L>0\) and the sequence is strictly increasing, \(N>1\). Define  
\[
T = N-1,\qquad L = B!.
\]  
Then \(T>0\) and \(a_{1+T}=a_1+L\).

\textbf{Lemma 7.} For every \(n\ge 1\) we have \(a_{n+T}=a_n+L\).  
\emph{Proof by induction on \(n\).}  
\emph{Base case \(n=1\):} \(a_{1+T}=a_1+L\) by construction.  
\emph{Inductive step.} Assume \(a_{n+T}=a_n+L\). Because the sequence is strictly increasing, \(a_{n+1}\) is good, so by Lemma 6, \(a_{n+1}+L\) is good. Moreover  
\[
a_{n+T}=a_n+L < a_{n+1}+L.
\]  
Since \(a_{n+T+1}\) is the smallest good number larger than \(a_{n+T}\), we obtain  
\[
a_{n+T+1}\le a_{n+1}+L.
\]  
Suppose for contradiction that \(a_{n+T+1}<a_{n+1}+L\). Set \(x=a_{n+T+1}-L\). Then  
\[
a_n < x < a_{n+1}.
\]  
By the definition of \(a_{n+1}\) (or the contrapositive of its minimality) there exists \(i\le n\) with \(\gcd(x,a_i)=1\).  
On the other hand, \(a_{n+T+1}\) is a term, hence good. Lemma 6 gives \(\text{good}(x)\) because \(\text{good}(x+L)\) holds. Therefore \(\gcd(x,a_i)>1\) for all \(i\), contradicting \(\gcd(x,a_i)=1\). Hence \(a_{n+T+1}=a_{n+1}+L\), which is exactly \(a_{(n+1)+T}=a_{n+1}+L\). \hfill\(\square\)

\textbf{Conclusion.} We have exhibited positive integers \(T\) and \(L\) (depending only on \(a_1\)) such that \(a_{n+T}=a_n+L\) for every \(n\ge 1\).

\[
\boxed{
\begin{gathered}
\text{There exist positive integers }T\text{ and }L\text{ such that}\\
a_{n+T}=a_n+L\quad\text{for all }n\ge 1.
\end{gathered}}
\]
\end{pipelinebox}

\end{document}